\documentclass[10pt,english,oneside,a4paper]{article}

\usepackage[utf8]{inputenc} 
\usepackage[english]{babel}

\usepackage[left=1.25in,right=1.25in,top=1.25in,bottom=1.25in]{geometry}

\usepackage{natbib}
\usepackage{amsmath,amssymb,amsfonts,amsthm}
\usepackage{mathtools}
\usepackage{bm}
\usepackage{dsfont} 

\usepackage{graphicx}
\usepackage[dvipsnames]{xcolor}
\usepackage{tikz}
\usetikzlibrary{matrix}
\usetikzlibrary{calc,decorations.pathreplacing}
\usepackage{pgfplots}
\pgfplotsset{compat=1.18}
\usepackage{caption}
\usepackage{subcaption}
\usepackage{multirow}

\definecolor{navyblue}{HTML}{000080}       
\definecolor{midnightblue}{HTML}{191970}   
\definecolor{cobaltblue}{HTML}{0047AB}     
\definecolor{sapphire}{HTML}{0F52BA}       
\definecolor{steelblue}{HTML}{4682B4}      
\definecolor{cornflower}{HTML}{6495ED}     
\definecolor{skyblue}{HTML}{87CEEB}        
\definecolor{babyblue}{HTML}{89CFF0}       
\definecolor{cyanblue}{HTML}{00FFFF}       
\definecolor{electricblue}{HTML}{7DF9FF}   

\usepackage{array}
\usepackage{enumitem}

\usepackage{algorithm}
\usepackage{algpseudocode}

\usepackage[style=american]{csquotes}
\usepackage[hidelinks]{hyperref}
\usepackage{cleveref}
\usepackage{xfrac} 

\usepackage{pgfplotstable}
\usepackage{booktabs}
\usepgfplotslibrary{groupplots}
\pgfplotsset{compat=1.18}

\pgfplotsset{compat=1.18}
\definecolor{modelone}{RGB}{31,119,180}   
\definecolor{modeltwo}{RGB}{0,150,136}    
\definecolor{modelthree}{RGB}{148,103,189} 

\usepackage[normalem]{ulem} 

\usepackage{benstyle_arXiv}

\usepackage{ifpdf}
\usepackage[absolute,overlay]{textpos}

\ifpdf
  \DeclareGraphicsExtensions{.eps,.pdf,.png,.jpg}
\else
  \DeclareGraphicsExtensions{.eps}
\fi

\numberwithin{equation}{section}

\theoremstyle{plain}
\newtheorem{theorem}{Theorem}[section]

\newtheorem{proposition}[theorem]{Proposition}
\newtheorem{corollary}[theorem]{Corollary}

\theoremstyle{definition}
\newtheorem{definition}[theorem]{Definition}

\theoremstyle{definition}
\newtheorem{assumption}[theorem]{Assumption}

\theoremstyle{remark}
\newtheorem{remark}[theorem]{Remark}

\newcommand{\bigcupdot}{\mathop{\vphantom{\bigcup}\mathpalette\docupdot\relax}\displaylimits}
\newcommand{\docupdot}[2]{%
  \text{\ooalign{$#1\bigcup$\cr\hidewidth$#1\cdot$\hidewidth\cr}}%
}

\newcommand{\cupdot}{\mathbin{\mathaccent\cdot\cup}}
\makeatletter

\AtBeginEnvironment{figure}{\setlength{\tabcolsep}{2pt}}

\title{Deterministic Bounds on Accuracy and Stability of Metric-Space Regression and Classification}

\author{Nina M.~Gottschling
  \thanks{Computing and Computational Sciences, Oak Ridge National Laboratory, Oak Ridge, Tennessee.}
  \and Jakob Gawlikowski
  \thanks{German Aerospace Center (DLR), Remote Sensing Technology Institute, Wessling, Germany.}
\and Michele Caprio
  \thanks{Department of Computer Science, University of Warwick, Coventry, United Kingdom.}}

\usepackage{amsopn}

\ifpdf
\hypersetup{
  pdftitle={Accuracy Bounds for Regression and Classification},
  pdfauthor={Nina M. Gottschling, Jakob Gawlikowski and Michele Caprio}
}
\fi



\begin{document}

\maketitle

\begin{textblock}{13.3}(1.4,15)\noindent\fontsize{6}{6}\selectfont\textcolor{black!30}{\vspace{15pt}\newline
This manuscript has been co-authored by UT-Battelle, LLC, under contract DE-AC05-00OR22725 with the US Department of Energy (DOE). The US government retains and the publisher, by accepting the article for publication, acknowledges that the US government retains a nonexclusive, paid-up, irrevocable, worldwide license to publish or reproduce the published form of this manuscript, or allow others to do so, for US government purposes. DOE will provide public access to these results of federally sponsored research in accordance with the DOE Public Access Plan (http://energy.gov/downloads/doe-public-access-plan).}\end{textblock}

\begin{abstract}
Data-driven approaches often face ambiguity in datasets due to sensor or target noise, insufficient information, and noise in the input data. While best-fit machine learning methods can accommodate such noise through uncertainty estimates, they are prone to overfitting and it is challenging to quantify ambiguity of the relation between inputs and targets. We introduce a geometric framework that quantifies this ambiguity through the average kernel size, computed with data-dependent, i.e. model-independent, feasible sets derived from worst-case input and target noise. The framework provides computable, deterministic, non-asymptotic bounds on the best achievable empirical error and accuracy for a finite, constructed dataset, independent of the predictive model class, and applies to regression and classification with pseudo-metric input spaces and metric target spaces. The ambiguity of the input-target relation implies three consequences: an empirical error floor, an accuracy–stability trade-off, and, under standard probabilistic assumptions, a strictly positive distribution–training error gap when training below the accuracy bounds. Under standard Borel assumptions for Euclidean norms, the distributional analogue of the average kernel size is related to Bayes risk as it reduces to the optimal distribution root-mean-squared error times $\sqrt{2}$. Experiments spanning tabular regression, image regression and classification datasets yield nontrivial bounds on the best achievable empirical error and accuracy on a constructed dataset, alongside descriptive comparisons with model performance on the original datasets. The bounds provide a computable, model-independent reference scale on the constructed dataset and comparisons with model performance on the original dataset are descriptive.
\end{abstract}


\section{Introduction}

\begin{figure}[h!]
\centering
\includegraphics[width=0.85\textwidth]{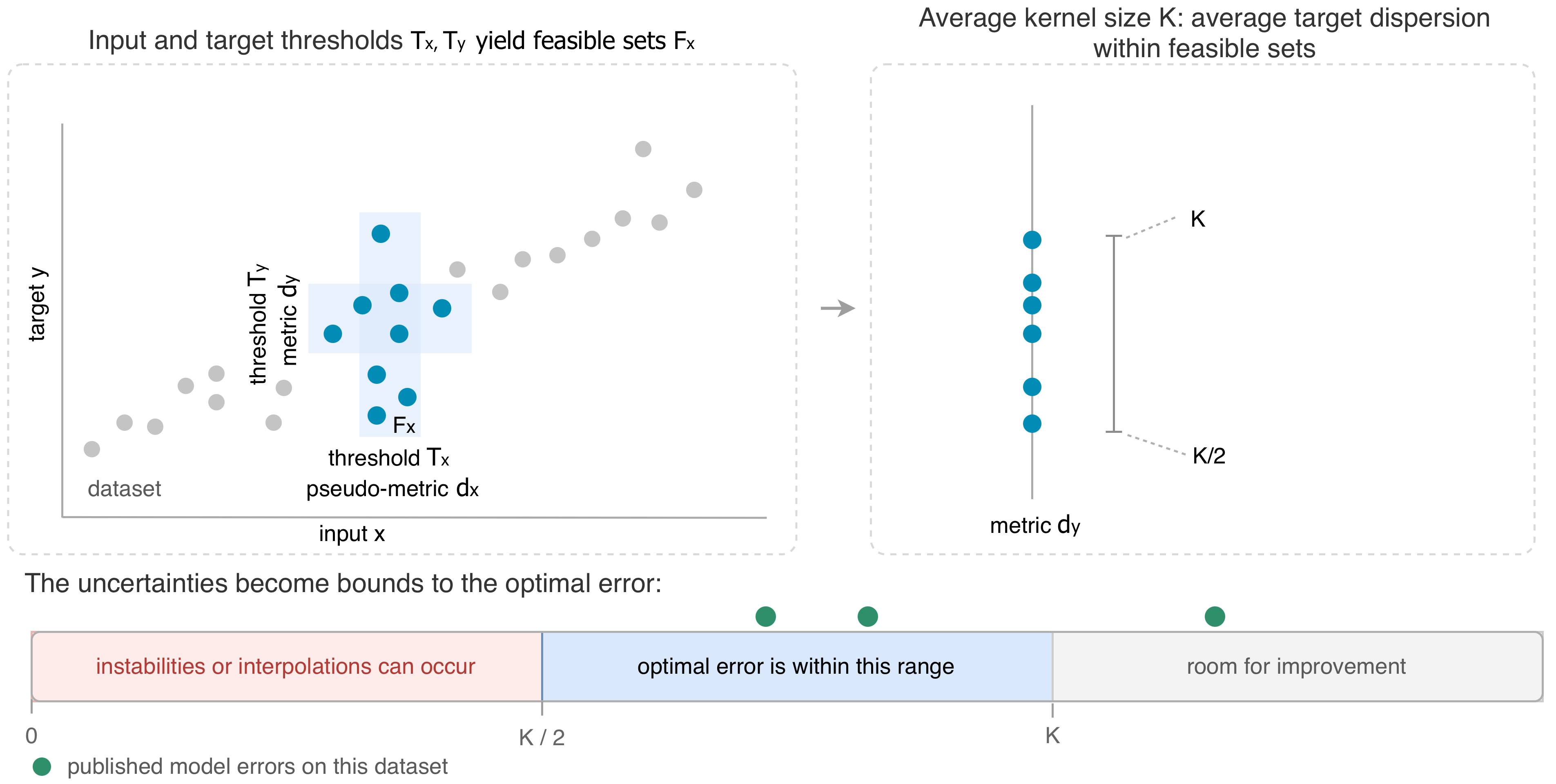}
\caption{\textbf{Accuracy bounds: from data ambiguity to a reference scale on the constructed dataset.} For an input $x$, the thresholds define a feasible set $F_x$ (Definition~\ref{def:feassetd}) of target values lying within the worst-case noise distance. The target values within this set may differ. The average kernel size, $K$, provides a reference scale for the best achievable empirical error on a finite constructed dataset; the term ``average kernel size'' is motivated by its namesake in inverse problems~\cite{gottschling2023existence}, but here we define it directly from the constructed dataset without requiring a forward model. In this schematic, the optimal error on the constructed dataset lies between $K/2$ and $K$. Model errors reported on the original dataset are shown for descriptive comparison. Both the finite feasible sets and the value of $K$ depend on the chosen pseudo-metric, metric, and thresholds. For models evaluated only on the original dataset, such comparisons are heuristic unless additional assumptions, such as those in Proposition~\ref{prop:originalenlarged}, are invoked.}\label{fig:hero}
\end{figure}

\noindent We express regression and classification problems using the following general framework:
\begin{align}\label{eq:sampling10}
	\text{predict } y \in \mathcal{M}_1 \subset \mathcal{Y} \text{ given } x \in \mathcal{M}_2 \subset \mathcal{X},
\end{align}
where the available input-target pairs $(y,x) \in \mathcal{M}_1\times\mathcal{M}_2$ are given by a finite dataset of tuples $\mathcal{D}_M = \{(y_m,x_m)\}_{m =1}^M \subset \mathcal{M}_1 \times \mathcal{M}_2$ for some $M \in \mathbb{N}$. Here, $y$ represents the data of interest, while $x$ denotes the available data from which we wish to predict $y$. The sets $\mathcal{M}_1\subset \mathcal{Y}$ and $\mathcal{M}_2\subset \mathcal{X}$ describe the sets of target and input signals of interest. The dataset $\mathcal{D}_M$ is deliberately defined in this general way to encompass many known cases and settings. For example, the ambient input set $\mathcal{X}$ may consist of tensors for image-based regression and classification problems or of a structure supporting tabular data, and the set $\mathcal{M}_2 \subset \mathcal{X}$ may consist of all admissible images. For classification, the ambient target set $\mathcal{Y}$ consists of one-hot encoded vectors or more generally $\mathcal{Y}=\{y \in [0,1]^C: \sum_{c=1}^C y_c =1\}$ for $C \in \mathbb{N}$ classes.

\noindent In applications, an input may correspond to multiple admissible targets, even if the observed dataset $\mathcal{D}_M$ contains a unique target for that input. To characterize such ambiguity, we introduce a multiset of admissible input-target pairs $\mathcal{R}$ satisfying $\mathcal{D}_M\subset\mathcal{R}$. For a fixed input $x$, we define its feasible set in the following Definition \ref{def:feassetd}. If, for a fixed input $x$, the targets are ambiguous, then the feasible set can be a non-singleton. As the disjoint union preserves multiplicities, it can be a multiset, which is an unordered collection of elements where each element can appear more than once.

\begin{definition}[Feasible sets]\label{def:feassetd} 
Let $\mathcal{R}$ be an underlying admissible input-target multiset. For an input $x$ obtained from a sample $(y,x) \in \mathcal{R}$ let the \emph{feasible set} $F_x$ be defined as 
\begin{align}\label{eq:feasset}
	F_x:= \bigcupdot_{(y,x) \in \mathcal{R}}\{y\} \subset \mathcal{M}_1.
\end{align}
As $\cupdot$ is the disjoint union, the \emph{feasible set} $F_x$ is a multiset. 
\end{definition}

\noindent A non-singleton feasible set indicates that multiple targets are admissibly associated with the same input. If $y,y'\in F_x$ with $y\neq y'$, then $(y,x),(y',x) \in \mathcal{R}$ although these pairs need not both be contained in the observed finite dataset $\mathcal{D}_M$.  Such ambiguity may arise, for example, from noisy observations, where it is generally impossible for a given sample to determine whether the input, target, or both have been corrupted, or to identify the underlying noise mechanism. Consequently, for an individual input-target sample we generally cannot determine whether the sample has been corrupted by noise. To account for this ambiguity, we consider an admissible enlargement \(\mathcal D_{M'}\) of the observed dataset and impose the following assumption.

\begin{assumption}[Existence of an admissible enlargement]\label{as:feassetd} 
We assume that there exists a multiset of admissible input-target pairs \(\mathcal{R}\) with \(\mathcal{D}_M \subset \mathcal{R}\). Relative to this admissible multiset, we assume that there exists a multiset
\begin{align}
\mathcal{D}_{M'} = \{(y_m,x_m)\}_{m =1}^{M'} \subset \mathcal{R}
\end{align}
such that \(\mathcal{D}_M \subset \mathcal{D}_{M'}\) and, for some \(x \in \mathcal{M}_2\), there exist distinct \(y,y' \in \mathcal{M}_1\) with \(y,y' \in F_x\).
\end{assumption}

\noindent Here, \(\mathcal D_{M'}\) is not meant to be uniquely determined or algorithmically recovered from \(\mathcal D_M\); it is an admissible enlargement of the observed sample used to represent latent ambiguity in the input-target relation. The condition that some \(F_x\) contains two distinct values \(y \neq y'\) excludes trivial cases and ensures a genuinely ambiguous regime.

\noindent The goal of solving regression and classification problems of the form \eqref{eq:sampling10} is to produce an approximation or set of approximations $\hat{y} \subset \mathcal{Y}$ to the true target $y$ from an input $x$. Any algorithm or method that produces a potentially \textbf{set-valued regressor or classifier} mapping data points $x$ in $\mathcal M_2$ to approximations of the true target $y$ is denoted by 
\begin{align*}
\phi:\mathcal M_2\rightrightarrows\mathcal Y,
\end{align*}
with $\phi(x)=\hat y \subset \mathcal{Y}$. Examples include optimization-based methods, such as random forests, \cite{segal2004machine,shwartz2022tabular,mcelfresh2023neural} which in some cases outperform deep neural networks on tabular data, and supervised or unsupervised deep learning methods \cite{lathuiliere2019comprehensive}, such as transformer-based architectures \cite{han2022survey}. 

\noindent The \textbf{empirical error of a (set-valued) regressor} $\phi: \mathcal{M}_2 \rightrightarrows \mathcal{Y}$ for approximately solving \eqref{eq:sampling10} is often quantified by the empirical root-$p$ mean of the residuals between the predictions and targets over the $M \in \mathbb{N}$ samples $\mathcal{D}_M=\{(y_m,x_m)\}_{m=1}^M$, and can be denoted by
\begin{align}\label{eq:empiricalerror}
    \mathcal{L}(\mathcal{D}_M, \phi,p) = \left(\frac{1}{M} \sum_{m=1}^M d_{\mathcal{Y}}^H(\{y_m\},\phi(x_m))^p\right)^{1/p},
\end{align}
\noindent for some $p \in [1,\infty)$ and a metric $d_{\mathcal{Y}}$ on $\mathcal{M}_1$, where the Hausdorff distance is defined by $d_{\mathcal{Y}}^H(\{y\},\phi(x)) := \sup_{z\in \phi(x)}d_{\mathcal{Y}}(y,z)$. For $p=1$, this is the mean absolute error (MAE), and for $p=2$, it corresponds to the root-mean-squared error (RMSE); its square is the empirical risk or also mean squared loss. The MAE and RMSE are commonly used to evaluate the error of regressors, see, e.g., \cite[Chapter 1]{bickel2015mathematical} and \cite[Chapter 1]{murphy2022probabilistic}.

\noindent For classification, the \textbf{empirical accuracy of a (set-valued) classifier} $\phi: \mathcal{M}_2 \rightrightarrows \mathcal{Y}$ can be measured by the Hausdorff distance induced by the normalized Hamming metric \(d_H\), so that
\begin{align}\label{eq:empiricalaccuracy}
\operatorname{Acc}(\mathcal{D}_M,\phi)=1-\frac{1}{M}\sum_{m=1}^{M} d_H^H\!\bigl(\{y_m\},\phi(x_m)\bigr).
\end{align}
\noindent The normalization of the Hamming metric is introduced in detail in Section \ref{sec:preliminaries}. 

\noindent A common objective in regression problems across various applications is to find a regressor that minimizes the loss \eqref{eq:empiricalerror} in a given class of functions, such as finding the optimal architecture and parameter values for deep neural networks or, for example, in least squares regression \cite{arlot2009data}. In classification problems, this changes to maximizing the accuracy in \eqref{eq:empiricalaccuracy}. This is commonly referred to as \emph{empirical risk minimization}, where the \textit{"[...] true goal is to minimize the expected loss on future data that we have not yet seen. That is, we want to generalize, rather than just do well on the training set."}, \cite[Chapter 1]{murphy2022probabilistic}. Under additional assumptions, the hope is that this procedure solves regression and classification problems of the form \eqref{eq:sampling10} and that the resulting regressors and classifiers produce, for any input \(x\), an approximation or set of approximations $\hat{y}$ to the true target $y$.

\subsection{Challenges}\label{sec:challenges}

Before summarizing the related work, we highlight three challenges that motivate the framework developed here.

\noindent \textbf{(i) Non-uniqueness of the target given the observed input.}
In many regression and classification problems, the available input does not determine a unique target, satisfying Assumption \ref{as:feassetd}. This may occur because of measurement noise, omitted variables, or annotation uncertainty and is commonly referred to as \emph{aleatoric uncertainty} in the field of machine learning, see \cite{murphy2022probabilistic, der2009aleatory,gawlikowski2023survey}. In such settings, even an ideal regressor or classifier cannot recover a single deterministic response from the observed input alone. Consequently, part of the prediction error is intrinsic to the data-generating relation and cannot be removed by choosing a more expressive model class or a better optimization method.

\noindent \textbf{(ii) The need for computable, model-independent bounds on finite datasets.}
In computationally demanding settings, it is useful to estimate achievable empirical error and accuracy before extensive model selection and training. This motivates finite-sample bounds that are computable from the dataset itself and independent of a specific model class. To the best of our knowledge, related work does not provide such bounds in this form; see Table \ref{tab:comparison} and Section \ref{sec:relwork}.

\noindent \textbf{(iii) Expressive regressors and classifiers do not evade intrinsic accuracy limits.} A third challenge is that expressive regressors and classifiers can fit the training dataset without resolving ambiguity inherent in the input-target relation. For example, for a finite dataset $\mathcal{D}_M$ with no ambiguous targets that have the same input, interpolation by a large enough model is possible, but may come at the cost of poor generalization or overfitting, \cite[Chapter 1]{murphy2022probabilistic} and \cite{zhang2016understanding,zhang2021understanding}, and potentially a large Lipschitz constant. Hence, approximation power alone does not determine the best achievable test error and accuracy.

\begin{table}[t]
\centering
\scriptsize
\renewcommand{\arraystretch}{1.15}
\begin{tabular}{>{\raggedright}p{3cm}|>{\raggedright}p{1.5cm}p{1.8cm}>{\raggedright}p{2.6cm}>{\raggedright\arraybackslash}p{2.7cm}}
\toprule
\textbf{Quantity} & \textbf{Object} & \textbf{Model-class dependent?}
& \textbf{Sampling assumptions} & \textbf{Directly computable from one dataset?} \\
\midrule
Bayes risk
& Distribution decision risk
& No
& A data-generating law is required
& Generally not \\
Conditional variance
& Distribution conditional dispersion
& No
& Usually i.i.d. sampling plus regularity or replication
& Only under additional structure \\
VC/Rademacher complexity
& Model-class complexity
& Yes
& Usually i.i.d. sampling
& Sometimes, bounds a restricted class \\
\hline 
\textbf{Average kernel size (proposed)}
& Quantity in accuracy bounds
& No
& None for the deterministic Theorem \ref{thm:lowerboundapprx}
& Yes, once metrics and thresholds are fixed \\
\hline 
\textbf{Distributional analogue average kernel size  (proposed)}
& Conditional pairwise dispersion
& No
& Standard Borel setting and integrability
& Not without a sampling or estimation scheme \\
\bottomrule
\end{tabular}
\caption{The average kernel size and its distributional analogue play different roles and are distinguished explicitly.}
\label{tab:comparison}
\end{table}

\subsection{Contributions}

We develop a framework for computing deterministic bounds on achievable regression and classification accuracy from finite datasets. Here, we introduce a metric-space framework that encodes input-target ambiguity through data-dependent feasible sets, with pseudo-metrics on inputs and  metrics on targets. The main contributions are as follows.

\smallskip
\noindent We establish an \emph{accuracy-stability trade-off}, Proposition \ref{prop:instabilityaccuracy}. We show that a too small pointwise error on separated, ambiguous targets at nearby, but distinct, inputs forces a large (local) Lipschitz constant and that ambiguous targets produce a pointwise error floor.

\smallskip
\noindent We derive computable, deterministic, non-asymptotic \emph{lower and upper bounds on the best achievable empirical error} and accuracy on datasets defined in Assumption \ref{as:feassetd}. The lower bounds in Theorem~\ref{thm:lowerboundapprx} and Corollary~\ref{thm:lowerboundapprxclass} hold for all regressors and classifiers, while for distinct inputs the upper bound holds and the optimum is attained by an upper hemicontinuous compact-valued regressor or classifier. 

\smallskip
\noindent Using a graph-based argument, we show that for Lipschitz continuous regressors and classifiers, the \emph{difference between empirical error, respectively accuracy, on the constructed and original datasets is bounded} in terms of the Lipschitz constant and input-target ambiguity for finite feasible sets of equal cardinality (Proposition~\ref{prop:originalenlarged}).

\smallskip
\noindent Under standard Borel assumptions, we show that regressors and classifiers \emph{achieving errors below the conditional ambiguity scale, necessarily incur a strictly positive generalization gap} (Proposition~\ref{thm:gengapbd}). Under Euclidean squared loss, the Bayes squared risk equals one half the squared distributional analogue average kernel size.

\smallskip
\noindent Under an ideal two-stage design with inputs sampled from the input law and conditionally independent repeated targets, we prove a \emph{finite-sample concentration bound for the empirical average kernel size estimator} (Proposition~\ref{thm:convavgalgnon}). This differs from the average kernel size computed by Algorithm~\ref{alg:feasset}, which is a data-dependent geometric quantity unless additional sampling assumptions are imposed.

\smallskip
\noindent We \emph{provide open-source code for reproducibility and experiments illustrating the computation of the deterministic bounds across regression and classification tasks}, their dependence on the chosen pseudo-metric, thresholds, and feasible set sizes, and their comparison with benchmark models and locally trained models. The experiments examine whether the bounds are nontrivial, how selected model performances compare numerically with the resulting intervals, and how sensitive the intervals are to the finite feasible sets (Section~\ref{sec:applications}). Figures~\ref{fig:regression-intervals} and \ref{fig:classification-intervals} provide descriptive visual comparisons with selected model errors and accuracies; detailed qualifications are deferred to Section~\ref{sec:applications}.

\begin{figure}[h!]

\centering
\begin{minipage}{\textwidth}
\begin{tikzpicture}

\begin{axis}[
    xmode=log,
    width=0.87\linewidth,
    height=5cm,
    xmin=0.008,
    xmax=350,
    ymin=0.5,
    ymax=6.5,
    xlabel={RMSE [units of target]},
    ytick={1,2,3,4,5,6},
    yticklabels={
        Critical Heat Flux,
        Concrete Strength,
        Energy Efficiency $Y_2$,
        Energy Efficiency $Y_1$,
        Tropical Cyclone,
        Synthetic
    },
    xmajorgrids=true,
    xminorgrids=true,
    grid style={gray!20},
    tick label style={font=\small},
    label style={font=\small},
    legend style={
    at={(0.5,-0.28)},
    anchor=north,
    font=\small,
    draw=none,
    legend columns=5,
    column sep=1.2em,
    /tikz/every even column/.append style={column sep=1.2em}
},
]


\addplot[
    very thick,
    blue!70!black,
    forget plot
] coordinates {(0.0102,6) (0.0203,6)};

\draw[very thick,blue!70!black]
    (axis cs:0.0102,5.88) -- (axis cs:0.0102,6.12);
\draw[very thick,blue!70!black]
    (axis cs:0.0203,5.88) -- (axis cs:0.0203,6.12);

\addplot[
    very thick,
    blue!70!black,
    forget plot
] coordinates {(3.83,5) (7.65,5)};

\draw[very thick,blue!70!black]
    (axis cs:3.83,4.88) -- (axis cs:3.83,5.12);
\draw[very thick,blue!70!black]
    (axis cs:7.65,4.88) -- (axis cs:7.65,5.12);

\addplot[
    very thick,
    blue!70!black,
    forget plot
] coordinates {(0.28,4) (0.55,4)};

\draw[very thick,blue!70!black]
    (axis cs:0.28,3.88) -- (axis cs:0.28,4.12);
\draw[very thick,blue!70!black]
    (axis cs:0.55,3.88) -- (axis cs:0.55,4.12);

\addplot[
    very thick,
    blue!70!black,
    forget plot
] coordinates {(0.78,3) (1.55,3)};

\draw[very thick,blue!70!black]
    (axis cs:0.78,2.88) -- (axis cs:0.78,3.12);
\draw[very thick,blue!70!black]
    (axis cs:1.55,2.88) -- (axis cs:1.55,3.12);

\addplot[
    very thick,
    blue!70!black,
    forget plot
] coordinates {(1.41,2) (2.81,2)};

\draw[very thick,blue!70!black]
    (axis cs:1.41,1.88) -- (axis cs:1.41,2.12);
\draw[very thick,blue!70!black]
    (axis cs:2.81,1.88) -- (axis cs:2.81,2.12);

\addplot[
    very thick,
    blue!70!black,
    forget plot
] coordinates {(106.49,1) (212.97,1)};

\draw[very thick,blue!70!black]
    (axis cs:106.49,0.88) -- (axis cs:106.49,1.12);
\draw[very thick,blue!70!black]
    (axis cs:212.97,0.88) -- (axis cs:212.97,1.12);


\addplot[
    very thick,
    dashed,
    orange!85!black,
    forget plot
] coordinates {(0.0113,6.2) (0.0225,6.2)};

\draw[very thick,orange!85!black]
    (axis cs:0.0113,6.08) -- (axis cs:0.0113,6.32);
\draw[very thick,orange!85!black]
    (axis cs:0.0225,6.08) -- (axis cs:0.0225,6.32);

\addplot[
    very thick,
    dashed,
    orange!85!black,
    forget plot
] coordinates {(3.38,5.2) (6.76,5.2)};

\draw[very thick,orange!85!black]
    (axis cs:3.38,5.08) -- (axis cs:3.38,5.32);
\draw[very thick,orange!85!black]
    (axis cs:6.76,5.08) -- (axis cs:6.76,5.32);



\addplot[
    only marks,
    mark=*,
    mark size=3pt,
    color=modelone,
    forget plot
] coordinates {
    (0.0169,5.82)
};

\addplot[
    only marks,
    mark=square*,
    mark size=3pt,
    color=modeltwo,
    forget plot
] coordinates {
    (0.0161,6.00)
};

\addplot[
    only marks,
    mark=triangle*,
    mark size=3.5pt,
    color=modelthree,
    forget plot
] coordinates {
    (0.0607,6.18)
};


\addplot[
    only marks,
    mark=*,
    mark size=3pt,
    color=modelone,
    forget plot
] coordinates {(6.26,4.84)};

\addplot[
    only marks,
    mark=square*,
    mark size=3pt,
    color=modeltwo,
    forget plot
] coordinates {(6.42,5.00)};

\addplot[
    only marks,
    mark=triangle*,
    mark size=3.5pt,
    color=modelthree,
    forget plot
] coordinates {(6.46,5.16)};


\addplot[
    only marks,
    mark=*,
    mark size=3pt,
   color=modelone,
    forget plot
] coordinates {(0.51,3.84)};

\addplot[
    only marks,
    mark=square*,
    mark size=3pt,
    color=modeltwo,
    forget plot
] coordinates {(0.56,4.00)};

\addplot[
    only marks,
    mark=triangle*,
    mark size=3.5pt,
    color=modelthree,
    forget plot
] coordinates {(0.67,4.16)};


\addplot[
    only marks,
    mark=*,
    mark size=3pt,
    color=modelone,
    forget plot
] coordinates {(1.28,2.84)};

\addplot[
    only marks,
    mark=square*,
    mark size=3pt,
    color=modeltwo,
    forget plot
] coordinates {(1.64,3.00)};

\addplot[
    only marks,
    mark=triangle*,
    mark size=3.5pt,
    color=modelthree,
    forget plot
] coordinates {(1.79,3.16)};


\addplot[
    only marks,
    mark=*,
    mark size=3pt,
    color=modelone,
    forget plot
] coordinates {(1.97,1.84)};

\addplot[
    only marks,
    mark=square*,
    mark size=3pt,
    color=modeltwo,
    forget plot
] coordinates {(2.08,2.00)};

\addplot[
    only marks,
    mark=triangle*,
    mark size=3.5pt,
    color=modelthree,
    forget plot
] coordinates {(2.10,2.16)};


\addplot[
    only marks,
    mark=*,
    mark size=3pt,
    color=modelone,
    forget plot
] coordinates {(247.76,0.84)};

\addplot[
    only marks,
    mark=square*,
    mark size=3pt,
    color=modeltwo,
    forget plot
] coordinates {(251.46,1.00)};

\addplot[
    only marks,
    mark=triangle*,
    mark size=3.5pt,
    color=modelthree,
    forget plot
] coordinates {(266.98,1.16)};

\end{axis}


\draw[
    very thick,
    blue!70!black
]
(0.5,-1.15) -- (1.1,-1.15);

\draw[
    very thick,
    blue!70!black
]
(0.5,-1.22) -- (0.5,-1.08);

\draw[
    very thick,
    blue!70!black
]
(1.1,-1.22) -- (1.1,-1.08);

\node[
    anchor=west,
    font=\scriptsize
]
at (1.2,-1.15)
{train interval};

\draw[
    very thick,
    dashed,
    orange!85!black
]
(3.1,-1.15) -- (3.6,-1.15);

\draw[
    very thick,
    orange!85!black
]
(3.1,-1.22) -- (3.1,-1.08);

\draw[
    very thick,
    orange!85!black
]
(3.6,-1.22) -- (3.6,-1.08);

\node[
    anchor=west,
    font=\scriptsize
]
at (3.7,-1.15)
{test interval};


\fill[
    modelone
]
(5.7,-1.15) circle (3pt);

\node[
    anchor=west,
    font=\scriptsize
]
at (5.85,-1.15)
{$\phi_{1st}$};

\fill[
    modeltwo
]
(6.73,-1.07)
rectangle 
(6.9,-1.25);

\node[
    anchor=west,
    font=\scriptsize
]
at (6.98,-1.15)
{$\phi_{2nd}$};

\fill[
    modelthree
]
(8.0,-1.06)
--
(7.9,-1.25)
--
(8.1,-1.25)
--
cycle;

\node[
    anchor=west,
    font=\scriptsize
]
at (8.15,-1.15)
{$\phi_{3rd}$};
\end{tikzpicture}
\caption{
\small{\textbf{Deterministic regression accuracy intervals and regressor errors.} The theoretical interval from Theorem~\ref{thm:lowerboundapprx} is
$\left[
1/2\operatorname{Kersize}(\mathcal D_{M'},p)_K,\,
\operatorname{Kersize}(\mathcal D_{M'},p)_K
\right]$. All scores are RMSE values in the units of the target variables of three regressors on the test datasets
$\mathcal{D}_{M_{\mathrm{test}}}$. Theoretical intervals apply to the admissible enlarged dataset \(\mathcal D_{M'}\) constructed from the observed dataset \(\mathcal D_M\); plotted RMSE values are on $\mathcal D_{M_{\mathrm{test}}}$. Numerical values and details are provided in Table \ref{tab:regression-results} in Section \ref{sec:applications}.}
}
\label{fig:regression-intervals}
\end{minipage}

\begin{minipage}{\textwidth}
\centering
\vspace{5pt}
\begin{tikzpicture}

\begin{axis}[
    xmode=normal,
    width=0.9\linewidth,
    height=4cm,
    xmin=75,
    xmax=101,
    ymin=-0.2,
    ymax=3.5,
    xlabel={Classification Accuracy [\%]},
    ytick={0,1,2,3},
    yticklabels={
        MNIST,
        EUROSAT,
        RESISC45$^*$,
        CIFAR-10
    },
    xmajorgrids=true,
    xminorgrids=true,
    grid style={gray!20},
    tick label style={font=\small},
    label style={font=\small},
    legend style={
        at={(0.5,-0.28)},
        anchor=north,
        font=\small,
        draw=none,
        legend columns=5,
        column sep=1.2em,
        /tikz/every even column/.append style={column sep=1.2em}
    },
]


\addplot[
    very thick,
    blue!70!black,
    forget plot
] coordinates {(85.61,3) (92.81,3)};

\draw[very thick,blue!70!black]
    (axis cs:85.61,2.88) -- (axis cs:85.61,3.12);
\draw[very thick,blue!70!black]
    (axis cs:92.81,2.88) -- (axis cs:92.81,3.12);

\addplot[
    very thick,
    blue!70!black,
    forget plot
] coordinates {(83.85,2) (91.93,2)};

\draw[very thick,blue!70!black]
    (axis cs:83.85,1.88) -- (axis cs:83.85,2.12);
\draw[very thick,blue!70!black]
    (axis cs:91.93,1.88) -- (axis cs:91.93,2.12);

\addplot[
    very thick,
    blue!70!black,
    forget plot
] coordinates {(97.78,1) (98.89,1)};

\draw[very thick,blue!70!black]
    (axis cs:97.78,0.88) -- (axis cs:97.78,1.12);
\draw[very thick,blue!70!black]
    (axis cs:98.89,0.88) -- (axis cs:98.89,1.12);

\addplot[
    very thick,
    blue!70!black,
    forget plot
] coordinates {(100,0) (100,0)};

\draw[very thick,blue!70!black]
    (axis cs:100,-0.12) -- (axis cs:100,0.12);
\draw[very thick,blue!70!black]
    (axis cs:100,-0.12) -- (axis cs:100,0.12);


\addplot[
    very thick,
    dashed,
    orange!85!black,
    forget plot
] coordinates {(83.76,3.2) (91.88,3.2)};

\draw[very thick,orange!85!black]
    (axis cs:83.76,3.08) -- (axis cs:83.76,3.32);
\draw[very thick,orange!85!black]
    (axis cs:91.88,3.08) -- (axis cs:91.88,3.32);

\addplot[
    very thick,
    dashed,
    orange!85!black,
    forget plot
] coordinates {(82.36,2.2) (91.18,2.2)};

\draw[very thick,orange!85!black]
    (axis cs:82.36,2.08) -- (axis cs:82.36,2.32);
\draw[very thick,orange!85!black]
    (axis cs:91.18,2.08) -- (axis cs:91.18,2.32);

\addplot[
    very thick,
    dashed,
    orange!85!black,
    forget plot
] coordinates {(96.89,1.2) ( 98.44,1.2)};

\draw[very thick,orange!85!black]
    (axis cs:96.89,1.08) -- (axis cs:96.89,1.32);
\draw[very thick,orange!85!black]
    (axis cs:98.44,1.08) -- (axis cs:98.44,1.32);

\addplot[
    very thick,
    dashed,
    orange!85!black,
    forget plot
] coordinates {(100,0.2) (100,0.2)};

\draw[very thick,orange!85!black]
    (axis cs:100,0.08) -- (axis cs:100,0.32);
\draw[very thick,orange!85!black]
    (axis cs:100,0.08) -- (axis cs:100,0.32);



\addplot[
    only marks,
    mark=*,
    mark size=3pt,
    color=modelone,
    forget plot
] coordinates {(91.19,3)};

\addplot[
    only marks,
    mark=square*,
    mark size=3pt,
    color=modeltwo,
    forget plot
] coordinates {(89.61,3)};

\addplot[
    only marks,
    mark=triangle*,
    mark size=3.5pt,
    color=modelthree,
    forget plot
] coordinates {(83.20,3)};


\addplot[
    only marks,
    mark=*,
    mark size=3pt,
    color=modelone,
    forget plot
] coordinates {(90.36,2)};

\addplot[
    only marks,
    mark=square*,
    mark size=3pt,
    color=modeltwo,
    forget plot
] coordinates {(86.02,2)};

\addplot[
    only marks,
    mark=triangle*,
    mark size=3.5pt,
    color=modelthree,
    forget plot
] coordinates {(85.16,2)};


\addplot[
    only marks,
    mark=*,
    mark size=3pt,
    color=modelone,
    forget plot
] coordinates {(98.00,1.15)};

\addplot[
    only marks,
    mark=square*,
    mark size=3pt,
    color=modeltwo,
    forget plot
] coordinates {(98.00,0.9)};

\addplot[
    only marks,
    mark=triangle*,
    mark size=3.5pt,
    color=modelthree,
    forget plot
] coordinates {(97.44,1)};


\addplot[
    only marks,
    mark=*,
    mark size=3pt,
    color=modelone,
    forget plot
] coordinates {(99.53,0.26)};

\addplot[
    only marks,
    mark=square*,
    mark size=3pt,
    color=modeltwo,
    forget plot
] coordinates {(99.45,0.03)};

\addplot[
    only marks,
    mark=triangle*,
    mark size=3.5pt,
    color=modelthree,
    forget plot
] coordinates {(99.43,0.0)};

\end{axis}



\draw[
    very thick,
    blue!70!black
]
(0.5,-1.15) -- (1.1,-1.15);

\draw[
    very thick,
    blue!70!black
]
(0.5,-1.22) -- (0.5,-1.08);

\draw[
    very thick,
    blue!70!black
]
(1.1,-1.22) -- (1.1,-1.08);

\node[
    anchor=west,
    font=\scriptsize
]
at (1.2,-1.15)
{train interval};

\draw[
    very thick,
    dashed,
    orange!85!black
]
(3.1,-1.15) -- (3.6,-1.15);

\draw[
    very thick,
    orange!85!black
]
(3.1,-1.22) -- (3.1,-1.08);

\draw[
    very thick,
    orange!85!black
]
(3.6,-1.22) -- (3.6,-1.08);

\node[
    anchor=west,
    font=\scriptsize
]
at (3.7,-1.15)
{test interval};


\fill[
    modelone
]
(5.7,-1.15) circle (3pt);

\node[
    anchor=west,
    font=\scriptsize
]
at (5.85,-1.15)
{$\phi_{1st}$};

\fill[
    modeltwo
]
(6.73,-1.07)
rectangle
(6.9,-1.25);

\node[
    anchor=west,
    font=\scriptsize
]
at (6.98,-1.15)
{$\phi_{2nd}$};

\fill[
    modelthree
]
(8.0,-1.06)
--
(7.9,-1.25)
--
(8.1,-1.25)
--
cycle;

\node[
    anchor=west,
    font=\scriptsize
]
at (8.15,-1.15)
{$\phi_{3rd}$};

\end{tikzpicture}

\caption{
\small{\textbf{Deterministic classification accuracy intervals and classifier accuracies.} Theoretical intervals from Corollary~\ref{thm:lowerboundapprxclass} are reported as percentages; explicitly, the
interval is 
$
100 \times \left[
1-\operatorname{Kersize}(\mathcal D_{M'},1)_K,\,
1-1/2\operatorname{Kersize}(\mathcal D_{M'},1)_K
\right].
$
Accuracies are reported on the \([0,100]\) scale and from classifiers on the test datasets $\mathcal{D}_{M_{\mathrm{test}}}$. Intervals are for $\mathcal D_{M'}$; plotted benchmark accuracies are on $\mathcal D_M$ and may use different splits/protocols, so they are descriptive only. For CIFAR-10 the order of classifier names is changed relative to the numerical values provided in Table \ref{tab:classification-results} in Section \ref{sec:applications}.}
}

\label{fig:classification-intervals}
\end{minipage}
\end{figure}

\noindent \textbf{Scope of the bounds.} The proposed bounds are not intrinsic properties of the dataset alone, but depend on the chosen pseudo-metric on the input space, the evaluation metric on the target space, and the finite-sample feasible set construction based on the thresholds from input and target noise. In the experimental section, whenever prior information on input (and/or target) measurement uncertainty was available, the thresholds $\tau_{\mathcal X}$ and $\tau_{\mathcal Y}$ were set to conservative worst-case values implied by the corresponding noise models.  In this setting the framework isolates a model-independent component of prediction difficulty, but does not directly account for optimization difficulty, finite model capacity, or model misspecification. 

\subsection{Relation to previous work}\label{sec:relwork}

The proposed framework to tackle the three challenges in Section \ref{sec:challenges} is related to several existing lines of work, including optimal recovery, approximation theory, statistical learning theory, and modern deep learning. 

\noindent\textbf{Optimal recovery and worst-case approximation.}
Classical approximation-theoretic frameworks quantify best achievable worst-case error in settings related to \eqref{eq:sampling10}. For noiseless linear problems, examples include Gelfand widths \cite{pinkus2012n} and best \(k\)-term approximation \cite{CoDaDe-08}. Worst-case bounds for set-valued recovery maps were established on Banach spaces in \cite{arestov1986optimal} and extended to metric spaces in \cite{magaril1991optimal}. For noisy problems, \cite{binev2022optimal} introduced optimal learning and upper bounds based on the Chebyshev radius, while \cite{cohen2022optimal} studied the optimal accuracy of nonlinear approximation methods; see also the radius of information in \cite{traub1983information} and the survey \cite{micchelli1977survey}. These works show that non-uniqueness can impose unavoidable error. Our approach differs in using metric and Hausdorff distances on feasible sets and optimal (set-valued) regressors and classifiers, without requiring a nested family of approximate prediction sets, and in providing a computable finite-dataset perspective relevant to modern machine learning.

\noindent\textbf{Statistical learning theory and model-dependent bounds.}
Another line of work studies regressor and classifier error through sampling assumptions, hypothesis-class complexity, or specific model classes, including neural-network architectural restrictions. For example, \cite{zhang2009some} derives sharp bounds for parameter estimation in linear regression under specific assumptions, and \cite{qi2020analyzing} gives upper bounds for the mean absolute error of neural-network regressors under architectural assumptions. Vapnik-Chervonenkis (VC) theory, see \cite{vapnik2006estimation,vapnik1999overview}, and related measures such as Rademacher complexity, see \cite{bartlett2002rademacher}, provide generalization bounds for restricted model classes, while \cite{cherkassky1999model} show the usefulness of VC-type bounds for model selection in small i.i.d. datasets and linear regression. However, such bounds are typically not directly computable for arbitrary finite datasets and broad model classes, and they often rely on assumptions, such as i.i.d. sampling, that may fail in practice or be difficult to verify \cite{arnold1979linear}.

\noindent\textbf{Modern deep learning and the limits of classical explanations.}
Neural networks are universal approximators, see \cite{hornik1989multilayer,maiorov1999lower,bach2017breaking,beck2022full,petersen2018optimal,yarotsky2017error}, but, as emphasized in \cite{adcock2021gap}, approximation-theoretic expressivity does not by itself provide guarantees on trainability or practical error and accuracy. \cite{zhang2016understanding,zhang2021understanding} show that deep networks can fit datasets with random labels, and argue that classical notions such as VC dimension, Rademacher complexity, and uniform stability do not explain the generalization behavior of state-of-the-art neural networks. Separately, \cite{northcutt2021pervasive} identified substantial label errors in widely used datasets, such as CIFAR-10, highlighting that both label uncertainty and annotation errors can limit measurable test accuracy. In some applications, deep learning is not even the best empirical choice: for instance, tree-based methods can outperform deep neural networks on tabular data \cite{grinsztajn2022tree}. These observations motivate the search for accuracy limits that depend on the task and dataset rather than on the expressive power of a specific model family. The perspective of this paper is closest to geometric and recovery-based approaches that quantify ambiguity in the input-target relation. It is complementary to information-theoretic minimax lower bounds, such as Fano and Le Cam methods \cite{tsybakov2003introduction,le2012asymptotic,fano1961transmission}, and to bias-variance decompositions of prediction risk \cite{geman1992neural}. 

\noindent\textbf{Bayes risk, irreducible error, and Bayes-error estimation.}
In statistical decision theory, the model-independent benchmark under distribution $\mu$ for a loss $\ell$ is the Bayes risk $\inf_{\phi} \mathbb E_{\mu}[\ell(Y,\phi(X))]$, where the infimum is over measurable decision rules $\phi$, see
\cite{berger1985statistical} and \cite{murphy2022probabilistic}, which could be either regressors or classifiers. Under Euclidean squared loss, the Bayes rule is the conditional mean and the irreducible term is the expected conditional variance. \cite{cover1967nearest} showed that the asymptotic error of a nearest-neighbor classifier is bounded above by twice the Bayes error; in their conditional sampling formulation, this error is characterized by pairwise disagreement between conditionally independent labels, which is the distributional analogue average kernel size in our settin. The problem of estimating this irreducible squared-loss term, often called the residual variance, has also been studied in nonparametric regression under i.i.d. sampling assumptions; see, for example, \cite{DevroyeGyorfiLugosiWalk2018}, who analyze a nearest-neighbor estimator together with asymptotic normality and concentration results.
Our distributional analogue average kernel size gives an equivalent pairwise representation, as its square is exactly twice the Bayes squared risk. Recent work develops statistical estimators of Bayes error and optimal classification error and accuracy, including multiclass and computationally efficient approaches, see \cite{chen2023evaluating,jeong2023demystifying,chen2025efficient,ushio2026practical}. These methods estimate distribution-level limits under probabilistic sampling assumptions, such as i.i.d. sampling. Related empirical studies refine the classical bias-variance picture in overparameterized regimes; see, for example, \cite{neal2018modern,yang2020rethinking}.  In contrast, our bounds are deterministic and finite-sample, apply to the realized empirical loss on the constructed dataset from Assumption \ref{as:feassetd}, extend to metric or pseudo-metric target spaces, and do not require i.i.d. sampling.

\noindent\textbf{Nonparametric estimation of conditional structure.}
Estimating conditional means, conditional variances, and Bayes risk from finite data is a central problem in nonparametric statistics, see 
\cite{stone1977consistent,gyorfi2002distribution}. Such estimators typically rely on i.i.d. sampling together with smoothness assumptions. Here the bounds are computed from deterministic and geometric finite feasible sets with a chosen pseudo-metric and threshold rule. It should be distinguished from an estimator based on genuine conditionally i.i.d. replicates. The convergence theorem connects the deterministic bounds to such an ideal sampling design.

\noindent\textbf{Stability and generalization.}
Classical algorithmic-stability results control expected generalization error by studying how a learning rule changes when observations in the training sample are replaced  \cite{bousquet2002stability}. The accuracy-stability result concerns a different notion: the metric regularity of the fitted regressors and classifiers as a function of their input, which is a deterministic geometric obstruction. If nearby inputs are paired with well-separated targets, a regressor or classifier that fits both targets accurately must have a large local Lipschitz constant, independently of how the regressor or classifier was trained.

\noindent\textbf{Dimensionality reduction and geometric preservation.}
Results such as the Johnson-Lindenstrauss lemma \cite{johnson1984extensions} show that low-dimensional embeddings can approximately preserve pairwise geometry, but this does not by itself quantify achievable prediction accuracy or ambiguity in the input-target relation.

\noindent\textbf{Positioning of the present work.}
We use a feasible-set geometry to derive computable, deterministic, non-asymptotic lower and upper bounds on the best achievable empirical error on a realized, finite, constructed dataset. The bounds are model-class independent and apply to upper hemicontinuous compact-valued regressors and classifiers, a class containing single-valued, continuous regressors and classifiers. The input space has a pseudo-metric and the target space a metric, so the construction is not restricted to Euclidean regression. The provided distribution dependent results show that, they relate conditional target ambiguity to Bayes risk and to generalization gaps under probabilistic assumptions.

\subsection{Outlook and impact}

An important practical feature of the proposed framework is that the bounds on empirical error on the constructed dataset \(\mathcal{D}_{M'}\) from Assumption \ref{as:feassetd} are computable before model optimization and training. As the bounds depend on the dataset, the chosen metric structure, and the finite samples used in the analysis, they provide task-dependent information prior to model selection and optimization. The framework can therefore help answer questions such as: \textit{Is the available data sufficiently informative for the task under the chosen evaluation metric?} and \textit{How much improvement remains possible before the observed empirical error reaches the model-independent floor induced by the chosen input space pseudo-metric and finite feasible sets?} The bounds quantify this remaining room for improvement in empirical error, but they do not imply that additional tuning, a different architecture, or a particular optimization method will
attain it. Although the proof of Theorem~\ref{thm:lowerboundapprx} is constructive and provides an explicit variational formula for the regressor obtaining the best achievable empirical error, this regressor is computationally expensive as it requires solving an optimization problem for each input. The analogous formula and conclusion apply to classification.

\noindent This makes the bounds potentially useful for dataset assessment, model selection, and training diagnostics. More broadly, they provide a principled way to distinguish limitations imposed by the data from limitations imposed by a specific learning method.

\subsection{Outline}
The remainder of the paper is organized into three logically distinct parts. First, we develop deterministic finite-sample bounds on empirical loss and accuracy for a constructed dataset derived from feasible-set geometry. Second, under additional probabilistic assumptions, we relate the distributional analogue of the average kernel size to Bayes risk and generalization gaps. Here Section~\ref{sec:main} develops the framework and assumptions and presents the main theoretical results, including the algorithm used to compute the accuracy bounds. Third, we provide experiments illustrating the behavior of the deterministic construction across several datasets and choices of evaluation metric. Here Section~\ref{sec:applications} presents the experimental results. Proofs are deferred to the end of the manuscript, while additional material for reproducibility, including implementation details and descriptions of the datasets considered, is provided in the Appendix. Our code is publicly available at this \href{https://github.com/nm19000/AccuracyBoundsRegression}{GitHub repository}.

\section{Main results}\label{sec:main}
In the following subsections, we first define a minimal framework for regression and classification problems that is required to present the corresponding main theoretical results. Then, we present the main deterministic and finite-sample results, hereafter we introduce additional probabilistic assumptions required to present the distribution level results.

\subsection{Problem formulation}\label{sec:preliminaries}

We model a regression or classification problem by a dataset of tuples
\[
\mathcal{D}_M \subset \mathcal{M}_1 \times \mathcal{M}_2 \subset \mathcal{Y} \times \mathcal{X},
\]
for some \(M \in \mathbb{N}\), together with the prediction task \eqref{eq:sampling10}. The ambient input space \(\mathcal{X}\) is equipped with a pseudo-metric \(d_{\mathcal X}\), and the ambient target space \(\mathcal{Y}\) is equipped with a metric \(d_{\mathcal Y}\). For image-based regression and classification tasks in Section \ref{sec:applications} we chose a metric based on wavelet transforms concatenated with the $\ell_2$ norm. We endow \(\mathcal{X}\) and \(\mathcal{Y}\) with the corresponding topologies generated by the open balls
\[
B_{d_{\mathcal X}}(x,r)=\{z\in\mathcal X:d_{\mathcal X}(x,z)<r\},
\qquad
B_{d_{\mathcal Y}}(y,r)=\{z\in\mathcal Y:d_{\mathcal Y}(y,z)<r\},
\]
for \(r>0\). We equip $\mathcal{M}_2$ with the induced subspace topology. In addition, we assume the following.

\begin{assumption}[Heine-Borel property]\label{as:heineBy}
The metric space \((\mathcal{Y},d_{\mathcal{Y}})\) satisfies the Heine-Borel property, that is, every closed and bounded subset of \(\mathcal{Y}\) is compact.
\end{assumption}

\noindent Throughout the paper, we denote by \(\pi_1:\mathcal{M}_1\times\mathcal{M}_2\to\mathcal{M}_1\) the projection \((y,x)\mapsto y\) and by \(\pi_2:\mathcal{M}_1\times\mathcal{M}_2\to\mathcal{M}_2\) the projection \((y,x)\mapsto x\). The disjoint union of sets \(A,B\subset\mathcal X\) is denoted by \(A\cupdot B\). A regressor or classifier is modeled by a set-valued map
\begin{align}\label{eq:decodermethod2}
\phi:\mathcal{M}_2\rightrightarrows\mathcal{Y},
\qquad
x\mapsto \phi(x)=\hat y.
\end{align}
For \(S\subset\mathcal{M}_2\), we define the image of \(S\) under \(\phi\) by
\[
\phi(S)=\bigcup_{x\in S}\phi(x).
\]
To accommodate set-valued regressors and classifiers, we extend the metric \(d_{\mathcal Y}\) by the Hausdorff distance. For bounded sets \(S_1,S_2\subset\mathcal Y\), define
\[
d_{\mathcal Y}^H(S_1,S_2)
=
\max\Bigl\{
\sup_{z\in S_1}\inf_{y\in S_2} d_{\mathcal Y}(z,y),
\;
\sup_{z\in S_2}\inf_{y\in S_1} d_{\mathcal Y}(z,y)
\Bigr\},
\]
which is well-defined for nonempty compact sets. Considering set-valued regressors and classifiers allows us to include stochastic models, such as diffusion-based generative models. For example, a finite collection of predictions \(\{y_\ell\}_{\ell=1}^L\) produced for a given input \(x\), yields a prediction set $\phi(x)=\{y_\ell\}_{\ell=1}^L$. For a dataset $\mathcal D_M=\{(y_m,x_m)\}_{m=1}^M$ and \(p\in[1,\infty)\), the empirical regression error of \(\phi\) is defined by \eqref{eq:empiricalerror}, which depends only on the sampled pairs in \(\mathcal D_M\).

\noindent For classification problems with \(C\in\mathbb N\) classes, we use the normalized Hamming distance and restrict the target ambient space $\mathcal Y$ as follows.

\begin{assumption}[Classification decisions and loss]
\label{as:convclassextra}
For classification, the target and decision space is
$\mathcal Y_C=\{e_1,\ldots,e_C\}$, equipped with its power-set sigma-algebra and the
normalized Hamming metric
\[
d_H(e_c,e_{c'})=\mathbf 1_{\{c\ne c'\}}.
\]
If a model outputs a score or probability vector in $\Delta^{C-1}$, it is converted to a one-hot decision by a fixed Borel-measurable tie-breaking $\arg\max$ rule.
\end{assumption}

\noindent Given a classifier \(\phi:\mathcal M_2\rightrightarrows\mathcal Y\), the empirical accuracy is defined by \eqref{eq:empiricalaccuracy} with \(d_{\mathcal Y}=d_H\).

\begin{remark}[Loss convention for set-valued predictors]
\label{rem:setvaluedloss}
Our loss uses the Hausdorff distance defined with the supremum over the prediction set and therefore measures average prediction error over pointwise worst-case distances. This convention is natural when a set-valued prediction is used in a downstream decision problem requiring error and accuracy to be controlled uniformly over the outcomes regarded as possible, as in robust or minimax decision theory \cite{gilboa1989maxmin,chamberlain2020robust}. Here, enlarging the prediction set cannot reduce the loss, and only provide benefits in downstream decision problems. This differs from coverage-oriented conventions, such as those used in conformal prediction \cite{angelopoulos2023conformal}, where a prediction set is rewarded for containing the realized target. Under such a convention, sufficiently large prediction sets could attain zero loss, so the lower bounds developed here would not directly apply.
\end{remark}

\noindent We quantify the stability of a regressor or classifier \(\phi:\mathcal M_2\rightrightarrows\mathcal Y\) by local and global Lipschitz constants. For \(\epsilon>0\) and \(x\in\mathcal M_2\), the local \(\epsilon\)-Lipschitz constant is defined by
\[
L^\epsilon(\phi,x)
=
\sup_{\substack{z\in\mathcal M_2\\ 0<d_{\mathcal X}(z,x)\le \epsilon}}
\frac{d_{\mathcal Y}^H(\phi(x),\phi(z))}{d_{\mathcal X}(x,z)},
\]
where in the classification setting \(d_{\mathcal Y}=d_H\). The global Lipschitz constant is defined by
\[
L(\phi)
=
\sup_{\substack{x,z\in\mathcal M_2\\ 0<d_{\mathcal X}(x,z)}}
\frac{d_{\mathcal Y}^H(\phi(x),\phi(z))}{d_{\mathcal X}(x,z)}.
\]
Related notions of Lipschitz continuity for set-valued maps may be found in \cite{li1994sharp,aubin1990set}. 

\noindent Following \cite[Definition 17.2]{guide2006infinite} we define upper hemicontinuity of a set-valued regressor or classifier $\phi : \mathcal{M}_2 \rightrightarrows \mathcal{Y}$. For this we define the \textit{upper inverse} $\phi^{u}$ of a subset $A \subset \mathcal{Y}$ by
    \[
        \phi^{u}(A) = \{ x \in \mathcal{M}_2 : \phi(x) \subset A \}.
    \]
Then, $\phi : \mathcal{M}_2 \rightrightarrows \mathcal{Y}$ is \emph{upper hemicontinuous} at \(x\) if for every neighborhood \(U\) of \(\phi(x)\), there is a neighborhood \(V\) of \(x\) such that \(z \in V\) implies \(\phi(z) \subset U\) or, equivalently, if \(\phi^u(U)\) is a neighborhood of \(x\) in \(\mathcal{X}\). As with functions, we say \(\phi\) is upper hemicontinuous on \(\mathcal{M}_2\), if it is upper hemicontinuous at every point of \(\mathcal{M}_2\). For more context, see Section \ref{sec:proofs}.

\begin{remark}[Level of generality] We deliberately adopt this level generality to accommodate practical data representations. The distance \(d_{\mathcal X}\) may be a pseudo-metric because distinct inputs can become indistinguishable after preprocessing or finite-precision representations. Multisets are used because multiplicities of admissible targets affect the feasible-set construction and empirical averages. The Heine-Borel assumption on \((\mathcal Y,d_{\mathcal Y})\) is the compactness condition needed for existence of minimizers; here finite-dimensional normed spaces are a standard special case.
\end{remark}

\subsection{Dataset-level stability and accuracy bounds}\label{sec:mainthm}

Motivated by the accuracy-stability trade-off for single-valued decoders solving inverse problems in \cite{gottschling2025troublesome}, we begin with an accuracy-stability trade-off showing that, when distinct targets correspond to nearby or identical inputs, a highly accurate regressor or classifier must become locally unstable. This observation motivates the lower bounds developed below.

\begin{proposition}[Accuracy--stability trade-off]\label{prop:instabilityaccuracy}
Let \(\eta,\tau_{\mathcal X}>0\), let \(M\in\mathbb N\), let \(\mathcal D_M \subset \mathcal M_1\times \mathcal M_2\) be a dataset, and let \(\phi:\mathcal M_2\rightrightarrows \mathcal Y\) be an upper hemicontinuous regressor or classifier with nonempty compact values and let $d_{\mathcal{X}}$ be a metric. Suppose there exist \((y,x),(y',x'),(y'',x'')\in \mathcal D_M\) such that
\begin{align}\label{eq:assinsta}
d_{\mathcal Y}^H(\phi(x),\{y\})<\eta,
\qquad
d_{\mathcal Y}^H(\phi(x'),\{y''\})<\eta,
\qquad
0<d_{\mathcal X}(x,x')< \tau_{\mathcal X}.
\end{align}
Then, there exists an open ball \(\mathcal B\subset \mathcal{M}_2\) centered at \(x\) such that, for every \(\epsilon>0\) with \(\tau_{\mathcal X}\le \epsilon\), the local \(\epsilon\)-Lipschitz constant satisfies
\[
L^\epsilon(\phi,\tilde x)
\ge
\frac{1}{\tau_{\mathcal X}}
\bigl(d_{\mathcal Y}(y,y'')-2\eta\bigr)
\qquad
\text{for all } \tilde x\in \mathcal B.
\]
\end{proposition}

\noindent While the proposition is, in essence, a geometric statement about an upper hemicontinuous \(\phi\), we formulate it using dataset points to situate it in the data-dependent framework used throughout the paper. The use of three dataset pairs in \eqref{eq:assinsta} is deliberate. The dataset contains $(y,x),(y',x')\in\mathcal D_M$ and, $(y'',x'')\in\mathcal D_M$, but the second accuracy condition evaluates the prediction at \(x'\) against the target \(y''\), rather than against the target \(y'\) paired with \(x'\). Thus, the proposition does not require \(\phi(x')\) to predict the target observed at \(x'\). It only requires that the prediction at the nearby input \(x'\) be close to some target \(y''\) that occurs in the dataset, possibly paired with a different input \(x''\). The more restrictive paired-target assumption is recovered by imposing \(y''=y'\). Since the proposition allows \(y''\ne y'\), it applies under a strictly weaker assumption and is therefore more general.

\noindent Under the assumptions of Proposition~\ref{prop:instabilityaccuracy}, the global Lipschitz constant satisfies
\[
L(\phi)
\ge
\sup_{\tilde x\in \mathcal M_2} L^\epsilon(\phi,\tilde x)
\ge
\frac{1}{\tau_{\mathcal X}}
\bigl(d_{\mathcal Y}(y,y'')-2\eta\bigr).
\]
Hence, if \(d_{\mathcal Y}(y,y'')-2\eta\ge C>0\), then letting \(\tau_{\mathcal X}\to 0^+\) forces \(L(\phi)\to\infty\). If two distinct targets are feasible for inputs with a very small distance, then uniformly small pointwise error is incompatible with global stability. 

\noindent If two targets $y,\tilde y\in F_x$ share exactly the same input and both are predicted with a pointwise error at most $\eta$, then as
$d_{\mathcal Y}(y,\tilde y)\le d_{\mathcal Y}^H(\{y\},\phi(x))+d_{\mathcal Y}^H(\{\tilde y\},\phi(x))\le 2\eta,$ necessarily $\eta\ge \frac12d_{\mathcal Y}(y,\tilde y)$. In worst-case form, this gives
\[
\eta \ge \frac{1}{2}\operatorname{diam}(F_x),
\qquad
\operatorname{diam}(F_x):=
\sup_{y,\tilde y\in F_x} d_{\mathcal Y}(y,\tilde y).
\]
While this yields a worst-case obstruction, in statistical learning it is often more informative to quantify the \emph{typical} ambiguity of the prediction task. This leads to the average kernel size and the computable empirical bounds below. The average kernel size is computed with the help of Algorithm~\ref{alg:feasset} which constructs finite-sample feasible sets from Definition \ref{def:feassetd}. It takes as input a collection of anchor points in \(\mathcal M_2\), a dataset \(\mathcal D_M \subset \mathcal M_1\times \mathcal M_2\) and threshold functions \(\tau_{\mathcal X}:\mathcal M_2\to (0,\infty)\) and \(\tau_{\mathcal Y}:\mathcal M_1\to (0,\infty)\), that satisfy the following assumption.

\begin{assumption}[Threshold compatibility with admissibility]\label{as:levelstauxy}
We assume that the threshold functions \(\tau_{\mathcal X}:\mathcal M_2\to (0,\infty)\) and \(\tau_{\mathcal Y}:\mathcal M_1\to (0,\infty)\) are such that for all candidate pairs $(y_k,x_k),(y_{k,n},x_{k,n}) \in \mathcal{D}_M$ whenever
\begin{align}\label{eq:leveltxty}
d_{\mathcal Y}(y_{k,n},y_k)\le \tau_{\mathcal Y}(y_k)
\quad\text{or}\quad
d_{\mathcal X}(x_{k,n},x_k)\le \tau_{\mathcal X}(x_k),
\end{align}
then
\[
(y_{k,n},x_k)\in \mathcal{R},
\]
where \(\mathcal{R}\) is the underlying admissible input-target multiset from Assumption~\ref{as:feassetd}.
\end{assumption}

\noindent Assumption~\ref{as:levelstauxy} is a compatibility condition between the threshold-based feasible-set rule \eqref{eq:leveltxty} and the underlying admissibility: it does not assume that \(\mathcal R\) is recovered from \((\tau_{\mathcal X},\tau_{\mathcal Y})\), but that the chosen thresholds are chosen such that reassignments remain in the multiset \(\mathcal R\). For a given sample, we generally cannot determine whether the observed input or target has been corrupted, nor which noise mechanism is responsible. Worst-case magnitudes from input-noise and target-/measurement-noise models motivate candidate choices for $\tau_{\mathcal X}$ and $\tau_{\mathcal Y}$; see Section\ref{sec:applications} for examples. Their use in Algorithm \ref{alg:feasset} is justified only when Assumption~\ref{as:levelstauxy} also holds, i.e. when the threshold-based reassignments are compatible with the underlying admissible multiset $\mathcal R$. The assumption requires that, for observed pairs already contained in $\mathcal{D}_M \subset \mathcal R$, a distance-based replacement of the input or target according to the corresponding threshold also yields an element of $\mathcal R$.\footnote{If the sample-level noise corruption is known, we instead denoise sample-wise and choose $\tau_{\mathcal X}$ and $\tau_{\mathcal Y}$ according to any remaining worst-case noise.} The disjunctive rule in Assumption~\ref{as:levelstauxy} and Algorithm~\ref{alg:feasset} reflects uncertainty about the source of corruption: a candidate pair is admitted whenever it is compatible by either the input-threshold or the target-threshold criterion. Thus, the ``or'' does not mean that proximity in input and target space is equivalent; it reflects that the source of corruption is unknown at the sample level. Unless the noise is heteroskedastic, see \cite{gawlikowski2023survey}, the threshold functions \(\tau_{\mathcal X}\) and \(\tau_{\mathcal Y}\) are constant. If no reliable prior information is available for choosing \(\tau_{\mathcal X}\), another choice is to calibrate it using Proposition~\ref{prop:instabilityaccuracy}.

\begin{algorithm}
\caption{Approximation of feasible sets}\label{alg:feasset}
\begin{algorithmic}[1]
\Require \(K,N^f_{\max}\in\mathbb N\), dataset \(\mathcal D_M\)
\Require threshold functions \(\tau_{\mathcal X}(\cdot)>0\), \(\tau_{\mathcal Y}(\cdot)>0\)
\State \(\mathcal D_{M'}\gets \emptyset\)
\For{\(k=1,\dots,K\)}
    \State choose an anchor \((y_k,x_k)\in \mathcal D_M\)
     \Comment{Sample without replacement.}
    \State \(F_{x_k}^{N(k)}\gets \emptyset\), \(N(k)\gets 0\)
    \While{\(N(k)<N^f_{\max}\)}
    \Comment{Sample without replacement; until \(N(k)\geq N^f_{\max}\) or $\mathcal D_M$ is empty.}
        \State sample \((y_{k,n},x_{k,n})\in \mathcal D_M\)
        \If{\(d_{\mathcal Y}(y_{k,n},y_k)\le \tau_{\mathcal Y}(y_k)\) or \(d_{\mathcal X}(x_{k,n},x_k)\le \tau_{\mathcal X}(x_k)\)}
            \State \(F_{x_k}^{N(k)}\gets F_{x_k}^{N(k)}\cupdot\{y_{k,n}\}\) \Comment{"$\cupdot$" denotes the disjoint union.}
            \State \(\mathcal D_{M'}\gets \mathcal D_{M'}\cupdot\{(y_{k,n},x_k)\}\)
            \State \(N(k)\gets N(k)+1\)
        \EndIf
    \EndWhile
\EndFor
\State \Return \(\{F_{x_k}^{N(k)}\}_{k=1}^K\), \(\{N(k)\}_{k=1}^K\), and \(\mathcal D_{M'}\)
\end{algorithmic}
\end{algorithm}

\begin{remark}[Scope of Assumption~\ref{as:levelstauxy}]
\noindent Assumption~\ref{as:levelstauxy} ensures that the pairs introduced by Algorithm~\ref{alg:feasset} are admissible in $\mathcal R$, but it does not imply any probabilistic statement about their multiplicities. Membership in $F_x$ does not imply that the constructed frequencies reproduce conditional probabilities. Such a claim would require additional probabilistic assumptions. Here, the condition \eqref{eq:leveltxty} in Assumption~\ref{as:levelstauxy} should therefore be understood as generating an admissible multiset of alternative targets.
\end{remark}

\noindent Algorithm~\ref{alg:feasset} returns finite multisets \(\{F_{x_k}^{N(k)}\}_{k=1}^K\), the finite feasible sets, together with their cardinalities \(\{N(k)\}_{k=1}^K\) and a constructed dataset 
\[
\mathcal D_{M'}
=
\{(y_{k,n},x_k):1\le k\le K,\ 1\le n\le N(k)\},
\qquad
M'=\sum_{k=1}^K N(k).
\] 
If threshold functions \(\tau_{\mathcal X}\) and \(\tau_{\mathcal Y}\) satisfy Assumption \ref{as:levelstauxy} and additionally we chose the number of anchor samples such that $K =M$ and if each anchor's own pair is in its feasible set and at least one set must be non-degenerate, then, as the anchors are sampled without replacement, the constructed dataset $\mathcal D_{M'}$ from Algorithm~\ref{alg:feasset} satisfies Assumption \ref{as:feassetd} and makes this additional assumption obsolete. The sampling procedure in the algorithm is intentionally left open and may be implemented, for example, by brute-force search, Monte Carlo methods, or nearest-neighbor queries, depending on the application. In our minimal implementation and examples in Section \ref{sec:applications}, we adopt queries based on nearest-neighbor distance matrices.

\noindent With the outputs from Algorithm~\ref{alg:feasset} we define the quantity that is central to the lower and upper bounds on empirical accuracy and error.

\begin{definition}[Average kernel size]
Let \(p\in[1,\infty)\), let \(\mathcal D_M\) be a dataset, and fix \(K,N^f_{\max}\in\mathbb N\). Apply Algorithm~\ref{alg:feasset} to obtain finite feasible sets
\[
\left\{F_{x_k}^{N(k)}\right\}_{k=1}^K
=
\left\{\{y_{k,n}\}_{n=1}^{N(k)}\right\}_{k=1}^K
\]
and let \(M'=\sum_{k=1}^K N(k)\). The \emph{average kernel size} is defined by
\begin{align}\label{eq:lowerboundogboiz}
\operatorname{Kersize}(\mathcal D_{M'},p)_K
=
\left(
\frac{1}{M'}
\sum_{k=1}^K
\frac{2\,\mathds{1}_{\{N(k)>1\}}}{N(k)-1}
\sum_{1\le n<n'\le N(k)}
d_{\mathcal Y}(y_{k,n},y_{k,n'})^p
\right)^{1/p}.
\end{align}
\end{definition}

\noindent The diagonal terms \(d_{\mathcal Y}(y_{k,n},y_{k,n})=0\) are omitted, and the normalization averages pairwise distances within each feasible set before averaging across all allocated samples. Thus, \(\operatorname{Kersize}(\mathcal D_{M'},p)_K\) measures the typical spread of targets associated with locally indistinguishable inputs and is a sample-weighted within-group average pairwise distance. The average kernel size provides the following finite sample, deterministic lower and upper bounds on the empirical error of regressors.

\begin{theorem}[Deterministic accuracy bounds and upper-hemicontinuous attainment]
\label{thm:lowerboundapprx}
Let \(p\in[1,\infty)\), assume Assumptions~\ref{as:feassetd} and~\ref{as:heineBy}, and let $\mathcal D_{M'}=\{(y_{k,n},x_k):1\le k\le K,\ 1\le n\le N(k)\}$, with $M'=\sum_{k=1}^K N(k)$, be the constructed dataset returned by Algorithm~\ref{alg:feasset} and where \(N(k)\ge1\) for every \(k \in \{1,\ldots, K\}\). Then, every regressor \(\phi:\mathcal M_2\rightrightarrows \mathcal Y\) with nonempty values satisfies
\[
\frac12\operatorname{Kersize}(\mathcal D_{M'},p)_K
\le
\mathcal L(\mathcal D_{M'},\phi,p).
\]
If additionally $d_{\mathcal X}(x_k,x_{k'})>0$ for all $k\ne k'$, then
\[
\frac12\operatorname{Kersize}(\mathcal D_{M'},p)_K
\le
\inf_{\phi_{\mathrm{uhc}}}
\mathcal L(\mathcal D_{M'},\phi_{\mathrm{uhc}},p)
=
\inf_{\phi}
\mathcal L(\mathcal D_{M'},\phi,p)
\le
\operatorname{Kersize}(\mathcal D_{M'},p)_K,
\]
where the first infimum is taken over all upper hemicontinuous regressors with nonempty, compact values and the second over all maps with nonempty bounded values. The infimum is attained by an upper hemicontinuous regressor with nonempty compact values. The attaining regressor can be chosen singleton-valued on a neighborhood of every anchor $x_k$.
\end{theorem}

\noindent Theorem~\ref{thm:lowerboundapprx} shows that enlarging the admissible model class to include discontinuous set-valued regressors does not improve the optimal empirical error on the constructed dataset. This supports the intuition that the average kernel size quantifies the residual variation in the target given a fixed pseudo-metric on the input space. If this quantity is nonzero, then part of the prediction error is intrinsic to the task and cannot be removed by changing the model or learning architecture alone. Although Theorem~\ref{thm:lowerboundapprx} holds for all sample sizes $M,K \in \mathbb{N}$ and all feasible set cardinalities $N(k) \in \mathbb{N}$ for all $k \in \{1, \ldots, K\}$, in the numerical experiments in Section~\ref{sec:applications} to ensure the average kernel size is representative of the whole dataset with size $M \in \mathbb{N}$, if possible we chose $K=M$. The feasible set cardinalities $N(k) \in \mathbb{N}$ are determined by the thresholds, which we chose to satisfy Assumption \ref{as:levelstauxy}.

\noindent Under additional assumptions on the target ambient space the minimizer admits a continuous single-valued extension via a standard bump function construction. 

\begin{corollary}[Continuous attainment under linear extension assumptions]
Assume, in addition to the assumptions made in Theorem~\ref{thm:lowerboundapprx} including that $d_{\mathcal X}(x_k,x_{k'})>0$ for all $k\ne k'$, that $\mathcal Y$ is a convex subset of a normed linear space and that $d_{\mathcal Y}$ is a norm. Then, the finite assignment $x_k\mapsto z_k$ admits a continuous single-valued extension $\theta:\mathcal M_2\to\mathcal Y$, and $\theta$ attains the unrestricted empirical minimum on $\mathcal M_2$.
\end{corollary}

\noindent In the scalar-valued case, where $\mathcal Y\subset\mathbb R$, the minimizer admits a Lipschitz-continuous single-valued extension via the extension theorem from \cite{mcshane1934extension}. For classification, we define the average kernel size by using the Hamming distance in \eqref{eq:lowerboundogboiz} with \(p=1\). Since accuracy is an affine transformation of a metric rather than a metric itself, the classification statement is formulated as an upper bound on achievable accuracy and we obtain the upper bound in Corollary \ref{thm:lowerboundapprxclass} under the additional Assumption \ref{as:convclassextra}, as an immediate consequence of Theorem~\ref{thm:lowerboundapprx}. 

\begin{corollary}[Accuracy bounds for upper hemicontinuous classifiers]
\label{thm:lowerboundapprxclass}
Assume Assumptions~\ref{as:feassetd} and~\ref{as:heineBy} and \ref{as:convclassextra}, with $p=1$, $d_{\mathcal Y}=d_H$, and let $\mathcal D_{M'}=\{(y_{k,n},x_k):1\le k\le K,\ 1\le n\le N(k)\}$ with $M'=\sum_{k=1}^K N(k)$ be the dataset returned by Algorithm~\ref{alg:feasset} and where \(N(k)\ge1\) for every \(k \in \{1, \ldots, K\}\). Then, every classifier \(\phi:\mathcal M_2\rightrightarrows \mathcal Y\) with nonempty values satisfies
\[
\operatorname{Acc}(\mathcal D_{M'},\phi)
\le
1-\frac12\operatorname{Kersize}(\mathcal D_{M'},1)_K.
\]
If additionally $d_{\mathcal X}(x_k,x_{k'})>0$ for all $k\ne k'$, then
\[
1-\operatorname{Kersize}(\mathcal D_{M'},1)_K
\le
\sup_{\phi_{\mathrm{uhc}}}
\operatorname{Acc}(\mathcal D_{M'},\phi_{\mathrm{uhc}})
=
\sup_{\phi}
\operatorname{Acc}(\mathcal D_{M'},\phi)
\le
1-\frac12\operatorname{Kersize}(\mathcal D_{M'},1)_K.
\]
The common supremum is attained by an upper hemicontinuous classifier with nonempty
finite values.
\end{corollary}

\begin{remark}[Classification target set extension]
Since no classifier attaining the best accuracy needs to be constructed for the upper bound, the upper-bound statement in Corollary~\ref{thm:lowerboundapprxclass} also holds when the target space in Assumption~\ref{as:convclassextra} is replaced by
\[
\mathcal Y=\Delta^{C-1}
=
\left\{
p\in\mathbb R^C:\;
p_i\ge 0,\ i=1,\dots,C,\;
\sum_{i=1}^C p_i=1
\right\},
\]
equipped with the metric $d_{\mathcal Y}(y,y'):=\frac12\|y-y'\|_1$.
\end{remark}

\noindent Theorem~\ref{thm:lowerboundapprx} and Corollary~\ref{thm:lowerboundapprxclass} provide lower, respectively upper, bounds to the error, or respectively accuracy, of any model evaluated on a dataset constructed with Algorithm~\ref{alg:feasset}. A natural question is, whether the error, or accuracy, on the constructed dataset are near the error or accuracy on the original dataset. In the following we provide a deterministic comparison bound of these two quantities where we utilize a graph-based argument. Let $E\subset\{1,\ldots,M\}^2$ denote the collection of reassignment edges generated by Algorithm~\ref{alg:feasset}. An edge \(e=(k,j)\in E\) indicates that the target \(y_j\) is reassigned to the anchor input \(x_k\). Thus, \(M'=|E|\), and the constructed dataset empirical loss can be written as an average over \(e\in E\). For \(m\in\{1,\ldots,M\}\), define the original pointwise loss by
\[
\ell_m(\phi)
:=
d_{\mathcal Y}^H(\{y_m\},\phi(x_m)).
\]
For each edge \(e=(k,j)\in E\), we associate a reference index \(r(e)\in\{1,\ldots,M\}\) and a corresponding reassignment error \(\gamma_e\). If \(e\) is generated by the input distance criterion, we set
\[
r(e)=j, \qquad\gamma_e := d_{\mathcal Y}^H(\phi(x_k),\phi(x_j)).
\]
If \(\phi\) is \(L\)-Lipschitz, then $\gamma_e\le L\,d_{\mathcal X}(x_k,x_j)$. If \(e\) is generated by the target distance criterion, we instead set
\[
r(e)=k, \qquad \gamma_e:=d_{\mathcal Y}(y_j,y_k).
\]
If both criteria apply, we choose the reference index corresponding to
the smaller of the two available upper bounds. With this choice, the reverse triangle inequality for the Hausdorff
metric gives, for every \(e=(k,j)\in E\),
\begin{align*}
\left|d_{\mathcal Y}^H(\{y_j\},\phi(x_k)) -\ell_{r(e)}(\phi)\right|\le \gamma_e.
\end{align*}
We define the induced reference weights by
\[
w_m:=\frac{1}{M'}\#\{e\in E:r(e)=m\},\qquad m=1,\ldots,M.
\]
Thus \(w_m\) is the fraction of constructed tuples whose reference
index is \(m\). $\sum_{m=1}^M w_m=1$ and if \(N(m)\) denotes the number of constructed tuples for which \(m\) is the reference index, then $w_m=\frac{N(m)}{M'}$. Finally, define
\[
\Gamma_p(\phi):=\left(\frac{1}{M'}\sum_{e\in E}\gamma_e^p\right)^{1/p},\qquad B_\phi:=\max_{1\le m\le M}\ell_m(\phi).
\]
The quantity \(\Gamma_p(\phi)\) measures the average geometric error introduced by the reassignments, while the weights \(w_m\) quantify the frequency with which the original observations are represented through the reference construction.
\begin{proposition}[Original versus constructed dataset empirical loss]\label{prop:originalenlarged} 
Assume Assumptions~\ref{as:feassetd} and~\ref{as:heineBy}, and let $\mathcal D_{M'}=\{(y_{k,n},x_k):1\le k\le K,\ 1\le n\le N(k)\}$ with $M'=\sum_{k=1}^K N(k)$ be the dataset returned by Algorithm~\ref{alg:feasset} and where \(N(k)\ge1\) for every \(k \in \{1, \ldots, K\}\). For every $p\in[1,\infty)$ and every regressor \(\phi:\mathcal M_2\rightrightarrows \mathcal Y\) with nonempty values,
\[
\left|\mathcal L(\mathcal D_{M'},\phi,p)-\mathcal L(\mathcal D_M,\phi,p)\right|\le
\Gamma_p(\phi)+B_\phi\left(\sum_{m=1}^M\left|w_m-\frac1M\right|\right)^{1/p}.
\]
\end{proposition}

\begin{remark}[Role of the thresholds]
The thresholds \(\tau_{\mathcal X}\) and \(\tau_{\mathcal Y}\) control the geometric term \(\Gamma_p(\phi)\) in Proposition~\ref{prop:originalenlarged}. By definition of Algorithm~\ref{alg:feasset} for an input edge \(e=(k,j)\), we have that $d_{\mathcal X}(x_k,x_j)\le \tau_{\mathcal X}(x_k)$ and, if \(\phi\) is \(L\)-Lipschitz, $\gamma_e=d_{\mathcal Y}^H(\phi(x_k),\phi(x_j))\le L\tau_{\mathcal X}(x_k)$. For a target edge, we have that $\gamma_e=d_{\mathcal Y}(y_j,y_k) \le \tau_{\mathcal Y}(y_k)$. Consequently, if $\tau_{\mathcal X}(x)\le \overline{\tau}_{\mathcal X}$ and $\tau_{\mathcal Y}(y)\le \overline{\tau}_{\mathcal Y}$, then $\Gamma_p(\phi)\le\max\left\{L\overline{\tau}_{\mathcal X},\overline{\tau}_{\mathcal Y}\right\}$. If additionally the reference weights are balanced; meaning that $w_m=\tfrac{N(m)}{M'} = \tfrac{1}{M}$ for all $m \in \{1, \ldots, M\}$, then Proposition~\ref{prop:originalenlarged} implies that
\[
\left|\mathcal L(\mathcal D_{M'},\phi,p)-\mathcal L(\mathcal D_M,\phi,p)
\right|\le\max\left\{L\overline{\tau}_{\mathcal X},\overline{\tau}_{\mathcal Y}\right\}.
\]
The thresholds $\tau_{\mathcal X}$ and $\tau_{\mathcal{Y}}$ therefore control the geometric error. Control of $\Gamma_p(\phi)$ by $\tau_{\mathcal X}$ and $\tau_{\mathcal Y}$ does not by itself imply a nontrivial transfer of lower bounds from $\mathcal D_{M'}$ to $\mathcal D_M$. For predominantly target-based reassignments, $\Gamma_p(\phi)$ may already absorb the ambiguity on $\mathcal D_{M'}$, so the induced accuracy bounds on $\mathcal D_M$ can be vacuous. Note that no independence assumption appears anywhere in the proposition. 
\end{remark}

\subsection{Non-Vanishing Generalization Gaps}

Up to this point, all statements have been deterministic and require only a dataset together with a metric structure. Extending the work on inverse problems in \cite{gottschling2023existence} to regression and classification problems, we introduce a probabilistic setting to interpret the average kernel size as a model-independent component of irreducible prediction error. This requires measurability assumptions and the existence of conditional laws. For simplicity we consider single-valued measurable maps in the following and leave extensions to set-valued measurable regressors and classifiers to future work.

\begin{assumption}[Standard Borel quotient setting]
\label{as:completeY}
Let $x\sim x'$ if and only if $d_{\mathcal X}(x,x')=0$, let
\[
\overline{\mathcal M}_2:=\mathcal M_2/{\sim},
\]
and let $q:\mathcal M_2\to\overline{\mathcal M}_2$ be the quotient map. We assume that $\overline{\mathcal M}_2$ and $\mathcal Y$ are standard Borel spaces. For example, it
suffices that they are Polish spaces equipped with their Borel sigma-algebras. We also
assume that $(y,y')\mapsto d_{\mathcal Y}(y,y')^p$ is Borel measurable.
\end{assumption}

\begin{assumption}[Data-generating law]
\label{as:measure}
Let $(\Omega,\mathcal F,\mathbb P)$ be a probability space. Equip $\mathcal M_2$
with the Borel sigma-algebra of its pseudo-metric topology, and let
$X:\Omega\to\mathcal M_2$ and $Y:\Omega\to\mathcal Y$ be measurable. The
quotient map $q$ is Borel measurable, so we may define the
quotient input random variable
\[
\overline X:=q(X).
\]
Let $\mu$ denote the joint law of $(Y,\overline X)$.
\end{assumption}
 
\noindent Under Assumption \ref{as:measure}, Theorem~\ref{thm:exitsregdist} implies that there exists a regular conditional law
\[
\mu_{\bar x}(A)
:=
\mathbb P(Y\in A\mid\overline X=\bar x),
\qquad
A\in\mathcal B(\mathcal Y),
\]
defined for $\mu_{\overline X}$-almost every $\bar x$. Now let $Y$ and $Y'$ be conditionally independent with common conditional law
$\mu_{\overline X}$ given $\overline X$. Similar to the deterministic analog of the average kernel size in~\eqref{eq:lowerboundogboiz}, we define the distributional analogue average kernel size by
\[
\operatorname{Kersize}(\mathcal D_\mu,p)
:=
\left(
\mathbb E_{\overline X}
\mathbb E_{Y,Y'\mid\overline X}
\bigl[d_{\mathcal Y}(Y,Y')^p\bigr]
\right)^{1/p}.
\]

\noindent For classification, we define the distributional analogue average kernel size analogously, replacing \(d_{\mathcal Y}\) by the Hamming distance \(d_H\) and taking \(p=1\).  In the Euclidean setting we can show that the distributional analogue average kernel size also quantifies the irreducible error, akin to the bias-variance decomposition introduced in \cite{geman1992neural}.

\begin{proposition}[Exact Euclidean squared-loss identity]
\label{prop:bayes-kernel-identity}
Assume Assumptions~\ref{as:completeY} and~\ref{as:measure}, $\mathcal Y=\mathbb R^d$, $d_{\mathcal Y}(y,y')=\|y-y'\|_2$, and
$\mathbb E\|Y\|_2^2<\infty$. Let $m(\overline X):=\mathbb E[Y\mid\overline X]$ denote the conditional mean. Then, every measurable $\overline \phi:\overline{\mathcal M}_2\to\mathbb R^d$ with finite squared
risk satisfies
\[
\mathbb E\|Y-\overline \phi(\overline X)\|_2^2
=
\frac12\operatorname{Kersize}(\mathcal D_\mu,2)^2
+
\mathbb E\|m(\overline X)-\overline \phi(\overline X)\|_2^2.
\]
Consequently,
\[
\inf_{\overline \phi}
\left(\mathbb E\|Y-\overline \phi(\overline X)\|_2^2\right)^{1/2}
=
\tfrac{1}{\sqrt2}\operatorname{Kersize}(\mathcal D_\mu,2),
\]
where the infimum is attained by the conditional mean $m$.
\end{proposition}

\noindent A large distributional analogue average kernel size therefore indicates that a substantial part of the prediction error is caused by aleatoric uncertainty or non-identifiability rather than by an inadequate model class. We define the generalized risk of a measurable regressor $\overline\phi:\overline{\mathcal M}_2\to\mathcal Y$ by
\[
\mathcal L(\mathcal D_\mu,\overline\phi,p)
:=
\Bigl(
\mathbb E_{(Y,\overline X)\sim\mu}\bigl[d_{\mathcal Y}(\overline\phi(\overline X),Y)^p\bigr]
\Bigr)^{1/p},
\]
and, analogously, the generalized accuracy of a measurable classifier by
\[
\operatorname{Acc}(\mathcal D_\mu,\overline\phi)
:=
1-\mathbb E_{(Y,\overline X)\sim \mu}\bigl[d_H(\overline\phi(\overline X),Y)\bigr].
\]

\noindent This probabilistic interpretation also yields lower bounds on generalization gaps whenever the training error is substantially smaller than the ambiguity encoded by the conditional law.

\begin{proposition}[Positive generalization gaps from conditional ambiguity]
\label{thm:gengapbd}
Assume Assumptions~\ref{as:completeY} and~\ref{as:measure} and let \(M\in\mathbb N\) and $\mathcal D_M=\{(y_m,x_m)\}_{m=1}^M \subset \mathcal{M}_1 \times \mathcal{M}_2$.

\noindent\textbf{Regression.}
Let $p\ge1$ and $\epsilon \in (0, \tfrac{1}{2}\operatorname{Kersize}(\mathcal D_\mu,p))$. Let $\overline\phi:\overline{\mathcal M}_2\to\mathcal Y$ be measurable and let $\phi=\overline\phi\circ q:\mathcal M_2\to\mathcal Y$ with  \(\mathcal L(\mathcal D_\mu,\overline\phi,p)<\infty\). If its empirical training loss satisfies
\[
\mathcal L(\mathcal D_M,\phi,p)\le\epsilon,
\]
then
\[
0<
\tfrac12\operatorname{Kersize}(\mathcal D_\mu,p)-\epsilon
\le
\mathcal L(\mathcal D_\mu,\overline\phi,p)
-
\mathcal L(\mathcal D_M,\phi,p).
\]
If $p=2$ and $\mathcal Y=\mathbb R^d$ with Euclidean loss, then the sharper exact
Bayes bound holds:
\[
0<
\tfrac{1}{\sqrt2}\operatorname{Kersize}(\mathcal D_\mu,2)-\epsilon
\le
\mathcal L(\mathcal D_\mu,\overline\phi,2)
-
\mathcal L(\mathcal D_M,\phi,2),
\]
provided $\operatorname{Kersize}(\mathcal D_\mu,2)>\sqrt2\,\epsilon$.

\noindent\textbf{Classification.}
Let the target space be the finite one-hot set from Assumption~\ref{as:convclassextra} and let
$\epsilon \in (0, \tfrac{1}{2}\operatorname{Kersize}(\mathcal D_\mu,1))$. If a measurable quotient classifier satisfies
\[
\operatorname{Acc}(\mathcal D_M,\phi)\ge1-\epsilon,
\]
then
\[
0<
\tfrac12\operatorname{Kersize}(\mathcal D_\mu,1)-\epsilon
\le
\operatorname{Acc}(\mathcal D_M,\phi)
-
\operatorname{Acc}(\mathcal D_\mu,\overline\phi).
\]
\end{proposition}

\noindent Under ideal sampling conditions the following theorem relates the distributional analogue average kernel size to a finite-sample estimator.

\begin{proposition}[Concentration of the ideal conditional-sampling estimator]
\label{thm:convavgalgnon}
Assume Assumptions~\ref{as:completeY} and~\ref{as:measure}. Let
$p\in[1,\infty)$ and assume
\[
v_{\max}:=\sup_{y,y'\in\mathcal Y}d_{\mathcal Y}(y,y')^p<\infty.
\]
For $k=1,\ldots,K$, sample $\overline X_k$ independently from
$\mu_{\overline X}$. Let \(N_K\ge 2\) and conditional  on all anchors
$(\overline X_1,\ldots,\overline X_K)$, sample an array
$Y_{k,n}$ for $n \in \{1, \ldots, N_K\}$ mutually independent from $\mu_{\overline X_k}$. Assume that the arrays $Y_k$ and $Y_{k'}$ are independent for every
$k\neq k'$, regardless of whether the corresponding anchors
$\overline X_k$ and $\overline X_{k'}$ coincide. Define
\[
\widehat\kappa_{K,N_K}^p
:=
\frac1K\sum_{k=1}^K
\frac{2}{N_K(N_K-1)}
\sum_{1\le n<n'\le N_K}
 d_{\mathcal Y}(Y_{k,n},Y_{k,n'})^p.
\]
Then, for every $\delta>0$,
\[
\mathbb P\left(
\left|\widehat\kappa_{K,N_K}^p
-
\operatorname{Kersize}(\mathcal D_\mu,p)^p\right|
\ge\delta
\right)
\le
2\exp\left(-\frac{2K\delta^2}{v_{\max}^2}\right).
\]
$\widehat\kappa_{K,N_K}^p$ converges in probability to
$\operatorname{Kersize}(\mathcal D_\mu,p)^p$ as $K\to\infty$, for any choice
$N_K\ge2$.
\end{proposition}

\begin{remark}[Ideal conditional sampling versus Algorithm~\ref{alg:feasset}]
The estimator in Proposition~\ref{thm:convavgalgnon} assumes an ideal two-stage design: anchor inputs are sampled from the quotient input law and, at every anchor, repeated targets are sampled conditionally independently from the exact law of $Y\mid\overline X=\bar x$. Algorithm~\ref{alg:feasset} does not require generating such samples, as it selects and is allowed to reuse observations from one finite dataset according to input- and target-distance thresholds; feasible sets can overlap and their elements can be dependent. The average kernel size returned by Algorithm~\ref{alg:feasset} is a deterministic, data-dependent geometric surrogate. Proposition~\ref{thm:convavgalgnon} should not be read as a consistency theorem for that surrogate, as such a result would require additional assumptions controlling neighborhood shrinkage, effective within-neighborhood sample sizes, continuity of the conditional law, sample reuse, and the weights induced by the selection rule.
\end{remark}

\section{Applications}
\label{sec:applications}

In the following we provide an outline of experiments and implementation, however, comprehensive details on the implementation and considered datasets are in the Appendix in Section \ref{sec:expdet}. We applied the framework to a range of tabular and image-regression tasks and image-classification tasks. For each experiment, we sampled finite feasible sets, constructed the dataset \(\mathcal D_{M'}\), computed the average kernel size, and used Theorem~\ref{thm:lowerboundapprx} and Corollary~\ref{thm:lowerboundapprxclass} to obtain model-independent bounds on the optimal empirical regression loss and classification error, respectively. The experiments are intended as illustrations of the deterministic construction. They address three empirical questions: $i)$ whether the lower bound is nontrivial across domains, $ii)$ how reported or measured model performance compares numerically with the resulting interval, and $iii)$ how sensitive the bound is to the threshold and parameters used for the feasible set construction. We compared the theoretical intervals with the error and accuracy of several regressors and classifiers evaluated on both the original datasets and, where the models are available, on constructed datasets. We used a synthetic regression example to study how the accuracy bounds interact with the accuracy-stability trade-off from Proposition \ref{prop:instabilityaccuracy}. 

\noindent Across all experiments, the threshold-selection rule was as follows: when an input-noise model or target/label measurement-noise model was available, we set $\tau_{\mathcal X}$ and $\tau_{\mathcal Y}$ equal to conservative worst-case noise magnitudes implied by that model. If such knowledge was not available, for the input threshold $\tau_{\mathcal{X}}$ we followed a practical empirical heuristic based on quantiles detailed in the Appendix in Section \ref{sec:feassetsamp} and assumed ideal, noiseless targets with \(\tau_{\mathcal Y}=0\). For \(\tau_{\mathcal Y}=0\) the target criterion in the "or"-condition Algorithm~\ref{alg:feasset} was not included. We used a wavelet-based metric with a conversion to float$32$ for all image inputs described in the Appendix in Section \ref{subsec:common_image_metric}; for images represented in float$64$, this induces a pseudo-metric. All details on the datasets, wavelet-based metric and feasible set sampling are in the Appendix in Section \ref{sec:expdet}.

\subsection{Theoretical assumptions and guarantees versus experimental procedure}

\noindent In all experiments, if an input-noise model or target/label measurement-noise model was available, then $\tau_{\mathcal X}$ and $\tau_{\mathcal Y}$ were set to conservative worst-case values implied by that model; this first choice satisfies a part of Assumption \ref{as:levelstauxy} without taking into account the correct frequencies in the underlying admissible multiset. 

\noindent For smaller datasets, ($M \approx 1k$), we computed the bounds on the full dataset and did not compute these separately on the train and test splits. For large and high-dimensional datasets, such as Tropical Cyclone and RESISC45, with a full dataset size in the range $M \approx 100k$ and/or inputs with higher dimension, we computed the bounds on random subsets of anchor samples with size $K=1k$ or $K=3k$ from the train or test splits over $20$ different random seeds. As $K < M$ Assumption~\ref{as:levelstauxy} is not satisfied. Here we reported the mean and standard deviation of the resulting bounds. For even larger datasets $M \gg 100k$ where a sample number $K \approx 3k$ is not representative of the dataset, the implementation of Algorithm~\ref{alg:feasset} using nearest neighbors may not be feasible. Extending Algorithm~\ref{alg:feasset} to be applicable to this scale is beyond the scope of this work.

\noindent \textbf{Comparison to theoretical assumptions and guarantees:} Theorem~\ref{thm:lowerboundapprx} and Corollary~\ref{thm:lowerboundapprxclass} only provide bounds on the error, respectively accuracy, on the constructed dataset $\mathcal{D}_{M'}$ from Algorithm~\ref{alg:feasset}.  We explicitly computed the accuracy on the original and constructed dataset to validate the bounds for the synthetic regression example, MNIST, CIFAR-10 and EUROSAT. For the remaining datasets we did not compute the regressor errors or classifier accuracies on the constructed dataset $\mathcal{D}_{M'}$ from Algorithm~\ref{alg:feasset} and instead report errors and accuracies from benchmark dataset challenges and published state-of-the-art models on the same datasets. In these cases, for CIFAR-10 and RESISC45, differing evaluation protocols make published results not directly comparable to the $\mathcal{D}_{M'}$-based bounds and thus not violations of them. We did not compare the average kernel size against empirical conditional variance as a general baseline, as it is only directly estimable with additional assumptions, such as i.i.d. samples, in Euclidean regression settings with repeated or conditionally repeated observations. For the same reason, we did not compute the conditional mean, which is optimal according to Proposition \ref{prop:bayes-kernel-identity}. 

\subsection{Regression tasks}

\noindent For all regression datasets considered, the accuracy bounds provided by Theorem~\ref{thm:lowerboundapprx} in conjunction with Algorithm~\ref{alg:feasset} are provided in Table \ref{tab:regression-results} and the parameters used for computation are provided in Table \ref{tab:regression-parameters}. We followed the procedure summarized above and outlined in the Appendix in Section \ref{sec:feassetsamp}. We computed the bounds on a synthetic regression dataset, the Tropical Cyclone open-source dataset from \href{https://huggingface.co/datasets/TCBench/TCBench}{Huggingface}, the UCI Energy Efficiency open-source datasets \cite{tsanas2012accurate} from \href{https://archive.ics.uci.edu/dataset/242/energy+efficiency}{UCI}, the UCI Concrete Compressive Strength open-source dataset \cite{yeh1998modeling} from \href{https://archive.ics.uci.edu/dataset/165/concrete+compressive+strength}{UCI} and the KAERI Critical Heat Flux (CHF) open-source dataset with uniform axial heating~\cite{park2000effect} from \href{https://www.kaggle.com/datasets/kaustubhdhote/critical-heat-flux-dataset/data}{Kaggle}. Detailed dataset descriptions, as well as the RMSE of the optimal regressor on the constructed test dataset from the proof of Theorem \ref{thm:lowerboundapprx}, are in Appendix in Section \ref{sec:regdets}.

\begin{table}[h!]
\centering
\caption{\textbf{Parameters for the regression tasks}. Here, \(M\) denotes the original dataset size, \(M'\) the constructed dataset size generated by Algorithm~\ref{alg:feasset}, \(K\) the number of anchors, $N_{\max}$ the maximum number of nearest neighbors and \(\overline{N(k)}\) the mean size of the feasible sets from the test dataset. $^*$: for tropical cyclone the mean$\pm$ the standard deviation of $\tau_{\mathcal X}$ across $20$ subsets of size $1k$ are: $(\overline{\tau_{\mathcal X}})_{\mathrm{train}}=284.7\pm
11.5$, $(\overline{\tau_{\mathcal X}})_{\mathrm{test}}=227.7\pm 8.9$, $\overline{N(k)}=169.4\pm 2.7$}
\label{tab:regression-parameters}
\small
\setlength{\tabcolsep}{3pt}
\scalebox{0.83}{
\begin{tabular}{l  c c c r r r r r r cc cc}
\toprule
Dataset & 
$d_{\mathcal X}$ & $d_{\mathcal Y}$ & $p$ &
$M_{\mathrm{train}}$ & $M_{\mathrm{test}}$ & $M'_{\mathrm{test}}$ & $K_{\mathrm{train}}$ & $K_{\mathrm{test}}$ & $N_{\max}$ &
$\overline{N(k)}$ & q &
$(\tau_{\mathcal X})_{\mathrm{train}}$ & $(\tau_{\mathcal X})_{\mathrm{test}}$ \\
\midrule
Synthetic &
$\ell_2$ & $\ell_2$ & $2$ &
$90$ & $810$ & $42.9k$ & $90$ & $810$ & $300$ &
$53.1$ & - & $0.014$ & $0.014$ \\

Tropical Cyclone$^*$ &
Wavelet-$\ell_2$ & $\ell_2$ & $2$ &
$70k$ & $44k$ & $153.7k$ &
$1k$ & $1k$ & $300$ &
$153.7$ & $0.1$ &
$284.7$ & $227.7$ \\

Energy Efficiency $Y_1$ &$\ell_2$ & $\ell_2$ & $2$ & $768$ & - & $1.3k$ &
$768$ & - & $300$ &
$1.75$ & $0.02$ &
$0.65$ & - \\
Energy Efficiency $Y_2$ &$\ell_2$ & $\ell_2$ & $2$ & $768$ & - & $1.3k$ &
$768$ & - & $300$ &
$1.75$ & $0.02$ &
$0.65$ & - \\
Concrete Strength &$\ell_2$ & $\ell_2$ & $2$ & $1030$ & - & $1.1k$ &
$1030$ & - & $300$ &
$1.10$ & $0.02$ &
$0.0$ & - \\
Critical Heat Flux & $\ell_2$ & $\ell_2$ & $2$ & $651$ & - & $59.2k$ &
$651$ & - & $300$ &
$90.9$ & - &
$0.17$ & - \\
\bottomrule
\end{tabular}}
\end{table}

\begin{table}[h!]
\centering
\caption{\textbf{Optimal regression accuracy intervals and regressor errors}. The theoretical interval from Theorem~\ref{thm:lowerboundapprx} is
$\left[
1/2\operatorname{Kersize}(\mathcal D_{M'},p)_K,\,
\operatorname{Kersize}(\mathcal D_{M'},p)_K
\right]$. All scores for models $\phi_{n}$ with $n \in \{1,2,3\}$ are RMSE values, $ \mathcal{L}(\mathcal{D}_M, \phi, 2)$, in the units of the target variable reported in the three right columns. $^{(1)}$: models evaluated on a different test split. $^*$: for tropical cyclone the mean bounds across $20$ subsets of size $1k$ are reported.}
\label{tab:regression-results}
\small
\setlength{\tabcolsep}{4pt}
\scalebox{0.85}{
\begin{tabular}{l  c c | c c | c c | c c}
\toprule
Dataset &
Interval train &
Interval test &
$\phi_{1st}$, $M_{\mathrm{test}}$ & $M'_{\mathrm{test}}$ &
$\phi_{2nd}$, $M_{\mathrm{test}}$ & $M'_{\mathrm{test}}$ &
$\phi_{3rd}$, $M_{\mathrm{test}}$ & $M'_{\mathrm{test}}$ \\
\midrule
Synthetic &
$[0.0102,0.0203]$ &
$[0.0113,0.0225]$ &
$0.0169$ & $0.0221$  &
$0.0161$ & $0.0225$ &
$0.0607$ & $0.0720$ \\

Tropical Cyclone$^*$  &
$[3.83,7.65]$ &
$[3.38,6.76]$ &
$6.26$ & - &
$6.42$ & - &
$6.46$ & - \\
Energy Efficiency $Y_1$$^{(1)}$ & $[0.28,0.55]$ & - & $0.51$ & - & $0.56$  &- & $0.67$ & - \\
Energy Efficiency $Y_2$$^{(1)}$ & $[0.78,1.55]$ & -  & $1.28$ & - & $1.64$  &- & $1.79$ & - \\
Concrete Strength$^{(1)}$ & $[1.41, 2.81]$ & -  & $1.97$ & - & $2.08$  &- & $2.10$ & - \\
Critical Heat Flux$^{(1)}$ & $[106.49, 212.97]$ & -  & $247.76$ & - & $251.46$  &- & $266.98$ & - \\
\bottomrule
\end{tabular}}
\end{table}

\subsubsection{Relation to the accuracy-stability trade-off}
\label{subsec:accuracy_stability_trade-off}

\begin{figure}[h!]
\centering
\begin{tikzpicture}
\begin{axis}[
    width=14cm,
    height=7cm,
    xlabel={RMSE},
    ylabel={Lipschitz lower bound $\hat{L}$},
    xmode=log,
    xmin=1.6e-4,
    xmax=4e-1,
    ymin=0.5,
    ymax=4.0,
    grid=major,
    legend style={
        at={(0.1,0.42)},
        anchor=north west,
        legend columns=2
    }
]

\addplot[
    only marks,
    mark=*,
    blue
] coordinates {
    (0.000297,2.113529)
};
\addlegendentry{\small{$GP^o$ train}}

\addplot[
    only marks,
    mark=square*,
    blue,
] coordinates {
    (0.150666,2.113529)
};
\addlegendentry{\small{$GP^o$ test}}

\addplot[
    only marks,
    mark=*,
    red
] coordinates {
    (0.014644,0.930406)
};
\addlegendentry{\small{$GP$ train}}

\addplot[
    only marks,
    mark=square*,
    red,
] coordinates {
    (0.016850,0.930406)
};
\addlegendentry{\small{$GP$ test}}

\addplot[
    only marks,
    mark=*,
    purple
] coordinates {
    (0.000177,3.889222)
};
\addlegendentry{\small{$SVR^o$ train}}

\addplot[
    only marks,
    mark=square*,
    purple,
] coordinates {
    (0.309775,3.889222)
};
\addlegendentry{\small{$SVR^o$ test}}

\addplot[
    only marks,
    mark=*,
    brown
] coordinates {
    (0.013384,0.921996)
};
\addlegendentry{\small{$SVR$ train}}

\addplot[
    only marks,
    mark=square*,
    brown,
] coordinates {
    (0.016177,0.921996)
};
\addlegendentry{\small{$SVR$ test}}

\addplot[
    draw=none,
    fill=blue!20,
    fill opacity=0.35
] coordinates {
    (0.011261,0.5)
    (0.022522,0.5)
    (0.022522,4.5)
    (0.011261,4.5)
} \closedcycle;

\node[
    align=center,
    font=\small
] at (axis cs:0.015,3.8)
{$\frac{1}{2}\mathrm{Kersize}
\leq \mathrm{RMSE}
\leq\mathrm{Kersize}$};

\addplot[
    dashed,
    black, 
    forget plot
] coordinates {
    (1e-4,0.9)
    (5e-1,0.9)
};
\addlegendentry{\small{$L_f=0.9$ }}

\end{axis}
\end{tikzpicture}

\caption{\textbf{Accuracy-stability trade-off.} The empirical Lipschitz constant compared against the RMSE of GP and SVR regressors on the train and test dataset. The average kernel size was computed from the reference dataset $\mathcal{D}_{M'_{\mathrm{train}}}$. Regression models $\phi_{\mathrm{n}}^{o}$ satisfy $\mathrm{RMSE}_{\mathrm{train}}(\phi_{\mathrm{n}}^{o}, \mathcal{D}_{M_{\mathrm{train}}})<\frac12\operatorname{Kersize}(\mathcal{D}_{M'_{\mathrm{train}}},2)$, whereas models $\phi_{\mathrm{n}}$ satisfy $\frac12\operatorname{Kersize}(\mathcal{D}_{M'_{\mathrm{train}}},2)\leq\mathrm{RMSE}_{\mathrm{train}}(\phi_{\mathrm{n}}, \mathcal{D}_{M_{\mathrm{train}}})\leq\operatorname{Kersize}(\mathcal{D}_{M'_{\mathrm{train}}},2)$. The parameters for computing the average kernel size were $N_{\mathrm{max}}=300$, $\tau_{\mathcal{X}}=0.014$ and $\tau_{\mathcal{Y}}=0.03$. The exact numerical values and errors on the constructed test dataset are reported in the Appendix in Table \ref{tab:all-model-trade-off}.}
\label{fig:q015_trade-off}
\end{figure}
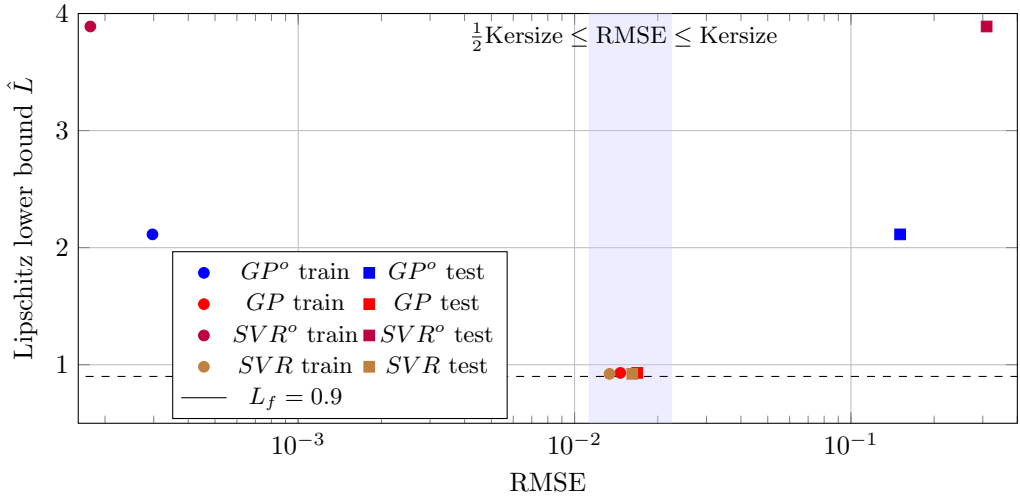

\noindent We studied the accuracy-stability trade-off from Proposition \ref{prop:instabilityaccuracy} for the synthetic regression
problem $f(x_1,x_2) =\sin\!\left(1+0.9\sqrt{x_1^2+x_2^2}\right)$ with input noise $U([0,0.01]^2)$ and target noise $U([0,0.03])$. The accuracy stability trade-off is visualized in Figure \ref{fig:q015_trade-off}. We trained two versions of each of the two regressors to approximate the target function $f$ from samples in a train dataset $\mathcal{D}_{M_{\mathrm{train}}}$ to obtain errors below half of the average kernel size and above on the train dataset, the latter attaining the accuracy within the bounds from Theorem \ref{thm:lowerboundapprx}. We considered two Gaussian Process regressors, and two Support Vector Regression (SVR) models with a radial basis function (RBF) kernel and optimized the model hyperparameters with a brute-force grid search to meet the accuracy bounds conditions. We report the RMSE of each model on both the original train and test datasets and the constructed test dataset generated by Algorithm~\ref{alg:feasset} and a lower bound to the Lipschitz constant of all models on the test dataset in the Appendix in Table \ref{tab:all-model-trade-off}. For noiseless inputs \(f\) is \(0.9\)-Lipschitz with respect to the Euclidean metric, since \(r(x) = \|x\|_2\) is \(1\)-Lipschitz and \(g(r) = \sin(1 + 0.9r)\) is \(0.9\)-Lipschitz, and hence, by composition, \(f = g \circ r\) is \(0.9\)-Lipschitz. Inspired by Proposition \ref{prop:instabilityaccuracy}, we compute $1/\tau_{\mathcal{X}}(\tau_{\mathcal{Y}}-2\mathrm{RMSE}_{\mathrm{train}}(\phi_{\mathrm{n}}^{o}, \mathcal{D}_{M_{\mathrm{train}}}))$ and compare it to the empirical Lipschitz constant of the unstable regressors, see Table \ref{tab:all-model-trade-offquant} in the Appendix.

\subsubsection{Sensitivity of the average kernel size to noise}

\noindent In Figures \ref{fig:kernel_size_noise} and \ref{fig:kernel_size_target_noise} we illustrate the effect of increasing input and target noise of the synthetic regression problem on the average kernel size. For this we chose a larger dataset size of $M=2.5k$ and increase the thresholds for feasible set sampling according to the increasing noise level on inputs or outputs. In Section \ref{sec:regdets} in the Appendix, we report the average kernel size for a noiseless control setting, evaluated the effect of misspecified thresholds \(\tau_{\mathcal X}\) and \(\tau_{\mathcal Y}\) and compare the empirical average kernel size with the Bayes RMSE under an additional independence assumption, as well as evaluate Assumption~\ref{as:levelstauxy} on admissible-pairs.

\begin{figure}[h!]
\centering

\begin{minipage}{0.48\linewidth}
\centering

\begin{tikzpicture}
\begin{axis}[
    width=\linewidth,
    height=0.65\linewidth,
    xlabel={Input noise level $\sigma_x$},
    ylabel={Average kernel size},
    grid=major,
    ymin=0.0,
    ymax=0.22,
    xtick={0.1,0.2,0.3,0.4,0.5,0.6,0.7,0.8,0.9,1.0},
    xticklabel style={rotate=45, anchor=east},
    scaled y ticks=false,
    yticklabel style={
        /pgf/number format/fixed,
        /pgf/number format/precision=3
    },
    legend style={at={(0.02,0.98)},anchor=north west},
]

\addplot[
    blue,
    thick,
    mark=*,
    mark size=2pt
]
coordinates {
    (0.01,0.0059615056)
    (0.05,0.0071006579)
    (0.10,0.0111933636)
    (0.20,0.0245780876)
    (0.30,0.0417339077)
    (0.40,0.0633912618)
    (0.50,0.0867726464)
    (0.60,0.1032731820)
    (0.70,0.1350157183)
    (0.80,0.1549749295)
    (0.90,0.1804439306)
    (1.00,0.2070803346)
};

\end{axis}
\end{tikzpicture}


\captionof{figure}{Average kernel size as a function of input noise level $\sigma_x$. 
Constant $N_{\max}=300$ and $\tau_{\mathcal{Y}}=0.03$, $\tau_{\mathcal{X}}=\sqrt{2}\sigma_x$. Corresponding values of $M'$ and $\tau_{\mathcal{X}}$ are in Table \ref{tab:kernel_size_noise} in the Appendix.}
\label{fig:kernel_size_noise}

\end{minipage}
\hfill
\vspace{-0.01cm}
\begin{minipage}{0.48\linewidth}
\centering
\begin{tikzpicture}
\begin{axis}[
    width=\linewidth,
    height=0.65\linewidth,
    xlabel={target noise level $\sigma_y$},
    ylabel={Average kernel size},
    grid=major,
    ymin=0,
    ymax=0.25,
    xtick={0.1,0.2,0.3,0.4,0.5,0.6,0.7,0.8,0.9,1.0},
    xticklabel style={rotate=45, anchor=east},
    scaled y ticks=false,
    yticklabel style={
        /pgf/number format/fixed,
        /pgf/number format/precision=3
    },
    legend style={at={(0.02,0.98)},anchor=north west},
]

\addplot[
    blue,
    thick,
    mark=*,
    mark size=2pt
]
coordinates {
    (0.01,0.0059615)
    (0.05,0.0155632)
    (0.10,0.0255227)
    (0.20,0.0474117)
    (0.30,0.0715549)
    (0.40,0.0978868)
    (0.50,0.1138221)
    (0.60,0.1434881)
    (0.70,0.1626441)
    (0.80,0.1917276)
    (0.90,0.2138467)
    (1.00,0.2370987)
};

\addlegendentry{Average kernel size}

\end{axis}
\end{tikzpicture}

\captionof{figure}{Average kernel size as a function of target noise level $\sigma_y$. 
Constant $N_{\max}=300$ and $\tau_{\mathcal{X}}=0.01414$, and $\tau_{\mathcal{Y}}=\sigma_y$.
Corresponding values of $M'$ are in Table \ref{tab:kernel_size_target_noise} in the Appendix.}
\label{fig:kernel_size_target_noise}

\end{minipage}

\end{figure}

\subsection{Classification tasks}

\noindent For all classification datasets considered, the accuracy bounds provided by Corollary~\ref{thm:lowerboundapprxclass} in conjunction with Algorithm~\ref{alg:feasset} are provided in Table \ref{tab:classification-results} and the parameters used for computation are provided in Table \ref{tab:classification-parameters} below. We followed the procedure summarized above and outlined in the Appendix in Section \ref{sec:feassetsamp}. We computed the bounds on the MNIST open-source dataset from \href{https://docs.pytorch.org/vision/main/generated/torchvision.datasets.MNIST.html}{Torch Vision}, the CIFAR-10 open-source dataset from \href{https://docs.pytorch.org/vision/main/generated/torchvision.datasets.CIFAR10.html}{Torch Vision}, the RESISC45 open-source dataset \cite{cheng2017remote} from \href{https://docs.torchgeo.org/en/v0.3.1/_modules/torchgeo/datamodules/resisc45.html}{TorchGeo} and the EUROSAT open-source dataset \cite{helber2019eurosat} from \href{https://docs.pytorch.org/vision/main/generated/torchvision.datasets.EuroSAT.html}{Torch Vision}. Detailed dataset descriptions, as well as the accuracy of the optimal classifier on the constructed test dataset from the Corollary~\ref{thm:lowerboundapprxclass}, are in Appendix in Section \ref{sec:classdets}.

\begin{table}[h!]
\centering
\caption{\textbf{Parameters for the classification tasks}. Here, \(M\) denotes the original dataset size, \(M'\) the constructed dataset size generated by Algorithm~\ref{alg:feasset}, \(K\) the number of anchors, $N_{\max}$ the maximum number of nearest neighbors and \(\overline{N(k)}\) the mean size of the feasible sets from the test dataset. $^*$: for RESISC45 the mean$\pm$ the standard deviation across $20$ subsets of size $3k$ are: $(\tau_{\mathcal X})_{\mathrm{train}}=31.5\pm
0.5$, $(\tau_{\mathcal X})_{\mathrm{test}}=31.8\pm 0.7$, $\overline{N(k)}=1.58\pm 0.08$.}
\label{tab:classification-parameters}
\small
\setlength{\tabcolsep}{3pt}
\scalebox{0.85}{
\begin{tabular}{l  c c c r r r  r r c c c c c}
\toprule
Dataset & 
$d_{\mathcal X}$ & $d_{\mathcal Y}$ & $p$ &
$M_{\mathrm{train}}$ & $M_{\mathrm{test}}$ &
$M'_{\mathrm{test}}$ & $K_{\mathrm{train}}$ & $K_{\mathrm{test}}$ &
$N_{\max}$ & $\overline{N(k)}$ &
$q$ & $(\tau_{\mathcal X})_{\mathrm{train}}$ & $(\tau_{\mathcal X})_{\mathrm{test}}$ \\
\midrule

MNIST &
Wavelet-$\ell_2$ & $d_H$ & $1$ &
$50k$ & $10k$ &
$25.1k$ & $50k$ & $10k$ &
$200$ & $2.51$ &
$0.1$ & $4.73$ & $5.38$ \\

CIFAR-10 &
Wavelet-$\ell_2$ & $d_H$ & $1$ &
$40k$ & $10k$ &
$13.0k$ & $40k$ & $10k$ &
$200$ & $1.30$ &
$0.05$ & $27.4$ & $30.8$ \\

RESISC45$^*$ &
Wavelet-$\ell_2$ & $d_H$ & $1$ &
$25.2k$ & $6.3k$
& $4.7k$ & $3k$ & $3k$ & $200$ & $1.58$ &
$0.05$ & $31.5$ & $31.8$ \\

EUROSAT &
Wavelet-$\ell_2$ & $d_H$ & $1$ &
$16.2k$ & $5.4k$ &
$15.0k$ & $16.2k$ &
$5.4k$ & $200$ & $2.77$ &
$0.09$ & $2.44$ & $2.70$ \\

\bottomrule
\end{tabular}}
\end{table}

\begin{table}[h!]
\centering
\caption{\textbf{Optimal classification accuracy intervals and classifier accuracies.} Theoretical intervals from Corollary~\ref{thm:lowerboundapprxclass} are reported as percentages; explicitly, the
interval is 
$
100 \times \left[
1-\operatorname{Kersize}(\mathcal D_{M'},1)_K,\,
1-1/2\operatorname{Kersize}(\mathcal D_{M'},1)_K
\right].
$
Accuracies, $\operatorname{Acc}(\mathcal{D}_M,\phi)$, for models $\phi_{n}$ with $n \in \{1,2,3\}$ are reported on the \([0,100]\) scale in the three right columns. $^{(1)}$: model(s) evaluated on a different test split. $^*$: for RESISC45 the mean bounds across $20$ subsets of size $3k$ are reported.}
\label{tab:classification-results}
\small
\setlength{\tabcolsep}{4pt}
\scalebox{0.9}{
\begin{tabular}{l  c c| c | c c| c c}
\toprule
Dataset & 
Interval train &
Interval test &
$\phi_{1st}$, $M_{\mathrm{test}}$& 
$\phi_{2nd}$, $M_{\mathrm{test}}$& $M'_{\mathrm{test}}$ &
$\phi_{3rd}$, $M_{\mathrm{test}}$ & $M'_{\mathrm{test}}$ \\
\midrule
MNIST &
$[100.0,100.0]$ &
$[100.0,100.0]$ & $99.53$ &
$99.45$ & $99.78$ &
$99.43$ & $99.77$ 
 \\
CIFAR-10 &
$[85.61,92.81]$ &
$[83.76,91.88]$ & $89.61$$^{(1)}$&
$91.19$ & $79.30$ &
$83.20$& $73.25$ 
 \\
RESISC45$^{*,(1)}$ &
$[83.85,91.93]$&
$[82.36,91.18]$ &
$90.36\pm0.18$ & 
$86.02\pm0.18$ & - &
$85.16\pm0.18$ & - \\

EUROSAT &
$[97.78,98.89]$ &
$[96.89,98.44]$ &
$98.00$$^{(1)}$ & 
$98.00$$^{(1)}$ & - &
$97.06$ & $96.63$\\

\bottomrule
\end{tabular}}
\end{table}

\subsubsection{Aleatoric uncertainty: dataset dependence and model independence of the accuracy bounds}

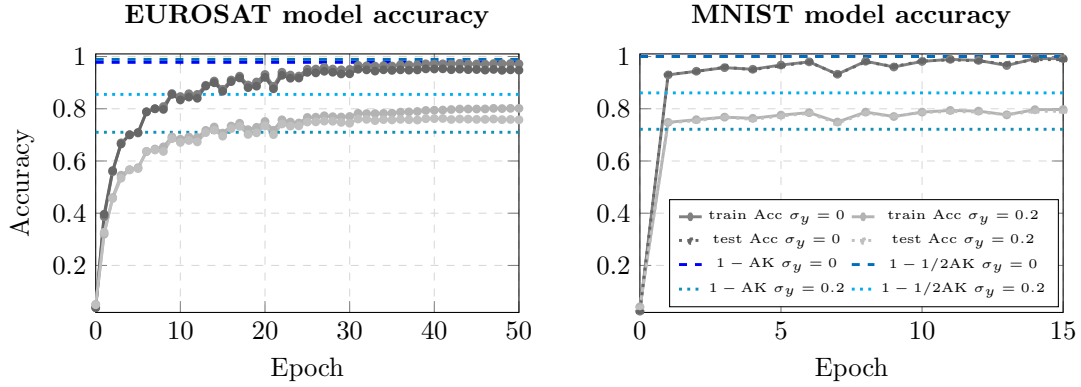
\begin{figure}[h!]
\begin{minipage}[t]{0.49\textwidth}
    \centering
\begin{tikzpicture}
\begin{axis}[width=\linewidth, 
height=5cm, 
xmin=0, 
xmax=50, 
ymin=0.02,
ymax=1.01,
xlabel={Epoch}, 
ylabel={Accuracy}, 
grid=major, 
grid style={dashed, gray!30}, 
every axis plot/.append style={line width=1.1pt}, 
title={\textbf{EUROSAT model accuracy}},]
\addplot[color=white!50!black, mark=*, mark size=1.1pt, ] coordinates {(0, 0.0386) (1, 0.3867) (2, 0.5588) (3, 0.6694) (4, 0.6996) (5, 0.7072) (6, 0.7886) (7, 0.8001) (8, 0.8084) (9, 0.8583) (10, 0.8468) (11, 0.8575) (12, 0.855) (13, 0.8898) (14, 0.9062) (15, 0.8744) (16, 0.9101) (17, 0.9248) (18, 0.8909) (19, 0.9025) (20, 0.9312) (21, 0.8862) (22, 0.9398) (23, 0.9291) (24, 0.9253) (25, 0.9514) (26, 0.9519) (27, 0.9588) (28, 0.952) (29, 0.9491) (30, 0.9509) (31, 0.9662) (32, 0.9643) (33, 0.9633) (34, 0.9651) (35, 0.9635) (36, 0.9683) (37, 0.9679) (38, 0.9681) (39, 0.9721) (40, 0.9714) (41, 0.972) (42, 0.9717) (43, 0.974) (44, 0.9714) (45, 0.9715) (46, 0.9712) (47, 0.973) (48, 0.9725) (49, 0.9733) (50, 0.9718)};
\addplot[color=white!70!black, mark=*, mark size=1.1pt, ] coordinates {(0, 0.051) (1, 0.3199) (2, 0.4599) (3, 0.5455) (4, 0.5668) (5, 0.5742) (6, 0.636) (7, 0.6458) (8, 0.654) (9, 0.692) (10, 0.6834) (11, 0.6919) (12, 0.6899) (13, 0.7171) (14, 0.7301) (15, 0.7067) (16, 0.7338) (17, 0.7445) (18, 0.72) (19, 0.729) (20, 0.7512) (21, 0.7176) (22, 0.7576) (23, 0.7491) (24, 0.7454) (25, 0.7667) (26, 0.7677) (27, 0.7731) (28, 0.7696) (29, 0.7683) (30, 0.7704) (31, 0.7825) (32, 0.7814) (33, 0.7815) (34, 0.7822) (35, 0.7831) (36, 0.787) (37, 0.7879) (38, 0.7896) (39, 0.7935) (40, 0.7944) (41, 0.7952) (42, 0.797) (43, 0.7989) (44, 0.799) (45, 0.799) (46, 0.7993) (47, 0.8015) (48, 0.8006) (49, 0.8015) (50, 0.8015)};
\addplot[color=white!40!black, mark=*, mark size=1.1pt, ] coordinates {(0, 0.038) (1, 0.3961) (2, 0.5631) (3, 0.6652) (4, 0.7006) (5, 0.7091) (6, 0.7878) (7, 0.8) (8, 0.7976) (9, 0.855) (10, 0.8339) (11, 0.8446) (12, 0.8402) (13, 0.8869) (14, 0.8996) (15, 0.867) (16, 0.9039) (17, 0.9181) (18, 0.8809) (19, 0.8865) (20, 0.9202) (21, 0.8765) (22, 0.928) (23, 0.9169) (24, 0.9176) (25, 0.9393) (26, 0.9374) (27, 0.9435) (28, 0.9398) (29, 0.9335) (30, 0.9339) (31, 0.9513) (32, 0.9489) (33, 0.9443) (34, 0.9493) (35, 0.9469) (36, 0.9491) (37, 0.9478) (38, 0.9487) (39, 0.9528) (40, 0.9524) (41, 0.953) (42, 0.9496) (43, 0.9515) (44, 0.9498) (45, 0.9502) (46, 0.9481) (47, 0.9506) (48, 0.9472) (49, 0.9491) (50, 0.9476)};
\addplot[color=white!75!black, mark=*, mark size=1.1pt, ] coordinates {(0, 0.0494) (1, 0.327) (2, 0.4565) (3, 0.5341) (4, 0.5657) (5, 0.5709) (6, 0.637) (7, 0.642) (8, 0.6365) (9, 0.6854) (10, 0.6711) (11, 0.6785) (12, 0.6731) (13, 0.7087) (14, 0.718) (15, 0.695) (16, 0.7239) (17, 0.7328) (18, 0.7017) (19, 0.7078) (20, 0.7348) (21, 0.7009) (22, 0.7417) (23, 0.7337) (24, 0.7344) (25, 0.7493) (26, 0.7498) (27, 0.7528) (28, 0.7522) (29, 0.7478) (30, 0.7446) (31, 0.7607) (32, 0.7594) (33, 0.7546) (34, 0.7585) (35, 0.7572) (36, 0.7593) (37, 0.758) (38, 0.7569) (39, 0.7619) (40, 0.7624) (41, 0.7628) (42, 0.7589) (43, 0.7607) (44, 0.7602) (45, 0.7593) (46, 0.7594) (47, 0.7606) (48, 0.7583) (49, 0.7591) (50, 0.7576)};
\addplot[color=blue, dashed, no markers, domain=\pgfkeysvalueof{/pgfplots/xmin}:\pgfkeysvalueof{/pgfplots/xmax}] {0.9777774156895495};
\addplot[color=NavyBlue, dashed, no markers, domain=\pgfkeysvalueof{/pgfplots/xmin}:\pgfkeysvalueof{/pgfplots/xmax}] {0.9888887078447747};
\addplot[color=cyan!70!black, dotted, no markers, domain=\pgfkeysvalueof{/pgfplots/xmin}:\pgfkeysvalueof{/pgfplots/xmax}] { 0.7099756613302273};
\addplot[color=cyan, dotted, no markers, domain=\pgfkeysvalueof{/pgfplots/xmin}:\pgfkeysvalueof{/pgfplots/xmax}] {0.8549878306651136};
\end{axis}
\end{tikzpicture}
\end{minipage}
\hfill
\begin{minipage}[t]{0.49\textwidth}
    \centering
\begin{tikzpicture}
\begin{axis}[width=\linewidth, 
height=5cm, 
xmin=0, 
xmax=15, 
ymin=0.02,
ymax=1.01,
xlabel={Epoch}, 
grid=major, 
grid style={dashed, gray!30}, 
legend style={
    at={(0.99,0.02)},
    anchor=south east,
    font=\tiny,
    legend columns=2,
    legend image post style={xscale=0.5},
}, 
every axis plot/.append style={line width=1.1pt}, 
title={\textbf{MNIST model accuracy}},]
\addplot[color=white!50!black, mark=*, mark size=1.1pt, ] coordinates {(0, 0.0249) (1, 0.9296) (2, 0.9437) (3, 0.9574) (4, 0.9512) (5, 0.9675) (6, 0.9795) (7, 0.9322) (8, 0.982) (9, 0.9599) (10, 0.9813) (11, 0.9892) (12, 0.9852) (13, 0.9656) (14, 0.9919) (15, 0.9921)};
\addlegendentry{train Acc $\sigma_y =0$}

\addplot[color=white!70!black, mark=*, mark size=1.1pt, ] coordinates {(0, 0.0416) (1, 0.748) (2, 0.7581) (3, 0.7679) (4, 0.7634) (5, 0.7756) (6, 0.7852) (7, 0.7492) (8, 0.7872) (9, 0.7703) (10, 0.7872) (11, 0.7936) (12, 0.791) (13, 0.7767) (14, 0.7966) (15, 0.7969)};
\addlegendentry{train Acc $\sigma_y =0.2$}

\addplot[color=white!40!black, mark=*, dotted, mark size=1.1pt, ] coordinates {(0, 0.0238) (1, 0.9311) (2, 0.9427) (3, 0.959) (4, 0.9498) (5, 0.9643) (6, 0.979) (7, 0.932) (8, 0.9797) (9, 0.9618) (10, 0.9827) (11, 0.9874) (12, 0.9833) (13, 0.9709) (14, 0.988) (15, 0.9882)};
\addlegendentry{test Acc $\sigma_y =0$}
\addplot[color=white!75!black, mark=*, dotted, mark size=1.1pt, ] coordinates {(0, 0.0394) (1, 0.7467) (2, 0.7539) (3, 0.7674) (4, 0.7593) (5, 0.7722) (6, 0.7823) (7, 0.7448) (8, 0.7836) (9, 0.7704) (10, 0.7856) (11, 0.7896) (12, 0.7871) (13, 0.7773) (14, 0.7903) (15, 0.7905)};
\addlegendentry{test Acc $\sigma_y =0.2$}
\addplot[color=blue, dashed, no markers, domain=\pgfkeysvalueof{/pgfplots/xmin}:\pgfkeysvalueof{/pgfplots/xmax}] {0.999981};
\addlegendentry{$1-\text{AK}$ $\sigma_y =0$}
\addplot[color=NavyBlue, dashed, no markers, domain=\pgfkeysvalueof{/pgfplots/xmin}:\pgfkeysvalueof{/pgfplots/xmax}] {0.99999};
\addlegendentry{$1-1/2\text{AK}$ $\sigma_y =0$}
\addplot[color=cyan!70!black, dotted, no markers, domain=\pgfkeysvalueof{/pgfplots/xmin}:\pgfkeysvalueof{/pgfplots/xmax}] {0.7214547872076542};
\addlegendentry{$1-\text{AK}$ $\sigma_y =0.2$}
\addplot[color=cyan, dotted, no markers, domain=\pgfkeysvalueof{/pgfplots/xmin}:\pgfkeysvalueof{/pgfplots/xmax}] {0.8607273936038271};
\addlegendentry{$1-1/2\text{AK}$ $\sigma_y =0.2$}
\end{axis}
\end{tikzpicture}
\end{minipage}
\caption{Accuracy of models trained on train datasets of EUROSAT and MNIST with noisy labels with $\sigma_y=0.2$. The models trained on noisy datasets are evaluated on both the clean and noisy test and train datasets. The lower and upper bounds from Corollary~\ref{thm:lowerboundapprxclass}, abbreviated as $[1-\text{AK}, 1-1/2\text{AK}]$, are computed on the noisy, $\sigma_y=0.2$, and clean, $\sigma_y=0$, train and test datasets. The number of anchor samples $K$ and the quantiles for computing $\tau_{\mathcal{X}}(q)$ are in Table \ref{tab:classification-parameters}. The average kernel size is computed with a larger $N_{max}=200$.}\label{fig:aleatoric-acc}
\end{figure}

\noindent Figure~\ref{fig:aleatoric-acc} compares the accuracy of networks trained on noisy versions of EUROSAT and MNIST with the ambiguity-based accuracy intervals derived from the average kernel size. We changed the labels of $\sigma_y$-fractions of the datasets to a different, random labels. In both datasets, adding label noise shifts the admissible accuracy range downward. On MNIST, the clean-data accuracy bounds are close to perfect accuracy, indicating that the task is nearly unambiguous under the chosen pseudo-metric on the input space, while the noisy-data accuracy bounds closely match the observed degradation in train and test accuracy. On EUROSAT, the clean-data accuracy bounds are already noticeably below \(1\) and reflect a higher baseline level of ambiguity.

\section{Discussion}

\noindent The deterministic bounds from Theorem~\ref{thm:lowerboundapprx} and Corollary~\ref{thm:lowerboundapprxclass} hold for the dataset constructed in Algorithm~\ref{alg:feasset}. Where a priori knowledge of the input and target noise is unavailable, we selected input thresholds so the average kernel size remained stable with changes in feasible set size. In these cases, for both regression and classification datasets, our stability analysis across quantiles for choosing input threshold $\tau_{\mathcal{X}}$ based on a practical empirical heuristic and feasible set size demonstrated that the bounds can be chosen to be relatively stable. We emphasize that the resulting kernel-size accuracy intervals are not an intrinsic property of the dataset alone, but rather, a property of the triple $(\mathcal D_M, d_{\mathcal X}, d_{\mathcal Y})$, and its interpretability depends on how well $d_{\mathcal X}$ captures task-relevant indistinguishability and which evaluation metric $d_{\mathcal Y}$ is used. After all, the evaluation metric can be optimized to obtain arbitrarily good accuracy \cite{redmon2018yolov3}. We chose a wavelet-based pseudo-metric for all image-based tasks to ensure comparability across tasks; however, other metrics may be more relevant to object classification tasks, such as MNIST or CIFAR-10.

\noindent \textbf{Regression accuracy bounds:} Table \ref{tab:regression-parameters} shows that the framework is applicable to a range of datasets of different sizes and ranging from image to tabular regression tasks. The bounds in Theorem \ref{thm:lowerboundapprx} and the regressor errors in Table \ref{tab:regression-results} show that the proposed intervals can be nontrivial and, in several cases, are numerically comparable to the performance reported for strong benchmark and published models. Figure \ref{fig:q015_trade-off} illustrates the accuracy-stability trade-off predicted by Proposition~\ref{prop:instabilityaccuracy}. The high-accuracy models have training errors smaller than half of the average kernel size, but this comes at the cost of significantly larger test errors and higher empirical Lipschitz constant estimates, $\widehat{L}$, substantially exceeding the Lipschitz constant of the noiseless function generating the dataset. The models that satisfy the prescribed accuracy interval on the training set from Theorem~\ref{thm:lowerboundapprx}, achieve lower test errors while maintaining $\widehat{L}$ close to the true regularity of the underlying data-generating function, see Table \ref{tab:all-model-trade-off} in the Appendix.

\noindent \textbf{Classification accuracy bounds:}  Table \ref{tab:classification-parameters} shows that the framework is applicable to a range of datasets of different sizes and image classification tasks. The accuracy bounds in Corollary \ref{thm:lowerboundapprxclass} and the classifier accuracies in Table \ref{tab:classification-results} show that the proposed intervals can be nontrivial and, in several cases,  provide a descriptive reference scale, but benchmark accuracies on $\mathcal D_M$, especially under different splits or protocols, are not directly comparable to the bounds on $\mathcal D_{M'}$. The train and test accuracies of the best model for the task of mapping images (I) to random classes (RC), reported in \cite[Table 1]{zhang2016understanding}, illustrate the lower bound on the generalization gap for classification in Proposition~\ref{thm:gengapbd}.

\noindent \textbf{Average kernel sizes and aleatoric uncertainty:} For the synthetic regression example, Figures \ref{fig:kernel_size_noise} and \ref{fig:kernel_size_target_noise} show that the average kernel sizes increase with increasing input and target noise. Combined with the model-independent bounds on empirical error in Theorem~\ref{thm:lowerboundapprx}, this indicates that the average kernel sizes capture increasing aleatoric uncertainty in the regression task. For MNIST and EUROSAT, Figure~\ref{fig:aleatoric-quantiles} shows that the average kernel sizes increase with increasing label noise. Together with Corollary~\ref{thm:lowerboundapprxclass} and Figure \ref{fig:aleatoric-acc}, this suggests that, under the chosen pseudo-metric and feasible-set construction, the resulting accuracy bounds capture a data-dependent component of aleatoric uncertainty. Here we note that, in the examples provided whenever we assumed ideal, noiseless targets with \(\tau_{\mathcal Y}=0\), our results may underestimate the aleatoric uncertainty of the corresponding tasks.

\section{Limitations \& Future Work}
As shown above, the proposed framework can provide realistic accuracy bounds for a wide range of applications and tasks. Nevertheless, several open points limit its functionality in certain aspects and could be investigated further in future work. 

\noindent First, the resulting bounds depend on the pair $(\mathcal D_M ,d_{\mathcal X})$ and the chosen evaluation metric $d_{\mathcal Y}$. The interpretability of the feasible sets and the resulting average kernel size depend on whether the chosen pseudo-metric $d_{\mathcal X}$ captures task-relevant indistinguishability. A poor metric may either merge unrelated inputs or fail to detect genuine ambiguity. In this work, we found that relatively simple metrics can already yield nontrivial bounds. For image data, metrics induced by transforms need not align with semantic similarity. Investigating different kinds of metrics, universal feature extractors (e.g., pretrained foundation models), or expert knowledge-based feature extractors could enable targeting the task-relevant features for the bounds more explicitly. 

\noindent Second, the feasible set construction depends on the threshold parameters $\tau_{\mathcal X}$ and $\tau_{\mathcal Y}$. Ideally, these can be selected from prior knowledge of input and target noise and otherwise, alternatives are selecting the input threshold $\tau_{\mathcal{X}}$ with a practical empirical heuristic based on quantiles or selecting the thresholds in order to guarantee a sufficiently small lower bound to the Lipschitz constant in the accuracy-stability trade-off from Proposition~\ref{prop:instabilityaccuracy}; however, Assumption~\ref{as:levelstauxy} may be difficult to verify in practice and further properties or rules for selecting the thresholds in the absence of prior knowledge on input and target noise should be evaluated at a larger scale. 

\noindent Third, Algorithm~\ref{alg:feasset} provides finite-sample approximations of the feasible sets. If the dataset under-samples rare but valid targets for a given input, then the empirical average kernel size may underestimate the true spread. Hence, the lower bound is data-dependent and may be conservative when the dataset significantly under-samples rare but valid targets. An important next step is to provide construction diagnostics with guarantees, including feasible-set size distributions, singleton fractions, in order to distinguish ambiguity arising from local input neighborhoods from ambiguity induced primarily by the target tolerance. Here a theoretical direction for future work is to bridge the gap between the ideal distributional analogue quantity and the finite-sample average kernel size. This includes consistency of Algorithm~\ref{alg:feasset} as $\tau_{\mathcal X}\to 0$ with growing local sample counts, and an accuracy-stability frontier obtained by minimizing empirical loss over predictors with prescribed Lipschitz constant.

\noindent Fourth, the optimal accuracy intervals, as in Theorem~\ref{thm:lowerboundapprx}, $\left[\tfrac{1}{2}\operatorname{Kersize}(\mathcal D_{M'},p)_K,\,\operatorname{Kersize}(\mathcal D_{M'},p)_K\right]$, can be non-sharp in some applications where additional assumptions that yield sharper bounds are satisfied. While these provide a nontrivial model-independent accuracy range, they may not always identify the exact optimal empirical error due to the reasons outlined above. The proposed bounds quantify only one component of prediction difficulty, namely the ambiguity of the input-target multiset in the chosen (pseudo-)metric spaces. They do not account directly for optimization difficulty or finite model capacity. The optimum regressor or classifier is explicit in some special cases, in which we provide the corresponding error and accuracy in the Appendix, yet the bounds remain applicable in general metric settings where the anchor-wise minimization need not admit a closed form and may require solving a separate optimization problem for each anchor. 

\section{Conclusion}

\noindent We introduced a geometric framework for deriving deterministic, non-asymptotic bounds on achievable empirical error and accuracy in regression and classification on general metric and pseudo-metric spaces. The central quantity, the average kernel size, measures target dispersion within finite feasible sets and yields a model-independent reference scale for empirical prediction difficulty on the constructed dataset. A complementary deterministic comparison bound relates empirical loss on the constructed dataset to empirical loss on the original dataset through a graph-based reassignment error, showing how the discrepancy is controlled by the thresholds, the predictor regularity, and the induced reference weights. Together with the stability result, the framework shows that fitting ambiguous targets below this scale entails increased local instability. Under additional probabilistic assumptions, the distributional analogue average kernel size is related to Bayes risk, implies positive generalization gaps below the ambiguity-compatible level, and admits a finite-sample concentration guarantee under an ideal conditional-sampling design. The average kernel size computed by Algorithm~\ref{alg:feasset} should therefore be interpreted as a computable geometric surrogate for this ideal quantity unless further assumptions justify a statistical consistency claim. Important directions for future work include task-dependent pseudo-metric design and bridging the gap between the ideal sampling theory and the finite-sample construction used in practice.

\section{Acknowledgements}
N. M. Gottschling acknowledges research sponsored by the Laboratory Directed Research and Development Program of Oak Ridge National Laboratory, managed by UT-Battelle, LLC, for the U. S. Department of Energy. J.~Gawlikowski acknowledges the funding provided by the German Aerospace Center from the department EO Data Science at the Earth Observation Center. This work was supported by the Helmholtz Association's Initiative and Networking Fund on the HAICORE (at) FZJ partition. M. Caprio gratefully acknowledges support from the Prob\_AI Hub and the London Mathematical Society. ChatGPT 5.6 Sol was used to assist with proofreading.

\bibliographystyle{plainnat}
\bibliography{references}

\section{Proofs}\label{sec:proofs} 
After introducing preliminaries, we prove the main results.
\noindent Each set-valued mapping $\phi : \mathcal{X} \rightrightarrows \mathcal{Y}$ has two natural inverses.
\noindent The \textit{upper inverse} $\phi^{u}$ (also called the \textit{strong inverse}) of a subset $A$ of $\mathcal{Y}$ is defined by
    \[
        \phi^{u}(A) = \{ x \in \mathcal{X} : \phi(x) \subset A \}.
    \]
\noindent The \textit{lower inverse} $\phi^{l}$ (also called the \textit{weak inverse}) is defined by
    \[
        \phi^{l}(A) = \{ x \in \mathcal{X} : \phi(x) \cap A \neq \emptyset \}.
    \]
\noindent Recall that a neighborhood of a set \(A\) is any set \(B\) for which there is an open set \(V\) satisfying \(A \subset V \subset B\). Any open set \(V\) that satisfies \(A \subset V\) is called an open
neighborhood of \(A\).

\begin{definition}[{\cite[Definition 17.2]{guide2006infinite}}]\label{def:hemicont}
A set-valued mapping \(\phi : \mathcal{X} \rightrightarrows \mathcal{Y}\) between topological spaces is:

\noindent \emph{upper hemicontinuous} at \(x\) if for every neighborhood \(U\) of \(\phi(x)\), there is a neighborhood \(V\) of \(x\) such that \(z \in V\) implies \(\phi(z) \subset U\) or, equivalently, if \(\phi^u(U)\) is a neighborhood of \(x\) in \(\mathcal{X}\).

\noindent \emph{lower hemicontinuous} at \(x\) if for every open set \(U\) that meets \(\phi(x)\), i.e. \(\phi(x) \cap U \neq \emptyset\), there is a neighborhood \(V\) of \(x\) such that \(z \in V\) implies \(\phi(z) \cap U \neq \emptyset\), equivalently, the lower inverse image \(\phi^l(U)\) is a neighborhood of \(x\).

\noindent \emph{continuous} at \(x\) if it is both upper and lower hemicontinuous at \(x\).
\end{definition}
\noindent As with functions, we say \(\phi\) is upper or lower hemicontinuous on \(X\), if it is upper or lower hemicontinuous at every point of \(X\).  It is continuous if it is continuous at each point.
\begin{definition}[Regular conditional distribution]
\label{def:regconddist}
Let \(Z:\Omega\to S\) and \(Y:\Omega\to T\) be random variables, where
\(S\) and \(T\) are measurable spaces with \(\sigma\)-algebras
\(\mathcal B(S)\) and \(\mathcal B(T)\), respectively. A regular conditional
distribution of \(Y\) given \(Z\) is a Markov kernel
\[
K:S\times\mathcal B(T)\to[0,1]
\]
such that
\begin{enumerate}
    \item for every \(z\in S\), the map $A\mapsto K(z,A)$ is a probability measure on \((T,\mathcal B(T))\);
    \item for every \(A\in\mathcal B(T)\), the map $z\mapsto K(z,A)$ is \(\mathcal B(S)\)-measurable;

    \item for every \(A\in\mathcal B(T)\) and \(B\in\mathcal B(S)\),
    \[
    \mathbb P(Y\in A,\ Z\in B)
    =
    \int_B K(z,A)\,\mathbb P_Z(dz).
    \]
\end{enumerate}
\end{definition}

\noindent Regular conditional distributions do not always exist. The following positive result guarantees that a regular conditional distribution exists, see \cite{kallenberg1997foundations}.

\begin{theorem}[Existence of regular conditional distributions]
\label{thm:exitsregdist}
If $S$ and $T$ are standard Borel spaces, then a regular conditional distribution of $Y$ given $Z$ exists and is unique up to $\mathbb P_Z$-almost everywhere equality.
\end{theorem}

\subsection{Proof of Proposition \ref{prop:instabilityaccuracy}}

\begin{proof}
Since $d_{\mathcal Y}^H(\phi(x),\{y\})<\eta$, we choose $\delta>0$ such that $d_{\mathcal Y}^H(\phi(x),\{y\})+\delta<\eta$. The set
\[
U_\delta:=\{v\in\mathcal Y: \inf_{z\in\phi(x)}d_{\mathcal Y}(v,z)<\delta\}
\]
is an open neighborhood of $\phi(x)$. By upper hemicontinuity of $\phi$ at $x$, there exists a neighborhood $B_1 \subset \mathcal{M}_2$ of $x$ such that $\phi(\tilde x)\subset U_\delta$  for all $\tilde x\in B_1$. Hence
\[
\sup_{u\in\phi(\tilde x)}d_{\mathcal Y}(u,y)<\eta\qquad\text{for all }\tilde x\in B_1.
\]
Together with $\sup_{v\in\phi(x')}d_{\mathcal Y}(v,y'')<\eta$, the reverse triangle inequality gives, for every $\tilde x\in B_1$,
\begin{align*}
d_{\mathcal Y}^H(\phi(\tilde x),\phi(x'))\geq d_{\mathcal Y}(y,y'')-\sup_{u\in\phi(\tilde x)}d_{\mathcal Y}(u,y)-\sup_{v\in\phi(x')}d_{\mathcal Y}(v,y'') >d_{\mathcal{Y}}(y,y'')-2\eta.
\end{align*}
Since $d_{\mathcal X}(x,x')<\tau_{\mathcal X}$, there exists an open ball $B_2$ centered at $x$ such that $x'\notin B_2$ and $d_{\mathcal X}(\tilde x,x')<\tau_{\mathcal X}$ for all $\tilde x\in B_2$. Set $B:=B_1\cap B_2$, which is not empty as $x \in B$. Then, for every $\tilde x\in B$ and every $\epsilon\geq\tau_{\mathcal X}$, we have that $0<d_{\mathcal X}(\tilde x,x')<\tau_{\mathcal X}$, as $x'\notin B_2$ and $\tilde x \in B \subset B_2$ and $d_{\mathcal X}$ is a metric. Thus, the point $x'$ is admissible in the definition of $L^\epsilon(\phi,\tilde x)$. Therefore,
\begin{align*}
L^\epsilon(\phi,\tilde x)\geq\frac{d_{\mathcal Y}^H(\phi(\tilde x),\phi(x'))}
{d_{\mathcal X}(\tilde x,x')} >\frac{d_{\mathcal Y}(y,y'')-2\eta}{\tau_{\mathcal X}},
\end{align*}
which proves the result.
\end{proof}

\subsection{Proof of Theorem \ref{thm:lowerboundapprx}}

\begin{proof}
Let
\[
F_{x_k}^{N(k)}=\{y_{k,n}\}_{n=1}^{N(k)},\qquad k=1,\dots,K,
\]
be the finite feasible sets returned by Algorithm~\ref{alg:feasset}, and let
$\mathcal D_{M'}=\{(y_{k,n},x_k):1\le k\le K,\ 1\le n\le N(k)\}$ with $M'=\sum_{k=1}^K N(k)$. Let \(\phi:\mathcal M_2\rightrightarrows\mathcal Y\) be any regressor.

\noindent\textbf{Step 1: Lower bound by half of the average kernel size.}
Fix \(k\in\{1,\dots,K\}\) and \(1\le n<n'\le N(k)\). For any \(\bar z\in\phi(x_k)\), the triangle inequality gives
\[
d_{\mathcal Y}(y_{k,n},y_{k,n'})\le d_{\mathcal Y}(y_{k,n},\bar z)+d_{\mathcal Y}(\bar z,y_{k,n'}).
\]
Since
\[
d_{\mathcal Y}^H(\{y_{k,n}\},\phi(x_k))=\sup_{\bar z\in\phi(x_k)}d_{\mathcal Y}(y_{k,n},\bar z),
\]
we obtain
\[
d_{\mathcal Y}(y_{k,n},y_{k,n'})\le d_{\mathcal Y}^H(\{y_{k,n}\},\phi(x_k))+d_{\mathcal Y}^H(\{y_{k,n'}\},\phi(x_k)).
\]
For \(p\ge1\), the convexity of \(a\mapsto a^p\) implies $(a+b)^p\le2^{p-1}(a^p+b^p)$ for $a,b\ge0$. Consequently,
\[
d_{\mathcal Y}(y_{k,n},y_{k,n'})^p\le 2^{p-1}\left(d_{\mathcal Y}^H(\{y_{k,n}\},\phi(x_k))^p+d_{\mathcal Y}^H(\{y_{k,n'}\},\phi(x_k))^p\right).
\]
Summing over all unordered pairs \(1\le n<n'\le N(k)\), and observing that each
index \(n\) occurs in exactly \(N(k)-1\) such pairs, gives
\[
\frac{2\,\mathds{1}_{\{N(k)>1\}}}{N(k)-1}\sum_{1\le n<n'\le N(k)}d_{\mathcal Y}(y_{k,n},y_{k,n'})^p\le 2^p\sum_{n=1}^{N(k)} d_{\mathcal Y}^H(\{y_{k,n}\},\phi(x_k))^p.
\]
When \(N(k)\le1\), the left-hand side is zero by definition, so the inequality
remains valid. Summing over \(k=1,\dots,K\) and dividing by \(2^pM'\), we obtain
\[
\frac{1}{2^pM'} \sum_{k=1}^K \frac{2\,\mathds{1}_{\{N(k)>1\}}}{N(k)-1}\sum_{1\le n<n'\le N(k)}d_{\mathcal Y}(y_{k,n},y_{k,n'})^p\le\frac{1}{M'}\sum_{k=1}^K\sum_{n=1}^{N(k)}d_{\mathcal Y}^H(\{y_{k,n}\},\phi(x_k))^p.
\]
By the definition of the empirical loss and of the average kernel size,
this is
\[
\frac{1}{2^p}\operatorname{Kersize}(\mathcal D_{M'},p)_K^p\le\mathcal L(\mathcal D_{M'},\phi,p)^p.
\]
Taking \(p\)-th roots yields
\[
\frac12\operatorname{Kersize}(\mathcal D_{M'},p)_K\le\mathcal L(\mathcal D_{M'},\phi,p).
\]
Since \(\phi\) was arbitrary, the first claim follows.

\noindent\textbf{Step 2: Existence of an optimal regressor.}
For each \(k\in\{1,\dots,K\}\), define
\[
f_k(z):=\frac{1}{N(k)}\sum_{n=1}^{N(k)}d_{\mathcal Y}(y_{k,n},z)^p,\qquad z\in\mathcal Y.
\]
The function \(f_k\) is continuous. We next show that it attains its minimum. Let
\[
r_k:=\max_{1\le n,n'\le N(k)}d_{\mathcal Y}(y_{k,n},y_{k,n'}).
\]
Since the group is finite, \(r_k<\infty\). If $d_{\mathcal Y}(z,y_{k,1})>2r_k$, then, for every \(n\),
\[
d_{\mathcal Y}(z,y_{k,n})\ge d_{\mathcal Y}(z,y_{k,1})- d_{\mathcal Y}(y_{k,1},y_{k,n})>2r_k-r_k=r_k.
\]
Hence, $f_k(z)>r_k^p$. On the other hand,
\[
f_k(y_{k,1})=\frac{1}{N(k)}\sum_{n=1}^{N(k)}d_{\mathcal Y}(y_{k,n},y_{k,1})^p\le r_k^p.
\]
Therefore every minimizer of \(f_k\) lies in the closed ball $\overline B_{d_{\mathcal Y}}(y_{k,1},2r_k)$. This ball is closed and bounded, and hence compact by Assumption~\ref{as:heineBy}. Since \(f_k\) is continuous, it attains its minimum on this compact set. Thus, we may choose
\[
z_k\in\arg\min_{z\in\mathcal Y} f_k(z).
\]
\noindent\textbf{Step 3: Characterization of the unrestricted empirical optimal regressor.}
Fix \(k\). For any nonempty prediction set \(A\subset\mathcal Y\),
\[
d_{\mathcal Y}^H(\{y_{k,n}\},A) \ge \inf_{z\in A}d_{\mathcal Y}(y_{k,n},z)
\]
for each \(n\). More importantly, since
\[
d_{\mathcal Y}^H(\{y_{k,n}\},A)=\max\left\{\inf_{a\in A}d_{\mathcal Y}(y_{k,n},a),\sup_{a\in A}d_{\mathcal Y}(y_{k,n},a)\right\},
\]
we have, for every \(n\),
\[
d_{\mathcal Y}^H(\{y_{k,n}\},A)\ge d_{\mathcal Y}(y_{k,n},a)
\]
for every \(a\in A\). Consequently,
\[
\sum_{n=1}^{N(k)} d_{\mathcal Y}^H(\{y_{k,n}\},A)^p\ge \sum_{n=1}^{N(k)} d_{\mathcal Y}(y_{k,n},a)^p
\]
for every \(a\in A\), and therefore
\[
\sum_{n=1}^{N(k)}d_{\mathcal Y}^H(\{y_{k,n}\},A)^p\ge\inf_{z\in\mathcal Y}
\sum_{n=1}^{N(k)}d_{\mathcal Y}(y_{k,n},z)^p.
\]
By the definition of \(z_k\),
\[
\inf_{z\in\mathcal Y}\sum_{n=1}^{N(k)}d_{\mathcal Y}(y_{k,n},z)^p=\sum_{n=1}^{N(k)}d_{\mathcal Y}(y_{k,n},z_k)^p.
\]
Equality is attained by the singleton prediction set $A=\{z_k\}$. Since the anchor points \(x_1,\dots,x_K\) are pairwise positively separated, the values of a regressor at the different anchors can be prescribed independently.
Hence the unrestricted empirical optimal regressor attains
\[
R_*^p :=\inf_{\phi}\mathcal L(\mathcal D_{M'},\phi,p)^p=\frac{1}{M'}\sum_{k=1}^K\sum_{n=1}^{N(k)}d_{\mathcal Y}(y_{k,n},z_k)^p.
\]

\noindent\textbf{Step 4: Upper bound by the average kernel size.}
Fix \(k\) with \(N(k)\ge2\). By the minimizing property of \(z_k\), for every
\(j\in\{1,\dots,N(k)\}\),
\[
\sum_{n=1}^{N(k)}d_{\mathcal Y}(y_{k,n},z_k)^p\le\sum_{n=1}^{N(k)}d_{\mathcal Y}(y_{k,n},y_{k,j})^p.
\]
Averaging this inequality over \(j=1,\dots,N(k)\) gives
\[
\sum_{n=1}^{N(k)}d_{\mathcal Y}(y_{k,n},z_k)^p\le\frac{1}{N(k)}\sum_{j=1}^{N(k)}\sum_{n=1}^{N(k)}d_{\mathcal Y}(y_{k,n},y_{k,j})^p.
\]
The diagonal terms vanish, and symmetry of the metric implies
\[
\sum_{j=1}^{N(k)}\sum_{n=1}^{N(k)}d_{\mathcal Y}(y_{k,n},y_{k,j})^p
=2\sum_{1\le n<n'\le N(k)}d_{\mathcal Y}(y_{k,n},y_{k,n'})^p.
\]
Therefore,
\[
\sum_{n=1}^{N(k)}d_{\mathcal Y}(y_{k,n},z_k)^p\le\frac{2}{N(k)}\sum_{1\le n<n'\le N(k)}d_{\mathcal Y}(y_{k,n},y_{k,n'})^p.
\]
Since \(N(k)\ge2\), $\frac{2}{N(k)}\le\frac{2}{N(k)-1}$, and hence
\[
\sum_{n=1}^{N(k)}d_{\mathcal Y}(y_{k,n},z_k)^p\le\frac{2}{N(k)-1}\sum_{1\le n<n'\le N(k)}d_{\mathcal Y}(y_{k,n},y_{k,n'})^p.
\]
For \(N(k)=1\), the left-hand side is zero as \(z_k=y_{k,1}\) is an exact
minimizer, while the corresponding kernel contribution is zero by definition. Summing over \(k\) and dividing by \(M'\), we obtain
\[
R_*^p\le\frac{1}{M'}\sum_{k=1}^K\frac{2\,\mathds{1}_{\{N(k)>1\}}}{N(k)-1}\sum_{1\le n<n'\le N(k)}d_{\mathcal Y}(y_{k,n},y_{k,n'})^p.
\]
Thus, $R_*\le\operatorname{Kersize}(\mathcal D_{M'},p)_K$. Together with Step 1, this gives
\[
\frac12 \operatorname{Kersize}(\mathcal D_{M'},p)_K\le\inf_{\phi}\mathcal L(\mathcal D_{M'},\phi,p)\le\operatorname{Kersize}(\mathcal D_{M'},p)_K.
\]

\noindent\textbf{Step 5: Construction of an upper hemicontinuous minimizer.}
It remains to show that the unrestricted empirical optimum can be attained by an
upper hemicontinuous map with nonempty compact values. Assume first that \(K\ge2\), and define $r:=\frac13\min_{k\neq k'}d_{\mathcal X}(x_k,x_{k'})>0$. For \(k=1,\dots,K\), let $C_k:=\overline B_{d_{\mathcal X}}(x_k,r)$. For \(k\neq k'\), the triangle inequality implies $d_{\mathcal X}(C_k,C_{k'})\ge d_{\mathcal X}(x_k,x_{k'})-2r>0$. Hence the sets \(C_1,\dots,C_K\) are pairwise disjoint. If \(K=1\), choose any
\(r>0\) and define \(C_1\) in the same way. Choose an arbitrary \(z_0\in\mathcal Y\), and define
\[
\Phi(x)=
\begin{cases}
\{z_k\}, & x\in\operatorname{int}(C_k),\\
\{z_0,z_k\}, & x\in\partial C_k,\\
\{z_0\}, & x\notin\displaystyle\bigcup_{k=1}^K C_k.
\end{cases}
\]
Every value of \(\Phi\) is nonempty and finite, and hence compact. We claim that \(\Phi\) is upper hemicontinuous. Away from the boundaries \(\partial C_k\), the map is locally constant and is therefore upper hemicontinuous. Now let \(x\in\partial C_k\). Since the sets \(C_k\) are positively separated, there exists a neighborhood \(U\) of \(x\) that intersects no \(C_{k'}\) with \(k'\neq k\). Thus, for every \(x'\in U\), $\Phi(x') \in \bigl\{\{z_k\},\{z_0\},\{z_0,z_k\}\bigr\}$. Let \(V\subset\mathcal Y\) be any open set containing $\Phi(x)=\{z_0,z_k\}$. Then \(z_0,z_k\in V\), and consequently
\[
\{z_k\}\subset V,\qquad\{z_0\}\subset V,\qquad\{z_0,z_k\}\subset V.
\]
Hence, after possibly shrinking \(U\), $\Phi(x')\subset V$ for all $x'\in U$. This proves upper hemicontinuity at \(x\), and therefore \(\Phi\) is upper
hemicontinuous on \(\mathcal M_2\). Finally, since \(x_k\in\operatorname{int}(C_k)\), $\Phi(x_k)=\{z_k\}$ for $ k=1,\dots,K$. Therefore,
\[
\mathcal L(\mathcal D_{M'},\Phi,p)^p=\frac{1}{M'}\sum_{k=1}^K\sum_{n=1}^{N(k)}d_{\mathcal Y}(y_{k,n},z_k)^p=R_*^p.
\]
Thus the unrestricted empirical infimum is attained by an upper hemicontinuous map
with nonempty finite, hence compact, values. \(\Phi\) is singleton-valued on the open neighborhood \(\operatorname{int}(C_k)\) of every anchor \(x_k\). Consequently,
\[
\inf_{\phi_{\mathrm{uhc}}}\mathcal L(\mathcal D_{M'},\phi_{\mathrm{uhc}},p)=\inf_{\phi}\mathcal L(\mathcal D_{M'},\phi,p),
\]
and both infima lie between the lower and upper bound from steps $1$ and $4$. This completes the proof.
\end{proof}

\subsection{Proof of Proposition~\ref{prop:originalenlarged}}

\begin{proof}
For each edge \(e=(k,j)\in E\), define $a_e:=d_{\mathcal Y}^H(\{y_j\},\phi(x_k))$ and $b_e:=\ell_{r(e)}(\phi)$. By the choice of the reference index \(r(e)\), we have $|a_e-b_e|\le \gamma_e$. Since \(M'=|E|\),
\begin{align*}
\mathcal L(\mathcal D_{M'},\phi,p)
=
\left(\frac{1}{M'}\sum_{e\in E}a_e^p\right)^{1/p}.
\end{align*}
By the reverse triangle inequality for the \(\ell_p\)-norm,
\begin{align*}
\left|
\mathcal L(\mathcal D_{M'},\phi,p)
-
\left(\frac{1}{M'}\sum_{e\in E}b_e^p\right)^{1/p}
\right|
\le
\left(\frac{1}{M'}\sum_{e\in E}|a_e-b_e|^p\right)^{1/p}
\le
\left(\frac{1}{M'}\sum_{e\in E}\gamma_e^p\right)^{1/p}=\Gamma_p(\phi).
\end{align*}
Next, by the definition of the reference weights \(w_i\),
\begin{align*}
\frac{1}{M'}\sum_{e\in E}b_e^p=\frac{1}{M'}\sum_{e\in E}\ell_{r(e)}(\phi)^p=\sum_{m=1}^M w_m\ell_m(\phi)^p.
\end{align*}
Hence, using \(0\le \ell_m(\phi)\le B_\phi\) for all $m \in \{1, \ldots, M\}$,
\begin{align*}
\left|\sum_{m=1}^M w_m\ell_m(\phi)^p-\frac{1}{M}\sum_{m=1}^M\ell_m(\phi)^p\right|
&=\left|\sum_{m=1}^M\left(w_m-\frac1M\right)\ell_m(\phi)^p\right|\\
&\le\sum_{m=1}^M\left|w_m-\frac1M\right|\ell_m(\phi)^p\le B_\phi^p\sum_{m=1}^M\left|w_m-\frac1M\right|.
\end{align*}
For nonnegative \(s,t\) and \(p\ge1\), the map \(u\mapsto u^{1/p}\) satisfies $|s^{1/p}-t^{1/p}|\le|s-t|^{1/p}$. Therefore,
\begin{align*}
\left|\left(\sum_{m=1}^M w_m\ell_m(\phi)^p\right)^{1/p}-\left(\frac1M\sum_{m=1}^M\ell_m(\phi)^p\right)^{1/p}\right|&\le B_\phi
\left(\sum_{m=1}^M\left|w_m-\frac1M\right|\right)^{1/p}.
\end{align*}
Combining the preceding two inequalities with the triangle inequality gives the claimed bound
\begin{align*}
\left|\mathcal L(\mathcal D_{M'},\phi,p)-\mathcal L(\mathcal D_M,\phi,p)\right|
\le\Gamma_p(\phi)+ B_\phi\left(\sum_{m=1}^M\left|w_m-\frac1M\right|
\right)^{1/p}
\end{align*}
\end{proof}

\subsection{Proof of Proposition \ref{prop:bayes-kernel-identity}}

\begin{proof}
Conditional on $\overline X$, let $Y'$ be an independent copy of $Y$. Then $\mathbb E[\|Y-Y'\|_2^2\mid\overline X]
=
2\,\mathbb E[\|Y-m(\overline X)\|_2^2\mid\overline X]$. The conditional Pythagorean identity gives
$\mathbb E[\|Y-\overline\phi(\overline X)\|_2^2\mid\overline X]
=
\mathbb E[\|Y-m(\overline X)\|_2^2\mid\overline X]
+
\|m(\overline X)-\overline\phi(\overline X)\|_2^2$. Taking expectations proves the result.
\end{proof}

\subsection{Proof of Proposition \ref{thm:gengapbd}}

\begin{proof}
\textbf{Regression generalization gap:} Let $\overline \phi:\overline{\mathcal M}_2\to \mathcal Y$ be a measurable regressor such that \(\mathcal L(\mathcal D_\mu,\overline\phi,p)<\infty\). Fix $\overline X$. Let $Y,Y' \overset{i.i.d.}{\sim} \mu(\cdot\mid \overline X)$. By the triangle inequality, $d_{\mathcal Y}(Y,Y')
\le d_{\mathcal Y}(Y,\overline\phi(\overline X)) + d_{\mathcal Y}(\overline\phi(\overline X),Y')$. Applying $(a+b)^p \le 2^{p-1}(a^p+b^p)$ gives $d_{\mathcal Y}(Y,Y')^p
\le
2^{p-1}\big(
d_{\mathcal Y}(Y,\overline\phi(\overline X))^p
+
d_{\mathcal Y}(Y',\overline\phi (\overline X))^p
\big)$. Now taking the conditional expectation given $\overline X$, gives 
\[
\mathbb E[d_{\mathcal Y}(Y,Y')^p \mid \overline X]
\le
2^{p-1}\Big(
\mathbb E[d_{\mathcal Y}(Y,\overline \phi(\overline X))^p \mid \overline X]
+
\mathbb E[d_{\mathcal Y}(Y',\overline \phi(\overline X))^p \mid \overline X]
\Big).
\]
Since $Y$ and $Y'$ are i.i.d. conditional on $\overline X$,
$\mathbb E[d_{\mathcal Y}(Y',\overline \phi(\overline X))^p \mid\overline X]
=
\mathbb E[d_{\mathcal Y}(Y,\overline\phi(\overline X))^p \mid \overline X]$. Hence,
$2^{-p}\mathbb E[d_{\mathcal Y}(Y,Y')^p \mid\overline  X]
\le
\mathbb E[d_{\mathcal Y}(Y,\overline \phi(\overline X))^p \mid \overline X]$. Taking the expectation over $\overline X$ and using the tower property on the right hand side, yields 
\[
2^{-p}
\mathbb E_{\overline X} \mathbb E_{Y,Y'\mid \overline X}[d_{\mathcal Y}(Y,Y')^p] \le \mathbb E_{(Y,\overline X)}[d_{\mathcal Y}(\overline \phi(\overline X),Y)^p].
\]
Now taking the $1/p$-th power on both sides yields
\[
2^{-1}
\left(\mathbb E_{\overline X} \mathbb E_{Y,Y'\mid \overline X}[d_{\mathcal Y}(Y,Y')^p]\right)^{1/p} \le \mathbb E_{(Y,\overline X)}[d_{\mathcal Y}(\overline \phi(\overline X),Y)^p]^{1/p}
\]
Thus, as $\operatorname{Kersize}(\mathcal{D}_{\mu},p) :=\Big(\mathbb E_{\overline X} \mathbb E_{Y,Y'\mid \overline X}[d_{\mathcal Y}(Y,Y')^p]\Big)^{1/p}$ we have that $2^{-1}\operatorname{Kersize}(\mathcal{D}_{\mu},p)\le \mathcal L(\mathcal{D}_{\mu},\overline\phi,p)$. Finally, since empirical risk is bounded by $\epsilon$, we have that $- \epsilon \le - \mathcal L(\mathcal{D}_M,\phi,p)$, and, thus $2^{-1}\operatorname{Kersize}(\mathcal{D}_{\mu},p) - \epsilon \le \mathcal L(\mathcal{D}_{\mu},\overline\phi,p)-\mathcal L(\mathcal{D}_M,\phi,p)$. The Euclidean sharpening follows from Proposition \ref{prop:bayes-kernel-identity}.
\textbf{Classification generalization gap:} The proof follows the same structure as for the regression case by bounding $\operatorname{Acc}(\mathcal{D}_{\mu},\overline\phi) := 1 - \mathbb E_{(Y,\overline X)\sim\mu}[d_{H}(Y,\overline\phi(\overline X))]$ from above.
\end{proof}

\subsection{Proof of Proposition \ref{thm:convavgalgnon}}

\begin{proof}
For each $k$, let $U_k:=\frac{2}{N_K(N_K-1)}
\sum_{1\le n<n'\le N_K}
 d_{\mathcal Y}(Y_{k,n},Y_{k,n'})^p$. Conditional on $\overline X_k$, the order-two U-statistic is unbiased, so
$\mathbb E[U_k\mid\overline X_k]
=
\mathbb E[d_{\mathcal Y}(Y,Y')^p\mid\overline X_k]$. Taking expectations gives
$\mathbb E U_k=\operatorname{Kersize}(\mathcal D_\mu,p)^p$. The variables $U_1,\ldots,U_K$ are independent and satisfy
$0\le U_k\le v_{\max}$. Since
$\widehat\kappa_{K,N_K}^p=K^{-1}\sum_{k=1}^K U_k$, Hoeffding's inequality \cite{hoeffding1994probability} gives
the stated bound.
\end{proof}

\newpage 
\section{Appendix}

\subsection{Experiment details}\label{sec:expdet}

\subsubsection{Pseudo-metric on image data}
\label{subsec:common_image_metric}

Across all image-based tasks we used the same metric to ensure comparability of image-based features. The metric on images was the composition of the $\ell_2$ norm with discrete wavelet features $g$. The image feature was obtained by the concatenation of the four subbands obtained from a single-level two-dimensional discrete wavelet transform (DWT) applied independently to each channel \cite{daubechies1992ten}. Given an image \(x\in\mathbb{R}^{C\times H\times W}\), the resulting feature vector is
denoted by \(g_{\mathrm{wav}}(x)\in\mathbb{R}^{D}\). The DWT was computed
using a chosen wavelet basis; in all experiments, we used the Haar wavelet
basis. Specifically, for each channel, the transform decomposes the image
into an approximation subband, denoted by LL (low-low), and three detail
subbands: LH (low-high), HL (high-low), and HH (high-high). These subbands correspond to the responses of low- and high-frequency filters
applied along the two spatial dimensions. The four resulting subbands were
flattened and concatenated to form the final feature, $g_{\mathrm{wav}}(x)$. For
wavelets with standard dyadic downsampling and even spatial dimensions, the resulting feature dimension is \(D=C H W\). The image similarity was measured by the metric
\begin{align}\label{eq:imagepseudomet}
d_{\mathcal X}(x,x')=\left\|g(x)-g(x')\right\|_2.
\end{align}
More generally, this construction can be viewed as a feature-induced pseudo-metric on the original image space: if the feature map $g$ is not injective, distinct images may have identical feature representations and therefore zero distance. For the specific implementation considered here, all four subbands of the Haar DWT are retained, so the transform is invertible under the corresponding DWT conventions. The pseudo-metric terminology is therefore primarily relevant to the general feature-map formulation, particularly when $g$ is replaced by a potentially non-injective feature encoder.

\subsubsection{Feasible set construction}\label{sec:feassetsamp}

We sampled finite feasible sets using Algorithm~\ref{alg:feasset}. For
image-based experiments, neighboring inputs were identified using the
wavelet-induced distance \(d_{\mathcal X}\) in
\eqref{eq:imagepseudomet}. Threshold choice followed a fixed rule throughout the experiments; whenever prior knowledge supplied an input-noise model and/or a target/label measurement-noise model, we set the thresholds $\tau_{\mathcal X}$ and $\tau_{\mathcal Y}$ to the corresponding conservative worst-case noise bounds. If such prior information was unavailable, we estimated $\tau_{\mathcal X}$ using the nearest-neighbor quantile rule in \eqref{eq:tauxqrule} and set $\tau_{\mathcal Y}=0$ unless target-noise information was available. Given the thresholds \(\tau_{\mathcal X}\) and
\(\tau_{\mathcal Y}\) approximately satisfying Assumption~\ref{as:levelstauxy}, and an
anchor \((y_k,x_k)\in\mathcal D_M\), the target \(y_{k,n}\in\pi_1(\mathcal D_M)\)
of a sample \((y_{k,n},x_{k,n})\in\mathcal D_M\) was admitted to the feasible
set \(F_{x_k}^{N(k)}\) whenever $d_{\mathcal X}(x_{k,n},x_k)\le \tau_{\mathcal X}(x_k)$ or $d_{\mathcal Y}(y_{k,n},y_k)\le \tau_{\mathcal Y}(y_k)$. Thus, \(F_{x_k}^{N(k)}\) contains targets associated with samples that are
close to the anchor either in input space or in target space. For one-hot labels, we set \(\tau_{\mathcal Y}=0\). In the case that prior knowledge did not supply an input-noise model, we used the following dataset-dependent procedure for the input threshold
\begin{align}\label{eq:tauxqrule}
\tau_{\mathcal X}(q)
=
\operatorname{Quantile}_{q}
\left(
\left\{
d_{\mathcal X}(x_m,x_{(1)}^{(m)})
\right\}_{m=1}^{M}
\right),
\end{align}
where \(x_{(1)}^{(m)}\) denotes the closest non-self neighbor of \(x_m\). The nearest neighbors for the dataset-dependent input threshold selection were computed with the nearest neighbor algorithm from \cite{pedregosa2011scikit}.\footnote{If the target noise is unknown, one could use the analogous target-space nearest-neighbor quantiles, $\tau_{\mathcal Y}(q)
=\operatorname{Quantile}_{q}(\{d_{\mathcal Y}(y_m,y_{(1)}^{(m)})\}_{1\le m \le M})$.} We used the standard settings, the metric corresponding to the choice of $d_{\mathcal{X}}$ and reported the maximum number of nearest neighbors $N_{\max}$ used in nearest neighbor algorithm from \cite{pedregosa2011scikit} in the experiments.\footnote{Note that the maximum number of nearest neighbors $N_{\max}$ for computing the threshold $\tau_{\mathcal X}(q)$ for a given quantile $q$ is not equal to the number of maximal elements $N^f_{\max}$ in a given feasible set obtained by Algorithm~\ref{alg:feasset}.} Additionally, we chose the maximum number of nearest neighbors $N_{\max}$ such that the average kernel size converged with an ablation study outlined below. If the worst-case noise threshold was unknown, in our experiments we did not estimate $\tau_{\mathcal Y}$ via quantiles and set $\tau_{\mathcal Y} = 0$, thereby assuming noiseless targets.

\noindent The quantile parameter \(q\) for choosing the threshold $\tau_{\mathcal{X}}(q)$ in \eqref{eq:tauxqrule} determines the locality scale at which samples are treated as indistinguishable with respect to the pseudo-metric
\(d_{\mathcal X}\): smaller values of \(q\) enforce more local feasible sets,
whereas larger values increase sample support but may merge points that are no
longer similar.  To focus on local neighborhoods we restricted the quantile to $q\in[0.01,0.1]$, that is, to the lower tail of the nearest-neighbor distance distribution. Within this range, \(q\) was chosen by a \emph{bandwidth stability heuristic} rather than by an optimality criterion: we selected the smallest \(q\) lying in a region where the estimated average kernel size changed only mildly across neighboring values of \(q\). We identified such a region by requiring both small relative changes in the average kernel size and approximately small local curvature, and then choose the smallest \(q\) in that plateau. This favors the most local scale at which the estimated ambiguity is not unduly sensitive to the threshold choice. We emphasize that this is a practical empirical heuristic for selecting a stable locality scale, not a claim of statistical optimality. To make the dependence on \(q\) transparent, we reported full sensitivity curves in Section \ref{sec:ablationstudies} below, which contains additional ablation studies, and verified that the average kernel size remained qualitatively stable throughout the selected range.

\noindent  The values of \(q\), \(\tau_{\mathcal X}\), and \(\tau_{\mathcal Y}\) and the average feasible set size $\overline{N(k)}$, as well as $M'$ and the number of anchor samples $K$, are reported for each experiment.

\noindent\textbf{Ablation study on maximum number of nearest neighbors and quantiles:} We assessed sensitivity to the quantile $q$ and maximum number of nearest neighbors $N_{\max}$ used in nearest neighbor algorithm from \cite{pedregosa2011scikit} for computing the threshold $\tau_{\mathcal X}(q)$ and varied
$q\in\{0.01,0.02,0.03,0.05,0.06, 0.08, 0.1\}$ and
$N_{\max}\in\{10,20,30,40,50,100,200,300,450\}$. In the ablation over the neighborhood size we selected $N_{\max}$ such that the average kernel size remained constant for larger $N_{\max}$. For varying the neighborhood size we set a convergence criterion at $10^{-4}$. For completeness, we extended the ablation study on quantiles to 
\[
q\in\{0.01,0.02,0.03,0.05,0.06, 0.08, 0.1,0.2,0.3,0.4,0.5,0.6,0.7,0.8,0.9,1.0\},
\]
\noindent although we do not recommend extending beyond the quantile $q=0.1$, as the resulting threshold $\tau_{\mathcal X}(q)$ is unlikely to satisfy Assumption \ref{as:levelstauxy} for larger quantiles. For each parameter choice, we recomputed the nearest-neighbor graph,
threshold \(\tau_{\mathcal X}\), feasible sets, and average kernel size.
This provides an empirical assessment of the stability of the resulting
feasible-set geometry and is reported in Section \ref{sec:ablationstudies}.

\subsubsection{Regression dataset details}\label{sec:regdets}

\begin{figure}[ht]
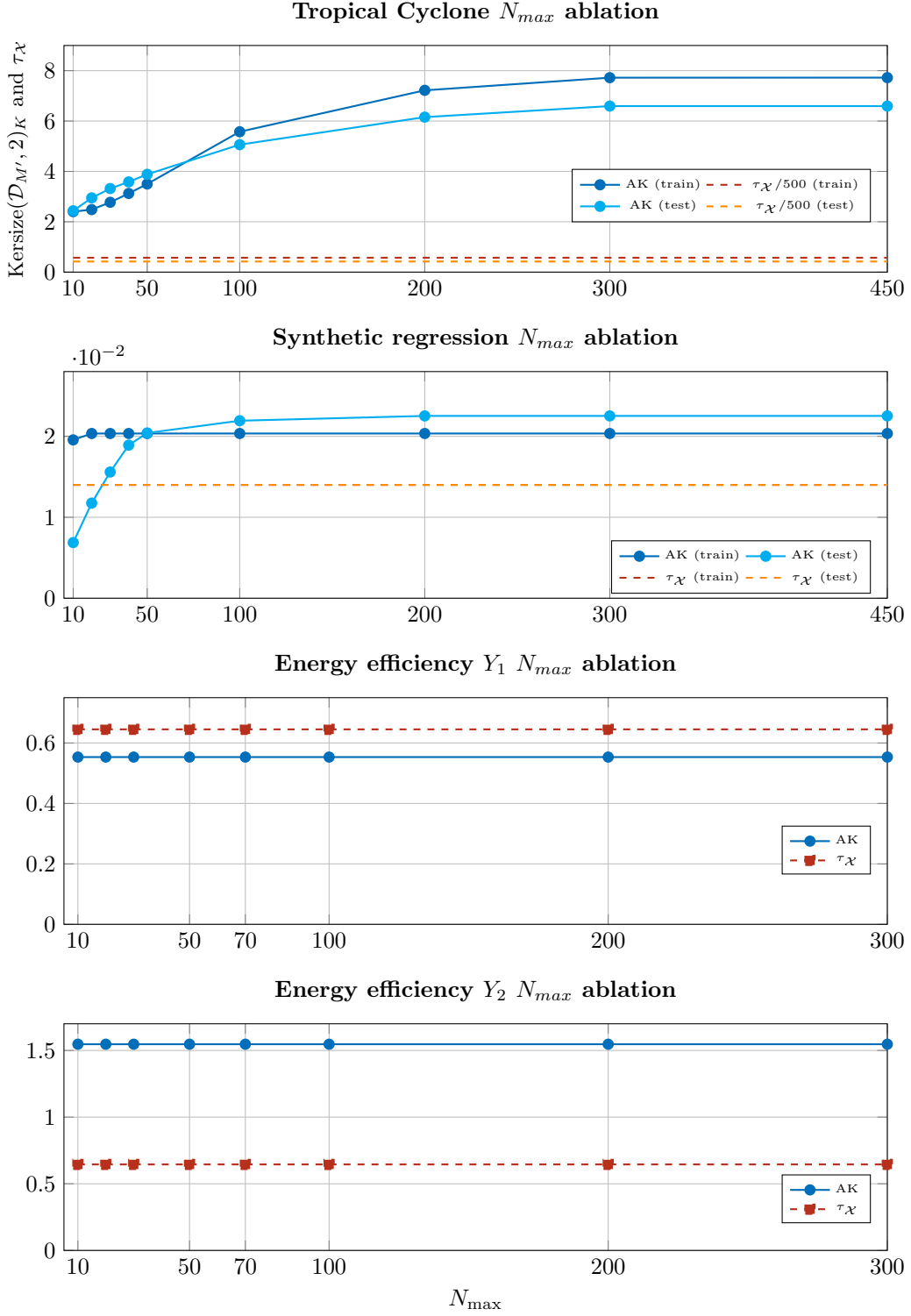

    \centering
    %
    \caption{Average kernel size, $AK$ or also $\operatorname{Kersize}(\mathcal D_{M'},2)_{K}$, for regression datasets as a function of the maximum number of nearest neighbors $N_{max}$ used in the threshold $\tau_{\mathcal{X}}$ and feasible set computation. For synthetic regression the thresholds are chosen based on the worst-case noise and otherwise, the input threshold is chosen as the $q^{th}$ quantile of the nearest-neighbor distances according to \eqref{eq:tauxqrule}. The quantile $q$ and number of anchor samples $K$ for train and test splits can be found in Table \ref{tab:regression-parameters}.}
\label{fig:regression-kersize-nmax-ablation1}
\end{figure}

\noindent For the \textbf{synthetic regression} dataset, we computed the average kernel size from feasible sets as a function of the maximum number of nearest neighbors and reached convergence with threshold $10^{-4}$ at $N_{\max} =300$, see Figure \ref{fig:regression-kersize-nmax-ablation1}. We generated a mesh grid of $M=900$ noisy input points $(x_1,x_2) \in [-1,1]^2$ and a noisy surface from the regression target function $f(x_1,x_2) =\sin\!\left(1+0.9\sqrt{x_1^2+x_2^2}\right)$ evaluated on the mesh grid of input points, where we added uniform input noise $U([0,0.01]^2)$ and target noise $U([0,0.03])$. Because bounded input and target noise were prescribed in the data-generation process, we set $\tau_{\mathcal X}$ and $\tau_{\mathcal Y}$ equal to the corresponding worst-case noise magnitudes. Thus, we chose $\tau_{\mathcal{X}}= 0.014 = \sqrt{0.01^2+0.01^2}$ and $\tau_{\mathcal{Y}}= 0.03$. We trained three regressors to approximate the target function $f$ from samples in a train dataset $\mathcal{D}_{M_{\mathrm{train}}}$ and evaluated their predictive error on held-out test data. We considered a Gaussian Process regressor $\phi_{1st}$, a multilayer perceptron (MLP) regressor $\phi_{2nd}$, and a Support Vector Regression (SVR) model with a radial basis function (RBF) kernel, denoted by $\phi_{3rd}$. We report the RMSE of each model on both the original test dataset $\mathcal{D}_{M_{\mathrm{test}}}$ and the constructed test dataset $\mathcal{D}_{M'_{\mathrm{test}}}$ generated by Algorithm~\ref{alg:feasset}. The error of the optimal regressor on $\mathcal D_{M'_{\mathrm{test}}}$ is given by $R_*= 0.01656$. \newline
As a noiseless control, we repeated the feasible-set construction on data generated from the deterministic relation $Y=f(X)$, for which the Bayes RMSE is zero. For all tested positive input thresholds with $\tau_{\mathcal Y}=0$, the empirical average kernel size was also zero on both train and test splits. Hence, in this setting the geometric threshold construction did not itself induce target ambiguity. \newline
For a comparison to the Bayes risk, we assume the additive noise model
\[
Y = f(X) + \varepsilon,
\qquad
\varepsilon \sim U([0,0.03]),
\]
with \(\varepsilon\) independent of \(X\). Hence $\operatorname{Var}(Y \mid X) = \operatorname{Var}(\varepsilon) = \tfrac{0.03^2}{12}$, so the Bayes RMSE is $\frac{0.03}{\sqrt{12}} \approx 0.00866$. By Proposition~\ref{prop:bayes-kernel-identity}, the corresponding population comparison quantity is $\tfrac{1}{\sqrt{2}} \operatorname{Kersize}(\mathcal D_\mu,2)$. As we do not observe \(\mathcal D_\mu\), we compute the empirical proxy $\tfrac{1}{\sqrt{2}} \operatorname{Kersize}(\mathcal D_{M'_{\mathrm{test}}},2)_K$. For \(\tau_{\mathcal X}\in\{10^{-5},0.01,0.04,0.1\}\), the empirical results are identical, indicating that the input-threshold criterion is inactive in this regime. The corresponding test values of the average kernel size are
\[
0.007509,\quad 0.011363,\quad 0.015147,\quad 0.022562
\]
for corresponding values of \(\tau_{\mathcal Y}=0.01,0.015,0.02,0.03\), respectively. Comparing with the Bayes benchmark under the assumption that the noise is independently sampled of the input, \(0.00866\), we find that for \(\tau_{\mathcal Y}=0.01\) the average kernel size the closest, whereas larger output thresholds increasingly overestimate the irreducible error under this independence assumption. Under target threshold misspecification for \(\tau_{\mathcal Y}=0.01,0.015,0.02,0.03\), we tested Assumption~\ref{as:levelstauxy} as follows: a candidate reassignment \((j,k)\) is called admissible if the observed target value \(y_j\) lies in the model-consistent interval associated with the anchor input \(X_k\), namely
\[
f(X_k) \le y_j \le f(X_k)+0.03.
\]
Accordingly, the test precision is defined as
\[
\mathrm{Precision}_{\mathrm{test}}
=
\frac{\#\{\text{accepted test pairs }(j,k)\text{ that are admissible}\}}
{\#\{\text{accepted test pairs }(j,k)\}}.
\]
Its empirical value decreases from
\[
0.8760 \quad \text{to} \quad 0.8082,\ 0.7314,\ 0.5700
\]
as \(\tau_{\mathcal Y}\) increases. \newline

\noindent For the \textbf{Tropical Cyclone} open-source dataset from \href{https://huggingface.co/datasets/TCBench/TCBench}{Huggingface}, the input images were converted to tensors. We did not assume an explicit image-space noise model and therefore chose $\tau_{\mathcal X}$ with the quantile heuristic. For the wind-speed targets, we used a heteroskedastic measurement-noise model outlined below and set $\tau_{\mathcal Y}(y)$ accordingly. We computed the average kernel size from feasible sets as a function of the maximum number of nearest neighbors and reached convergence with threshold $10^{-4}$ at $N_{\max} =300$, see Figure \ref{fig:regression-kersize-nmax-ablation1}. For the ablation over quantiles see Figure \ref{fig:regression-kersize-quantile-ablation1}. For the tropical cyclone regression tasks in Table \ref{tab:regression-results}, the maximum sustained wind speed is predicted from a triplet of consecutive satellite images or from the storm category (SC) defined by the Saffir-Simpson scale \cite{simpson1974hurricane}. We report the optimal accuracy intervals from Theorem~\ref{thm:lowerboundapprx} mean over $20$ random subsets of size $K=1k$ sampled from the train and test sets,  $\overline{\operatorname{Kersize}}_{\mathrm{train}}=7.65\pm 0.19$, $\overline{\operatorname{Kersize}}_{\mathrm{test}}=6.76\pm 0.15$. The RMSE values, in knots, of the leaderboard winning models of \cite{drivendata_wind_winners_2021} were computed on the test dataset. The dataset consists of single-channel long-wave infrared satellite images acquired every 15 minutes, paired with one-minute maximum sustained wind speeds in knots. Using the challenge’s storm-level splits and \texttt{TorchGeo} data loader \cite{stewart2025torchgeo}, we obtained approximately $70\mathrm{k}$ training and $44\mathrm{k}$ test samples. We resized images to $366\times 366$ pixels and standardized the wavelet features on the train and test splits. The wind-speed annotations are from \cite{nasa_tropical_cyclone_wind_estimation_2024}. As these annotations are operational intensity estimates rather than direct point measurements, we modeled their noise as heteroskedastic, with uncertainty increasing with wind speed. This is motivated by previous studies of tropical-cyclone intensity and wind-speed retrieval uncertainty \cite{torn2012uncertainty,ruf2018assessment}, including approximately $1.4~\mathrm{m\,s^{-1}}$ low-wind and $17.2\%$ high-wind uncertainties \cite{ruf2018assessment}. With wind speed $y$ measured in knots, we used the phenomenological uncertainty model
$$\tau_{\mathcal{Y}}(y)=
\begin{cases}
2.7, & y<39,\\
0.17y, & y\geq 39,
\end{cases}$$ where $39$ knots approximates $20~\mathrm{m\,s^{-1}}$, $2.7$ knots corresponds to approximately $1.39~\mathrm{m\,s^{-1}}$, and the high-wind branch represents approximately $17\%$ relative uncertainty. These values motivated our label-noise model but are not intended as a direct recalibration of the uncertainty estimates. For $\tau_{\mathcal{X}}=0.0$ and the same heteroskedastic noise model for $\tau_{\mathcal{Y}}$ with otherwise the same parameters, we have that $\operatorname{Kersize}(\mathcal D_{M'_{\mathrm{test}}},2)_K =5.91$ kts. For the same input threshold and $\tau_{\mathcal{Y}}=0.01$ with otherwise the same parameters, we have that $\operatorname{Kersize}(\mathcal D_{M'_{\mathrm{test}}},2)_K =2.36$ kts. The error of the optimal regressor on $\mathcal D_{M'_{\mathrm{test}}}$ is given by $R_*= 4.3215\pm 0.1095$.  

\noindent As a sanity check, we considered the \textbf{Tropical Cyclone Storm Categories (SC)} dataset, we used only the labels from the tropical cyclone open-source dataset from \href{https://huggingface.co/datasets/TCBench/TCBench}{Huggingface} and consider the task of predicting the wind speed from the hurricane and storm categories of the Saffir-Simpson scale \cite{simpson1974hurricane}. We computed the average kernel size from the input of the storm category and bins containing the associated wind speeds in the train and test datasets. We assumed noiseless targets and set $\tau_{\mathcal{Y}}=0$. Here we obtained accuracy bounds on the train dataset of $[3.34,6.68]$ and $[3.23,6.45]$ on the test dataset. Note that these intervals are in the same order of magnitude as the ones computed on the tropical cyclone dataset reported in Table \ref{tab:regression-results}.

\noindent For \textbf{UCI Energy Efficiency} open-source datasets \cite{tsanas2012accurate} from \href{https://archive.ics.uci.edu/dataset/242/energy+efficiency}{UCI} with targets $Y_1$ and $Y_2$, we standardized the input variables using the full set. No measurement-noise models were available, so $\tau_{\mathcal X}$ was selected by the quantile-based rule and $\tau_{\mathcal Y}=0$. We computed the average kernel size from feasible sets as a function of the maximum number of nearest neighbors and reached convergence with threshold $10^{-4}$ at $N_{\max} =100$, see Figure \ref{fig:regression-kersize-nmax-ablation1}. For the ablation over quantiles see Figure \ref{fig:regression-kersize-quantile-ablation1}. This dataset \cite{tsanas2012accurate} contains eight input variables (relative compactness, surface area, wall area, roof area, overall height, orientation, glazing area, glazing area distribution) and two target variables, the heating load ($Y_1$) and cooling load ($Y_2$). The best model errors in RMSE were from \cite[Table 5]{vzegklitz2017symbolic}. We assumed noiseless targets and set $\tau_{\mathcal{Y}}=0$, which means that the target criterion was inactive in Algorithm~\ref{alg:feasset}. The error of the optimal regressor on $\mathcal D_{M'_{\mathrm{test}}}$ is given by $R_*= 0.2767$ for $Y_1$ and by $R_*= 0.7734$ for $Y_2$.

\noindent For the \textbf{UCI Concrete Compressive Strength} open-source dataset \cite{yeh1998modeling} from \href{https://archive.ics.uci.edu/dataset/165/concrete+compressive+strength}{UCI}, we standardized the input variables using the full dataset. No measurement-noise models were available, so $\tau_{\mathcal X}$ was selected by the quantile-based rule and $\tau_{\mathcal Y}=0$. We computed the average kernel size from feasible sets as a function of the maximum number of nearest neighbors and reached convergence with threshold $10^{-4}$ at $N_{\max} =50$, see Figure \ref{fig:regression-kersize-nmax-ablation2}. For the ablation over quantiles see Figure \ref{fig:regression-kersize-quantile-ablation1}. For this dataset \cite{yeh1998modeling} there are eight input variables (cement, blast furnace slag, fly ash, water, superplasticizer, coarse aggregate, fine aggregate, and age) and one target variable, the concrete compressive strength. The best model errors in RMSE were from the benchmark results reported in \cite[Table 10]{pan2024predicting}. The error of the optimal regressor on $\mathcal D_{M'_{\mathrm{test}}}$ is given by $R_*= 1.7136$.  

\noindent For the \textbf{KAERI Critical Heat Flux (CHF)} open-source dataset with uniform axial heating~\cite{park2000effect} from \href{https://www.kaggle.com/datasets/kaustubhdhote/critical-heat-flux-dataset/data}{Kaggle}, we standardized the input variables using the full set. We computed the average kernel size from feasible sets as a function of the maximum number of nearest neighbors and reached convergence with threshold $10^{-4}$ at $N_{\max} =50$, see Figure \ref{fig:regression-kersize-nmax-ablation2}. We used five input variables (diameter, heated length, pressure, mass flux, and equilibrium quality) and one target variable, the critical heat flux. We set $\tau_{\mathcal Y}$ equal to the worst-case noise magnitude of a heteroskedastic noise model. To account for input measurement uncertainty, we assigned uncertainties of $\pm1\%$ for diameter and length, $\pm0.5\%$ for pressure, $\pm2\%$ for mass flux, and $\pm0.01$ absolute for equilibrium quality. For observation $k$, the resulting input uncertainty radius was computed using\,
$\tau_{\mathcal{X},k}=\left(\sum_i\left(\sigma_{i,k}/s_i\right)^2\right)^{1/2}$, where $\sigma_{i,k}$ denotes the assumed physical measurement uncertainty of feature $i$ for observation $k$, and $s_i$ denotes the standard deviation of feature $i$ computed on the full dataset. We used the dimensionless worst-case threshold $\tau_{\mathcal{X}}= \max_k \tau_{\mathcal{X},k}=0.171840$. We modeled target measurement uncertainty as heteroskedastic, where for a critical heat flux measurement $y$ in $\mathrm{kW/m^2}$, we define
\[
\tau_{\mathcal{Y}}(y)=\sqrt{\sigma_{\mathrm{floor}}^2+\left(r y\right)^2},
\]
where $\sigma_{\mathrm{floor}}=1\mathrm{kW/m^2}$ represents a constant absolute uncertainty and $r=0.1$ represents a $10\%$ relative uncertainty between targets as reported in \cite[Section 2.4]{groeneveld2019}. This is an approximate heteroskedastic measurement model and should not be interpreted as independently verified measurement uncertainties. The best model errors were from Extra Trees, Random Forest, and Histogram Gradient Boosting regressors trained on $M_{\mathrm{train}}=521$ samples and tested on $M_{\mathrm{test}}=130$ samples. As we compute the average kernel size on the full dataset we do not obtain a constructed dataset solely for the test dataset and, thus, do not report the models' errors on this. For $\tau_{\mathcal{X}}=0.0$ and the same heteroskedastic noise model for $\tau_{\mathcal{Y}}$ with otherwise the same parameters, we have that $\operatorname{Kersize}(\mathcal D_{M'},2)_M =191.9\mathrm{kW/m^2}$. For $\tau_{\mathcal{X}}=0.171840$ and $\tau_{\mathcal{Y}}=0.01$ with otherwise the same parameters, we have that $\operatorname{Kersize}(\mathcal D_{M'},2)_M =341.4\mathrm{kW/m^2}$. The larger average kernel size for $\tau_{\mathcal{Y}}=0.01$ may be due to aggregation pathologies. The error of the optimal regressor on $\mathcal D_{M'}$ is given by $R_*=  149.574\mathrm{kW/m^2}$.

\subsubsection{Classification dataset details}\label{sec:classdets}

\begin{figure}[ht]
    \centering
    \begin{tikzpicture}

        \begin{groupplot}[
            group style={
                group size=1 by 4,
                horizontal sep=1.0cm,
                vertical sep=1.5cm
            },
]

    \nextgroupplot[
    width=14cm,
    height=5cm,
    title={\textbf{CIFAR-10 $N_{max}$ ablation}},
    ylabel={\small{$\operatorname{Kersize}(\mathcal D_{M'},1)_{K}$ and $\tau_{\mathcal{X}}$}},
    grid=both,
    legend style={
    at={(0.7,0.75)},
    anchor=north,
    font=\tiny,
    legend columns=2,
    },
    xtick={10,50,100,200,300,450},
    xmin=5,xmax=450,
    ymin=0,
    ymax=0.17,
]

\addplot[
    color=NavyBlue,
    mark=*,
    thick
]
coordinates {
    (10,0.0878471346)
    (20,0.1123155430)
    (30,0.1255929664)
    (40,0.1329737150)
    (50,0.1374842696)
    (100,0.1432052709)
    (200,0.1439125603)
    (300,0.1439125603)
    (450,0.1439125603)
};
\addlegendentry{AK (train)}

\addplot[
color=cyan,
mark=*,
thick
]
coordinates {
(10,0.1215018244)
(20,0.1447663750)
(30,0.1538434038)
(40,0.1582006879)
(50,0.1606952009)
(100,0.1624366655)
(200,0.1624366655)
(300,0.1624366655)
(450,0.1624366655)
};
\addlegendentry{AK (test)}
\addplot[
    color=BrickRed,
    dashed,
    thick
]
coordinates {
    (10,0.0548046314)
    (20,0.0548046314)
    (30,0.0548046314)
    (40,0.0548046314)
    (50,0.0548046314)
    (100,0.0548046314)
    (200,0.0548046314)
    (300,0.0548046314)
    (450,0.0548046314)
};
\addlegendentry{$\tau_{\mathcal{X}}/500$ (train)}

\addplot[
color=orange,
dashed,
thick
]
coordinates {
(10,0.0616035122)
(20,0.0616035122)
(30,0.0616035122)
(40,0.0616035122)
(50,0.0616035122)
(100,0.0616035122)
(200,0.0616035122)
(300,0.0616035122)
(450,0.0616035122)
};
\addlegendentry{$\tau_{\mathcal{X}}/500$ (test)}

\nextgroupplot[
    width=14cm,
    height=5cm,
    title={\textbf{MNIST $N_{max}$ ablation}},
    grid=both,
    legend style={
    at={(0.7,0.65)},
    anchor=north,
    font=\tiny,
    legend columns=2,
    },
    xtick={10,50,100,200,300,450},
    xmin=5,xmax=450,
    ymin=0,
    ymax=0.058,
]

\addplot[
    color=NavyBlue,
    mark=*,
    thick
]
coordinates {
(10, 0.0)
(50, 0.0)
(100, 0.0)
(200, 0.0)
(300, 0.0)
(450, 0.0)
};
\addlegendentry{AK (train)}

\addplot[
    color=cyan,
    mark=*,
    thick
]
coordinates {
(10, 0.0)
(50, 0.0)
(100, 0.0)
(200, 0.0)
(300, 0.0)
(450, 0.0)
};
\addlegendentry{AK (test)}

\addplot[
    color=BrickRed,
    dashed,
    thick
]
coordinates {
    (10,0.0094646047)
    (20,0.0094646047)
    (30,0.0094646047)
    (40,0.0094646047)
    (50,0.0094646047)
    (100,0.0094646047)
    (200,0.0094646047)
    (300,0.0094646047)
    (450,0.0094646047)
};
\addlegendentry{$\tau_{\mathcal{X}}/500$ (train)}

\addplot[
    color=orange,
    dashed,
    thick
]
coordinates {
    (10,0.0107638488)
    (20,0.0107638488)
    (30,0.0107638488)
    (40,0.0107638488)
    (50,0.0107638488)
    (100,0.0107638488)
    (200,0.0107638488)
    (300,0.0107638488)
    (450,0.0107638488)
};
\addlegendentry{$\tau_{\mathcal{X}}/500$ (test)}

\nextgroupplot[
    width=14cm,
    height=5cm,
    title={\textbf{RESISC45 $N_{max}$ ablation}},
    grid=both,
    legend style={
    at={(0.75,0.66)},
    anchor=north,
    font=\tiny,
    legend columns=2,
    },
    xtick={10,50,100,200,300,450},
    xmin=5,xmax=450,
    ymin=0,
    ymax=0.2,
]

\addplot[
    color=NavyBlue,
    mark=*,
    thick
]
coordinates {
    (10,0.0851665549)
    (20,0.1036849333)
    (30,0.1212809258)
    (40,0.1370669312)
    (50,0.1460322286)
    (100,0.1507288010)
    (200,0.1507288010)
    (300,0.1507288010)
    (450,0.1507288010)
};
\addlegendentry{AK (train)}

\addplot[
    color=cyan,
    mark=*,
    thick
]
coordinates {
    (10,0.0759207754)
    (20,0.1027658552)
    (30,0.1185768803)
    (40,0.1333866542)
    (50,0.1413067958)
    (100,0.1446808352)
    (200,0.1446808352)
    (300,0.1446808352)
    (450,0.1446808352)
};
\addlegendentry{AK (test)}

\addplot[
    color=BrickRed,
    dashed,
    thick
]
coordinates {
    (10,0.0625393383)
    (20,0.0625393383)
    (30,0.0625393383)
    (40,0.0625393383)
    (50,0.0625393383)
    (100,0.0625393383)
    (200,0.0625393383)
    (300,0.0625393383)
    (450,0.0625393383)
};
\addlegendentry{$\tau_{\mathcal{X}}/500$ (train)}

\addplot[
    color=orange,
    dashed,
    thick
]
coordinates {
    (10,0.0624344305)
    (20,0.0624344305)
    (30,0.0624344305)
    (40,0.0624344305)
    (50,0.0624344305)
    (100,0.0624344305)
    (200,0.0624344305)
    (300,0.0624344305)
    (450,0.0624344305)
};
\addlegendentry{$\tau_{\mathcal{X}}/500$ (test)}

        \nextgroupplot[
    width=14cm,
    height=5cm,
    title = {\textbf{EUROSAT $N_{max}$ ablation}},
    xlabel={$N_{max}$},
    grid=both,
    legend style={
    at={(0.75,0.97)},
    anchor=north,
    font=\tiny,
    legend columns=2,
    },
    xtick={10,50,100,200,300,450},
    xmin=5, xmax=450,
    ymin=0,
    ymax=0.05,
]

\addplot[
    color=NavyBlue,
    mark=*,
    thick
]
coordinates {
    (10,0.0204666395)
    (20,0.0211669845)
    (30,0.0221691757)
    (40,0.0226807700)
    (50,0.0251945850)
    (100,0.0279229006)
    (200,0.0222225843)
    (300,0.0221357677)
    (450,0.0221357677)
};
\addlegendentry{AK (train)}

\addplot[
    color=cyan,
    mark=*,
    thick
]
coordinates {
    (10,0.0283437321)
    (20,0.0364620921)
    (30,0.0377866107)
    (40,0.0350736087)
    (50,0.0332463368)
    (100,0.0311062814)
    (200,0.0311062814)
    (300,0.0311062814)
    (450,0.0311062814)
};
\addlegendentry{AK (test)}

\addplot[
    color=BrickRed,
    dashed,
    thick
]
coordinates {
    (10,0.0048869685)
    (20,0.0048869685)
    (30,0.0048869685)
    (40,0.0048869685)
    (50,0.0048869685)
    (100,0.0048869685)
    (200,0.0048869685)
    (300,0.0048869685)
    (450,0.0048869685)
};
\addlegendentry{$\tau_{\mathcal{X}}/500$ (train)}

\addplot[
    color=orange,
    dashed,
    thick
]
coordinates {
    (10,0.0054098130)
    (20,0.0054098130)
    (30,0.0054098130)
    (40,0.0054098130)
    (50,0.0054098130)
    (100,0.0054098130)
    (200,0.0054098130)
    (300,0.0054098130)
    (450,0.0054098130)
};
\addlegendentry{$\tau_{\mathcal{X}}/500$ (test)}

        \end{groupplot}

    \end{tikzpicture}
    \caption{Average kernel size, $AK$ or also $\operatorname{Kersize}(\mathcal D_{M'},1)_{K}$, for classification datasets as a function of the maximum number of nearest neighbors $N_{max}$ used in the threshold $\tau_{\mathcal{X}}$ computation. The threshold is chosen as the $q^{th}$ quantile of the nearest-neighbor distances according to \eqref{eq:tauxqrule}. The quantile $q$ and number of anchor samples $K$ for train and test splits can be found in Table \ref{tab:classification-parameters}.}
\label{fig:classification-kersize-nmax-ablation1}
\end{figure}
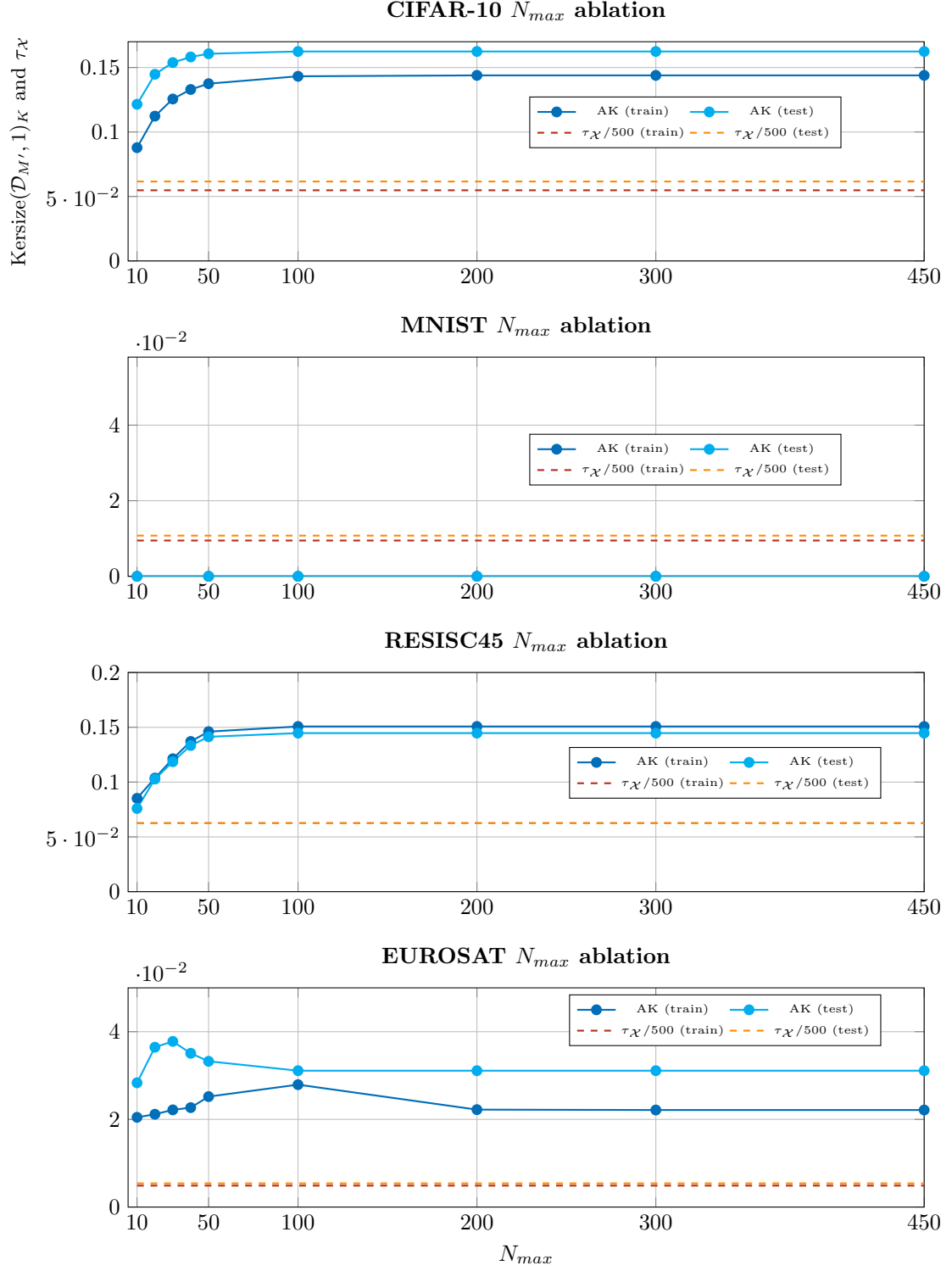

\noindent For the \textbf{MNIST} open-source dataset from \href{https://docs.pytorch.org/vision/main/generated/torchvision.datasets.MNIST.html}{Torch Vision}, we converted the input images to tensors and normalized them using the MNIST training-set channel means and standard deviations. We computed the average kernel size from feasible sets as a function of the maximum number of nearest neighbors and reached convergence with threshold $10^{-4}$ at $N_{\max} =50$; see Figure \ref{fig:classification-kersize-nmax-ablation1}. MNIST contains single-channel grayscale images of handwritten digits of size $28\times 28$ from $0$ to $9$. For MNIST, we computed the accuracies on the test set and the constructed test set for an EfficientNet B0, $\phi_{2nd}$ and $\phi_{3rd}$ with further random initializations of this model. The accuracy of the optimal classifier on $\mathcal D_{M'_{\mathrm{test}}}$ is given by $\operatorname{Acc}_*= 100.00$. 

\noindent For the \textbf{CIFAR-10} open-source dataset from \href{https://docs.pytorch.org/vision/main/generated/torchvision.datasets.CIFAR10.html}{Torch Vision}, we converted the input images to tensors and normalized them using the CIFAR-10 training-set channel means and standard deviations. We computed the average kernel size from feasible sets as a function of the maximum number of nearest neighbors and reached convergence with threshold $10^{-4}$ at $N_{\max} =50$, see Figure \ref{fig:classification-kersize-nmax-ablation1}. CIFAR-10 contains three-channel color images of objects of size $32\times 32$ from $10$ different classes. For CIFAR-10, we computed the accuracies on the test set and the constructed test set for a pretrained ResNet56, $\phi_{2nd}$, and pretrained ShuffleNetV2, $\phi_{3rd}$, from \cite{chenyaofo_pytorch_cifar_models} and retrained on the train dataset. The accuracy for $\phi_{1st}$ is reported from the CIFAR-10 Kaggle \href{https://www.kaggle.com/competitions/mu-cifar10/leaderboard}{leaderboard}. The accuracy of the optimal classifier on $\mathcal D_{M'_{\mathrm{test}}}$ is given by $\operatorname{Acc}_*= 88.43$. 

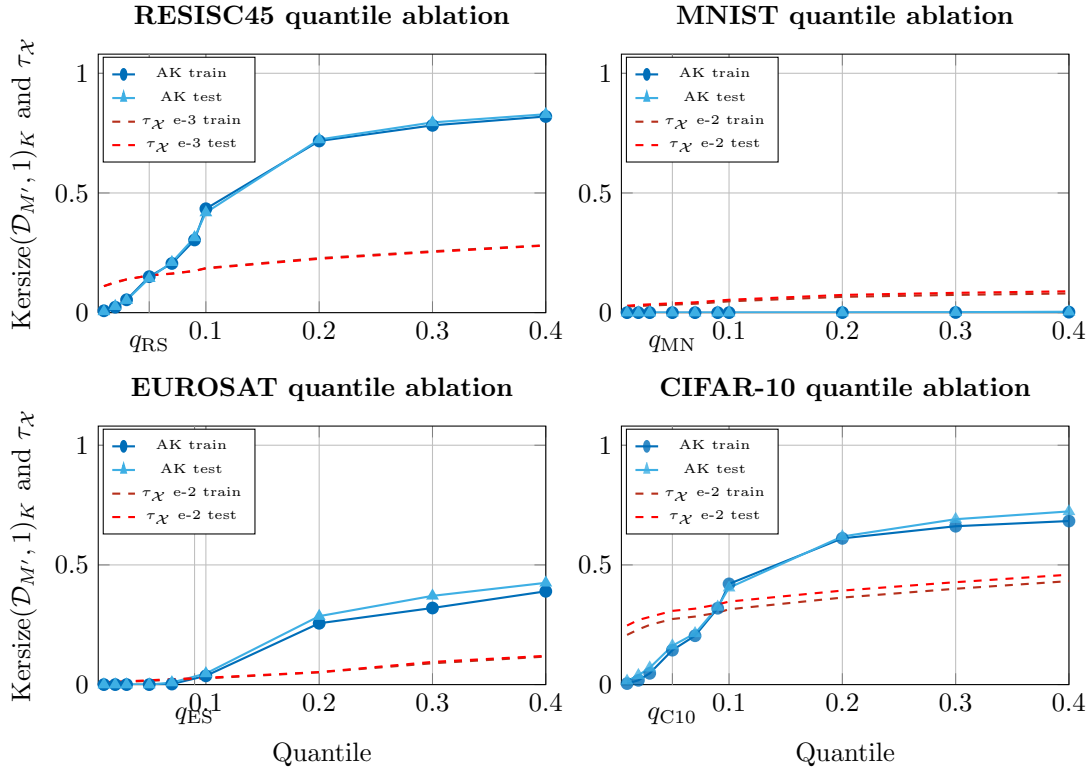
\begin{figure}[ht]
    \centering
    \begin{tikzpicture}

        \begin{groupplot}[
            group style={
                group size=2 by 2,
                horizontal sep=1.0cm,
                vertical sep=1.5cm
            },
            width=7.5cm,
            height=5cm,
    xmin=0.005, xmax=0.4,
    ymin=0, ymax=1.08,
    xtick={0.1,0.2,0.3,0.4},
    grid=major,
    legend style={at={(0.01,0.99)},anchor=north west, legend columns=1,font=\tiny,legend image post style={xscale=0.5},},
]

        \nextgroupplot[  title = {\textbf{RESISC45 quantile ablation}},
         ylabel={$\operatorname{Kersize}(\mathcal D_{M'},1)_K$ and $\tau_{\mathcal{X}}$},
            extra x ticks={0.05},
    extra x tick labels={$q_{\mathrm{RS}}$},
    extra x tick style={
    tick label style={yshift=-5pt},
    },]
\addplot[
    NavyBlue,
    thick,
    mark=*,
]
coordinates {
    (0.01,0.007732)
    (0.02,0.022722)
    (0.03,0.053641)
    (0.05,0.150383)
    (0.07,0.205463)
    (0.09,0.303586)
    (0.10,0.434087)
    (0.20,0.717094)
    (0.30,0.782460)
    (0.40,0.819585)
};
\addlegendentry{\tiny{$\mathrm{AK}$ train}}

\addplot[
    CornflowerBlue,
    thick,
    mark=triangle*,
]
coordinates {
    (0.01,0.004971)
    (0.02,0.026675)
    (0.03,0.051967)
    (0.05,0.143733)
    (0.07,0.210315)
    (0.09,0.314064)
    (0.10,0.418093)
    (0.20,0.723220)
    (0.30,0.794750)
    (0.40,0.828915)
};
\addlegendentry{\tiny{$\mathrm{AK}$ test}}

\addplot[
    BrickRed,
    thick,
    dashed,
]
coordinates {
    (0.01,0.110979)
    (0.02,0.124488)
    (0.03,0.138677)
    (0.05,0.155572)
    (0.07,0.162460)
    (0.09,0.173826)
    (0.10,0.186551)
    (0.20,0.227804)
    (0.30,0.256464)
    (0.40,0.281758)
};
\addlegendentry{\tiny{$\tau_{\mathcal{X}}$ e-3 train}}

\addplot[
    red,
    thick,
    dashed,
]
coordinates {
    (0.01,0.110496)
    (0.02,0.128347)
    (0.03,0.138966)
    (0.05,0.155344)
    (0.07,0.163611)
    (0.09,0.174838)
    (0.10,0.184542)
    (0.20,0.224964)
    (0.30,0.253612)
    (0.40,0.280437)
};
\addlegendentry{\tiny{$\tau_{\mathcal{X}}$ e-3 test}}

\nextgroupplot[title = {\textbf{MNIST quantile ablation}},
 extra x ticks={0.05},
    extra x tick labels={$q_{\mathrm{MN}}$},
    extra x tick style={
    tick label style={yshift=-5pt},
    },]
\addplot[
    NavyBlue,
    thick,
    mark=*,
]
coordinates {
(0.01, 0.0)
(0.02, 0.0)
(0.03, 0.0)
(0.05, 0.0)
(0.07, 0.0)
(0.09, 0.0)
(0.1, 0.0)
(0.2, 0.00024171093939953121)
(0.3, 0.0007164978711050381)
(0.4, 0.0021031606208374914)
};
\addlegendentry{\tiny{$\mathrm{AK}$ train}}

\addplot[
    CornflowerBlue,
    thick,
    mark=triangle*,
]
coordinates {
(0.01, 0.0)
(0.02, 0.0)
(0.03, 0.0)
(0.05, 0.0)
(0.07, 0.0)
(0.09, 0.0)
(0.1, 0.0)
(0.2, 0.0003569826942377741)
(0.3, 0.0018097645095006697)
(0.4, 0.003827350126388734)
};
\addlegendentry{\tiny{$\mathrm{AK}$ test}}

\addplot[
    BrickRed,
    thick,
    dashed,
]
coordinates {
(0.01, 0.0257184011220932)
(0.02, 0.027901278638839724)
(0.03, 0.029736305403709413)
(0.05, 0.03340562331676483)
(0.07, 0.037281861329078674)
(0.08, 0.04009519691467285)
(0.1, 0.04732302331924439)
(0.2, 0.0660067892074585)
(0.3, 0.07414066791534424)
(0.4, 0.08013680839538575)
};
\addlegendentry{\tiny{$\tau_{\mathcal{X}}$ e-2 train}}

\addplot[
    red,
    thick,
    dashed,
]
coordinates {
(0.01, 0.02887741577625275)
(0.02, 0.03233955659866333)
(0.03, 0.03476171231269836)
(0.05, 0.03896887290477753)
(0.07, 0.04315092167854309)
(0.1, 0.05381924390792848)
(0.2, 0.07402444648742676)
(0.3, 0.08275299453735352)
(0.4, 0.08913255500793458)
};
\addlegendentry{\tiny{$\tau_{\mathcal{X}}$ e-2 test}}

\nextgroupplot[title = {\textbf{EUROSAT quantile ablation}},
 ylabel={$\operatorname{Kersize}(\mathcal D_{M'},1)_K$ and $\tau_{\mathcal{X}}$},
     xlabel={Quantile},
    extra x ticks={0.09},
    extra x tick labels={$q_{\mathrm{ES}}$},
    extra x tick style={
    tick label style={yshift=-5pt},
    },]
\addplot[
    color=NavyBlue,
    mark=*,
    thick
]
coordinates {
    (0.01, 0.0)
    (0.02, 0.0)
    (0.03, 0.0)
    (0.05, 0.0)
    (0.07, 0.0022722993499584574)
    (0.1, 0.03655821495577588)
    (0.2, 0.25615785690176734)
    (0.3, 0.3200055651594985)
    (0.4, 0.38972338640999277)
};
\addlegendentry{\tiny{$\mathrm{AK}$ train}}

\addplot[
    color=CornflowerBlue,
    mark=triangle*,
    thick
]
coordinates {
    (0.01, 0.0)
    (0.02, 0.0)
    (0.03, 0.0)
    (0.05, 0.0)
    (0.07, 0.00847106670646249)
    (0.1, 0.0463588981232552)
    (0.2, 0.28587726040203193)
    (0.3, 0.37071700839075)
    (0.4, 0.4251600989388694)
};
\addlegendentry{\tiny{$\mathrm{AK}$ test}}

\addplot[
    color=BrickRed,
    dashed,
    thick
]
coordinates {
    (0.01, 0.010300685501098632)
    (0.02, 0.01086460874080658)
    (0.03, 0.011934553730487822)
    (0.05, 0.015672801256179811)
    (0.07, 0.0201067391872406)
    (0.1, 0.025918546199798587)
    (0.2, 0.05135299110412599)
    (0.3, 0.08907579517364502)
    (0.4, 0.11704438400268554)
};
\addlegendentry{\tiny{$\tau_{\mathcal{X}}$ e-2 train}}

\addplot[
    color=red,
    dashed,
    thick
]
coordinates {
    (0.01,  0.010416629314422607)
    (0.02, 0.011488429117202759)
    (0.03, 0.013619306468963623)
    (0.05, 0.018043369412422179)
    (0.07, 0.02277106354236603)
    (0.1, 0.02852147674560547)
    (0.2, 0.0522320671081543)
    (0.3, 0.09480838489532472)
    (0.4, 0.12016570281982421)
};
\addlegendentry{\tiny{$\tau_{\mathcal{X}}$ e-2 test}}

        \nextgroupplot[title = {\textbf{CIFAR-10 quantile ablation}},     
            xlabel={Quantile},
        extra x ticks={0.05},
    extra x tick labels={$q_{\mathrm{C10}}$},
    extra x tick style={
    tick label style={yshift=-5pt},
    },]
\addplot[
    NavyBlue,
    thick,
    mark=*,
    mark options={opacity=0.8},
]
coordinates {
    (0.01,0.004330)
    (0.02,0.017886)
    (0.03,0.047076)
    (0.05,0.143913)
    (0.07,0.204128)
    (0.09,0.319019)
    (0.10,0.420678)
    (0.20,0.610932)
    (0.30,0.661612)
    (0.40,0.683328)
};

\addlegendentry{\tiny{$\mathrm{AK}$ train}}

\addplot[
    CornflowerBlue,
    thick,
    mark=triangle*,
     mark options={opacity=0.8},
]
coordinates {
    (0.01,0.014990)
    (0.02,0.038940)
    (0.03,0.070249)
    (0.05,0.162437)
    (0.07,0.213772)
    (0.09,0.324273)
    (0.10,0.404681)
    (0.20,0.618829)
    (0.30,0.690964)
    (0.40,0.723554)
};
\addlegendentry{\tiny{$\mathrm{AK}$ test}}

\addplot[
    BrickRed,
    thick,
    dashed,
]
coordinates {
    (0.01,0.207579)
    (0.02,0.230604)
    (0.03,0.249360)
    (0.05,0.274023)
    (0.07,0.284213)
    (0.09,0.300279)
    (0.10,0.314679)
    (0.20,0.363492)
    (0.30,0.400448)
    (0.40,0.432261)
};
\addlegendentry{\tiny{$\tau_{\mathcal{X}}$ e-2 train}}

\addplot[
    red,
    thick,
    dashed,
]
coordinates {
    (0.01,0.246733)
    (0.02,0.270044)
    (0.03,0.282745)
    (0.05,0.308018)
    (0.07,0.317210)
    (0.09,0.334808)
    (0.10,0.346814)
    (0.20,0.392800)
    (0.30,0.427824)
    (0.40,0.459140)
};
\addlegendentry{\tiny{$\tau_{\mathcal{X}}$ e-2 test}}
        \end{groupplot}

    \end{tikzpicture}
    \caption{Average kernel size, $\operatorname{Kersize}(\mathcal D_{M'},1)_K$ ($\mathrm{AK}$), as a function of the quantile  $q$ for the classification datasets. The additional ticks on the x-axis indicate the quantiles for the computation of $\tau_{\mathcal{X}}(q)$. The number of anchor samples $K$ are in Table \ref{tab:classification-parameters}.}
\label{fig:classification-kersize-vs-q}
\end{figure}

\noindent For the \textbf{RESISC45} open-source dataset \cite{cheng2017remote} from \href{https://docs.torchgeo.org/en/v0.3.1/_modules/torchgeo/datamodules/resisc45.html}{TorchGeo}, the input images were converted to tensors. We computed the average kernel size from feasible sets as a function of the maximum number of nearest neighbors and reached convergence with threshold $10^{-4}$ at $N_{\max} =200$, see Figure \ref{fig:classification-kersize-nmax-ablation1}. The RESISC45 dataset \cite{cheng2017remote} is a publicly available benchmark for remote-sensing scene classification. It contains approximately $31.5k$ images spanning $45$ land-use and scene categories, with substantial variation in spatial resolution, viewpoint, illumination, and background. The task is to map each RGB image of size $256\times 256$ to one of the $45$ landcover classes. The accuracies on the test dataset with a $20/80$ train/test split are from the best models from \cite[Table 6]{cheng2017remote}. For RESISC45, the train and test intervals were computed from the mean average kernel sizes over \(20\) randomly selected subsets of \(3\mathrm{k}\) samples, $100\,\overline{\operatorname{Kersize}}_{\mathrm{train}}
=16.15\pm2.37\%$ and $100\,\overline{\operatorname{Kersize}}_{\mathrm{test}}
=17.64\pm1.73\%$. The values shown in the ablation figures, such as Figure \ref{fig:classification-kersize-vs-q}, correspond to individual subsets and are not used directly to compute these intervals. The accuracy of the optimal classifier on $\mathcal D_{M'_{\mathrm{test}}}$ is given by $\operatorname{Acc}_*= 91.33$. 

\noindent For the \textbf{EUROSAT} open-source dataset \cite{helber2019eurosat} we used the RGB version of the dataset provided \href{https://huggingface.co/datasets/torchgeo/eurosat}{by TorchGeo on HuggingFace} together with the provided split into training, validation and testing. The input images were converted to tensors and normalized to to [-1,1]. We computed the average kernel size from feasible sets as a function of the maximum number of nearest neighbors and reached convergence at $N_{\max} =200$, see Figure \ref{fig:classification-kersize-nmax-ablation1}. The EUROSAT dataset \cite{helber2019eurosat} is a publicly available benchmark for land-use and land-cover classification based on Sentinel-2 satellite imagery. It contains $27k$ labeled image patches distributed across $10$ classes, including forests, residential areas, agricultural land, rivers, and industrial sites. The dataset provides both RGB and multispectral imagery and captures substantial variation in geographic location, land-cover structure, and spectral characteristics. We chose the classification task mapping RGB images of size $64 \times 64$ to $10$ different landcover classes. The best model's test-set accuracies is from \cite[Table 5]{adegun2023review} and a locally trained DNNs. The accuracy of the optimal classifier on $\mathcal D_{M'_{\mathrm{test}}}$ is given by $\operatorname{Acc}_*= 97.94$.

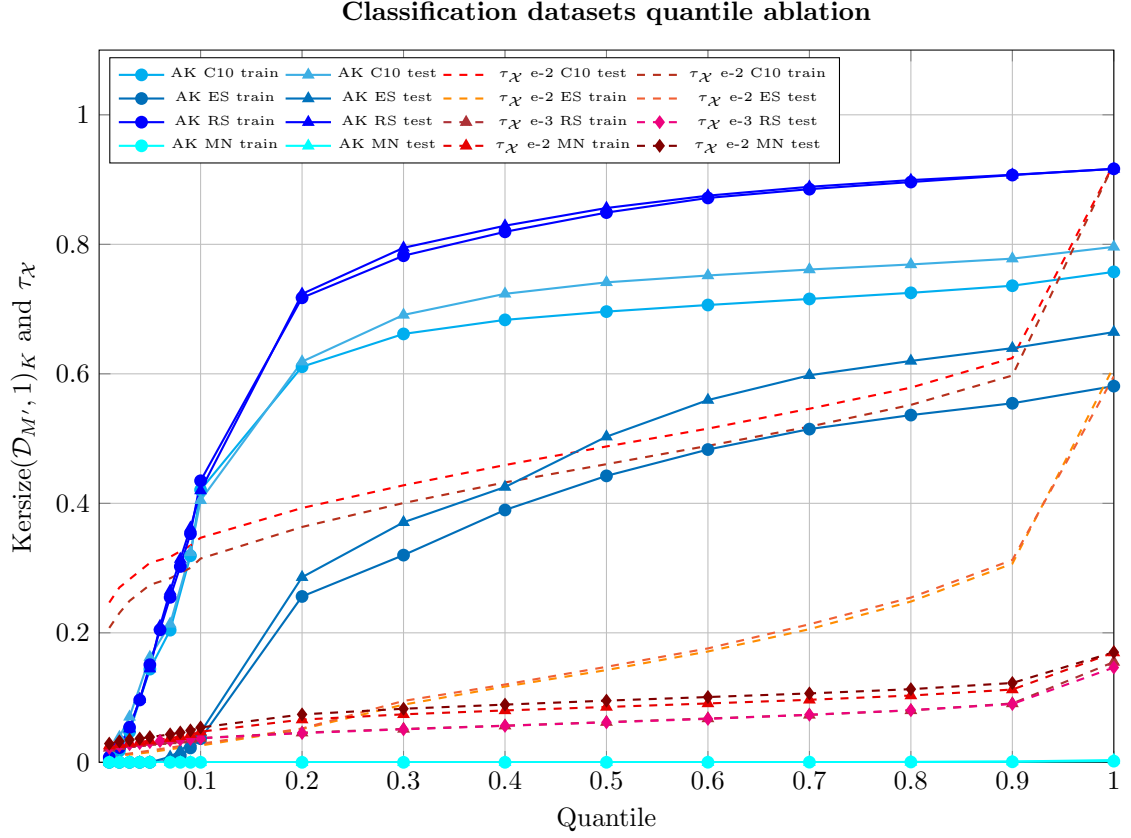
\begin{figure}[h!]
\centering
\begin{tikzpicture}
\begin{axis}[
    width=15cm,
    height=11cm,
    title = {\textbf{Classification datasets quantile ablation}},
    xlabel={Quantile},
    ylabel={$\operatorname{Kersize}(\mathcal D_{M'},1)_{K}$ and $\tau_{\mathcal{X}}$},
    xmin=0.0, xmax=1.0,
    ymin=0, ymax=1.1,
    xtick={0.1,0.2,0.3,0.4,0.5,0.6,0.7,0.8,0.9,1.0},
    grid=major,
    legend style={
    at={(0.01,0.99)},
    anchor=north west,
    font=\tiny,
    legend columns=4,
    },
    mark options={solid},
]

\addplot[
    cyan,
    thick,
    mark=*,
]
coordinates {
    (0.01,0.004330)
    (0.02,0.017886)
    (0.03,0.047076)
    (0.05,0.143913)
    (0.07,0.204128)
    (0.09,0.319019)
    (0.10,0.420678)
    (0.20,0.610932)
    (0.30,0.661612)
    (0.40,0.683328)
    (0.50,0.696100)
    (0.60,0.706287)
    (0.70,0.715637)
    (0.80,0.725084)
    (0.90,0.736038)
    (1.00,0.757439)
};

\addlegendentry{\tiny{$\mathrm{AK}$ C10 train}}

\addplot[
    CornflowerBlue,
    thick,
    mark=triangle*,
]
coordinates {
    (0.01,0.014990)
    (0.02,0.038940)
    (0.03,0.070249)
    (0.05,0.162437)
    (0.07,0.213772)
    (0.09,0.324273)
    (0.10,0.404681)
    (0.20,0.618829)
    (0.30,0.690964)
    (0.40,0.723554)
    (0.50,0.741295)
    (0.60,0.751870)
    (0.70,0.761091)
    (0.80,0.768919)
    (0.90,0.777879)
    (1.00,0.796090)
};
\addlegendentry{\tiny{$\mathrm{AK}$ C10 test}}

\addplot[
    red,
    thick,
    dashed,
]
coordinates {
    (0.01,0.246733)
    (0.02,0.270044)
    (0.03,0.282745)
    (0.05,0.308018)
    (0.07,0.317210)
    (0.09,0.334808)
    (0.10,0.346814)
    (0.20,0.392800)
    (0.30,0.427824)
    (0.40,0.459140)
    (0.50,0.487826)
    (0.60,0.515381)
    (0.70,0.546092)
    (0.80,0.578716)
    (0.90,0.624352)
    (1.00,0.924051)
};
\addlegendentry{\tiny{$\tau_{\mathcal{X}}$ e-2 C10 test}}

\addplot[
    BrickRed,
    thick,
    dashed,
]
coordinates {
    (0.01,0.207579)
    (0.02,0.230604)
    (0.03,0.249360)
    (0.05,0.274023)
    (0.07,0.284213)
    (0.09,0.300279)
    (0.10,0.314679)
    (0.20,0.363492)
    (0.30,0.400448)
    (0.40,0.432261)
    (0.50,0.460472)
    (0.60,0.488797)
    (0.70,0.518430)
    (0.80,0.552026)
    (0.90,0.597364)
    (1.00,0.924033)
};
\addlegendentry{\tiny{$\tau_{\mathcal{X}}$ e-2 C10 train}}

\addplot[
    color=NavyBlue,
    mark=*,
    thick
]
coordinates {
(0.01, 0.0)
(0.02, 0.0)
(0.03, 0.0)
(0.04, 0.0)
(0.05, 0.0)
(0.07, 0.0022722993499584574)
(0.08, 0.009584928949550169)
(0.09, 0.02222258431045058)
(0.1, 0.03655821495577588)
(0.2, 0.25615785690176734)
(0.3, 0.3200055651594985)
(0.4, 0.38972338640999277)
(0.5, 0.44240157310155637)
(0.6, 0.48305361524305757)
(0.7, 0.5146237491770923)
(0.8, 0.5362533049542986)
(0.9, 0.5545048158931871)
(1.0, 0.580998836841235)
};
\addlegendentry{\tiny{$\mathrm{AK}$ ES train}}

\addplot[
    color=RoyalBlue,
    mark=triangle*,
    thick
]
coordinates {
(0.01, 0.0)
(0.02, 0.0)
(0.03, 0.0)
(0.04, 0.0)
(0.05, 0.0)
(0.07, 0.00847106670646249)
(0.08, 0.017765186624709738)
(0.09, 0.031106281434211912)
(0.1, 0.0463588981232552)
(0.2, 0.28587726040203193)
(0.3, 0.37071700839075)
(0.4, 0.4251600989388694)
(0.5, 0.502919618667436)
(0.6, 0.5594902416666211)
(0.7, 0.5977298539782429)
(0.8, 0.6199015441290467)
(0.9, 0.639641008881796)
(1.0, 0.6643990630334035)
};
\addlegendentry{\tiny{$\mathrm{AK}$ ES test}}

\addplot[
    color=orange,
    dashed,
    thick
]
coordinates {
(0.01, 0.010300685501098632)
(0.02, 0.01086460874080658)
(0.03, 0.011934553730487822)
(0.04, 0.013746134281158449)
(0.05, 0.01567280125617981)
(0.07, 0.0201067391872406)
(0.08, 0.02253253765106201)
(0.09, 0.024434842586517336)
(0.1, 0.025918546199798588)
(0.2, 0.051352991104125986)
(0.3, 0.08907579517364501)
(0.4, 0.11704438400268553)
(0.5, 0.14248043537139893)
(0.6, 0.17119126892089842)
(0.7, 0.20562964820861815)
(0.8, 0.24827328109741212)
(0.9, 0.3067802047729492)
(1.0, 0.6111174392700195)
};
\addlegendentry{\tiny{$\tau_{\mathcal{X}}$ e-2 ES train}}

\addplot[
    color=RedOrange,
    dashed,
    thick
]
coordinates {
(0.01, 0.010416629314422608)
(0.02, 0.01148842911720276)
(0.03, 0.013619306468963623)
(0.04, 0.01578873052597046)
(0.05, 0.01804336941242218)
(0.07, 0.02277106354236603)
(0.08, 0.025033998489379882)
(0.09, 0.02704906494617462)
(0.1, 0.02852147674560547)
(0.2, 0.0522320671081543)
(0.3, 0.09480838489532471)
(0.4, 0.12016570281982421)
(0.5, 0.14750155448913574)
(0.6, 0.17593668365478518)
(0.7, 0.21319600677490227)
(0.8, 0.2544483413696289)
(0.9, 0.3124808654785156)
(1.0, 0.5960328674316406)
};
\addlegendentry{\tiny{$\tau_{\mathcal{X}}$ e-2 ES test}}

\addplot[
    blue,
    thick,
    mark=*,
]
coordinates {
    (0.01,0.007737)
    (0.02,0.022735)
    (0.03,0.054074)
    (0.04,0.096290)
    (0.05,0.150729)
    (0.06,0.204630)
    (0.07,0.254814)
    (0.08,0.302472)
    (0.09,0.353133)
    (0.10,0.434819)
    (0.20,0.717433)
    (0.30,0.782379)
    (0.40,0.819314)
    (0.50,0.849022)
    (0.60,0.871633)
    (0.70,0.885197)
    (0.80,0.896172)
    (0.90,0.906950)
    (1.00,0.916603)
};
\addlegendentry{\tiny{$\mathrm{AK}$ RS train}}

\addplot[
    blue,
    thick,
    mark=triangle*,
]
coordinates {
    (0.01,0.004983)
    (0.02,0.026694)
    (0.03,0.051977)
    (0.04,0.097603)
    (0.05,0.144681)
    (0.06,0.210379)
    (0.07,0.264687)
    (0.08,0.313883)
    (0.09,0.362047)
    (0.10,0.419797)
    (0.20,0.723406)
    (0.30,0.794638)
    (0.40,0.828752)
    (0.50,0.856077)
    (0.60,0.875169)
    (0.70,0.888850)
    (0.80,0.899192)
    (0.90,0.907390)
    (1.00,0.916435)
};
\addlegendentry{\tiny{$\mathrm{AK}$ RS test}}

\addplot[
     Maroon,
    thick,
    dashed,
    mark=triangle*,
]
coordinates {
    (0.01,0.022336)
    (0.02,0.025008)
    (0.03,0.027898)
    (0.04,0.029572)
    (0.05,0.031270)
    (0.06,0.032629)
    (0.07,0.033793)
    (0.08,0.034894)
    (0.09,0.035968)
    (0.10,0.037501)
    (0.20,0.045809)
    (0.30,0.051550)
    (0.40,0.056605)
    (0.50,0.062143)
    (0.60,0.067729)
    (0.70,0.073261)
    (0.80,0.080222)
    (0.90,0.090929)
    (1.00,0.154887)
};
\addlegendentry{\tiny{$\tau_{\mathcal{X}}$ e-3 RS train}}

\addplot[
    RubineRed,
    thick,
    dashed,
    mark=diamond*,
]
coordinates {
    (0.01,0.022211)
    (0.02,0.025791)
    (0.03,0.027915)
    (0.04,0.029672)
    (0.05,0.031217)
    (0.06,0.032848)
    (0.07,0.034142)
    (0.08,0.035121)
    (0.09,0.036050)
    (0.10,0.037112)
    (0.20,0.045217)
    (0.30,0.050974)
    (0.40,0.056361)
    (0.50,0.061736)
    (0.60,0.067082)
    (0.70,0.073492)
    (0.80,0.080493)
    (0.90,0.090062)
    (1.00,0.146407)
};
\addlegendentry{\tiny{$\tau_{\mathcal{X}}$ e-3 RS test}}

\addplot[
    cyanblue,
    thick,
    mark=*,
]
coordinates {
(0.01, 0.0)
(0.02, 0.0)
(0.03, 0.0)
(0.04, 0.0)
(0.05, 0.0)
(0.07, 0.0)
(0.08, 0.0)
(0.09, 0.0)
(0.1, 0.0)
(0.2, 2.417104786695266e-06)
(0.3, 7.164978711050381e-06)
(0.4, 2.1031606208374913e-05)
(0.5, 5.1324741401618726e-05)
(0.6, 0.00012111309812438145)
(0.7, 0.0002628146160053209)
(0.8, 0.0004978486718215331)
(0.9, 0.0008065415144828198)
(1.0, 0.001825537673741513)
};
\addlegendentry{\tiny{$\mathrm{AK}$ MN train}}

\addplot[
    cyanblue,
    thick,
    mark=triangle*,
]
coordinates {
(0.01, 0.0)
(0.02, 0.0)
(0.03, 0.0)
(0.04, 0.0)
(0.05, 0.0)
(0.07, 0.0)
(0.08, 0.0)
(0.09, 0.0)
(0.1, 0.0)
(0.2, 3.569826942377741e-06)
(0.3, 1.8097645095006696e-05)
(0.4, 3.8273501263887336e-05)
(0.5, 7.608290071000816e-05)
(0.6, 0.0001476700806422648)
(0.7, 0.0002937790141110521)
(0.8, 0.0005929775269043756)
(0.9, 0.0012811547488877533)
(1.0, 0.003496870968012172)
};
\addlegendentry{\tiny{$\mathrm{AK}$ MN test}}

\addplot[
    red!90!black,
    thick,
    dashed,
    mark=triangle*,
]
coordinates {
(0.01, 0.0257184011220932)
(0.02, 0.027901278638839722)
(0.03, 0.029736305403709414)
(0.04, 0.03156243543624878)
(0.05, 0.03340562331676483)
(0.07, 0.037281861329078674)
(0.08, 0.04009519691467285)
(0.09, 0.04326951556205749)
(0.1, 0.04732302331924439)
(0.2, 0.06600678920745849)
(0.3, 0.07414066791534424)
(0.4, 0.08013680839538574)
(0.5, 0.08548875808715821)
(0.6, 0.09090850067138671)
(0.7, 0.09657788181304933)
(0.8, 0.1031530933380127)
(0.9, 0.11241935920715332)
(1.0, 0.16955711364746093)
};
\addlegendentry{\tiny{$\tau_{\mathcal{X}}$ e-2 MN train}}

\addplot[
    red!50!black,
    thick,
    dashed,
    mark=diamond*,
]
coordinates {
(0.01, 0.02887741577625275)
(0.02, 0.03233955659866333)
(0.03, 0.03476171231269836)
(0.04, 0.03666823863983154)
(0.05, 0.03896887290477753)
(0.07, 0.04315092167854309)
(0.08, 0.046107306480407714)
(0.09, 0.049276986837387085)
(0.1, 0.05381924390792848)
(0.2, 0.07402444648742676)
(0.3, 0.08275299453735352)
(0.4, 0.08913255500793457)
(0.5, 0.09527891159057617)
(0.6, 0.1008311424255371)
(0.7, 0.1064013671875)
(0.8, 0.11297747802734376)
(0.9, 0.12252912330627443)
(1.0, 0.1701069450378418)
};
\addlegendentry{\tiny{$\tau_{\mathcal{X}}$ e-2 MN test}}

\end{axis}

\end{tikzpicture}

\caption{Average kernel size, $AK$ or also $\operatorname{Kersize}(\mathcal D_{M'},1)_{K}$, and $\tau_{\mathcal{X}}(q)$ as a function of the quantile parameter $q$ for the different classification datasets, MNIST (MN), CIFAR-10 (C10), EUROSAT (ES) and RESISC45 (RS). The threshold $\tau_{\mathcal{X}}(q)$ is chosen for each quantile $q$ according to \eqref{eq:tauxqrule}. The number of anchor samples $K$ for train and test splits can be found in Table \ref{tab:classification-parameters}.}
\label{fig:classification-kersize-vs-qfull}
\end{figure}

\noindent In addition to Figure \ref{fig:classification-kersize-vs-q} indicating the choice of quantiles for the input space thresholds $\tau_{\mathcal X}$, the Figure \ref{fig:classification-kersize-vs-qfull} contains a full ablation across all quantiles $q \in [0,1]$.

\subsubsection{Sensitivity of the average kernel size to noise}

\begin{figure}[h!]
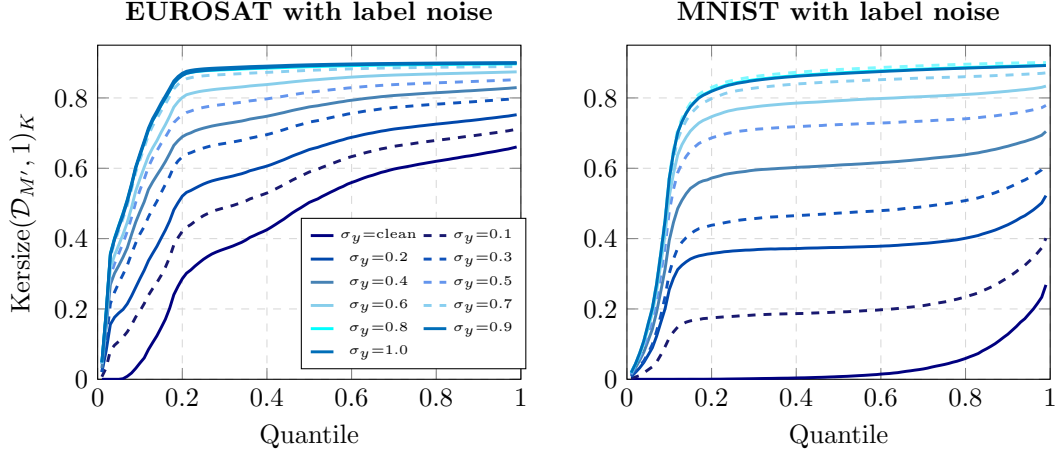

\begin{minipage}[t]{0.49\textwidth}
    \centering

\end{minipage}
\caption{EUROSAT and MNIST train datasets plots of average kernel size for different label noise levels versus the quantile $q$ used for computing the threshold $\tau_{\mathcal X}(q)$. The number of anchor samples $K$ is the full train set size (cmp. Table \ref{tab:classification-parameters}) and the average kernel size is computed with $N_{max}=200$.}\label{fig:aleatoric-quantiles}
\end{figure}

We applied different levels of label noise on MNIST and EUROSAT and computed the average kernel sizes across quantiles. Figure~\ref{fig:aleatoric-quantiles} shows the estimated average kernel size as a function of the quantile $q$ used to compute the threshold \(\tau_{\mathcal X}(q)\). For both datasets, the average kernel size increases with both the quantile $q$ and the target noise level \(\sigma_y\). Larger quantiles produce larger thresholds \(\tau_{\mathcal X}(q)\), thus Algorithm~\ref{alg:feasset} yields larger feasible sets and therefore more heterogeneous targets in a given feasible set, while larger noise levels increase label dispersion within these sets. The MNIST $\sigma_y=0$ curve in Figure~\ref{fig:aleatoric-quantiles} remains near zero for small quantiles. In contrast, EUROSAT exhibits substantial average kernel size already at moderate quantiles, indicating greater intrinsic ambiguity under the chosen pseudo-metric. Overall, the plots show that the average kernel size increases consistently with increasing aleatoric label uncertainty.

\subsection{Ablation studies}\label{sec:ablationstudies}

\begin{figure}[ht]
    \centering
    \begin{tikzpicture}

        \begin{groupplot}[
            group style={
                group size=1 by 2,
                horizontal sep=1.0cm,
                vertical sep=1.5cm
            },
]

    \nextgroupplot[
width=14cm,
height=5cm,
title={\textbf{Concrete strength $N_{max}$ ablation}},
ylabel={\small{$\operatorname{Kersize}(\mathcal D_{M'},2)_{K}$ and $\tau_{\mathcal{X}}$}},
grid=both,
    legend style={
    at={(0.98,0.08)},
    anchor=south east,
    font=\tiny,
    legend columns=1,
    },
xtick={10,50,70,100,200,300},
xmin=5,
xmax=300,
ymin=0,
ymax=3.1,
]

\addplot[
color=NavyBlue,
mark=*,
thick
]
coordinates {
(10,2.813861)
(20,2.813861)
(30,2.813861)
(50,2.813861)
(70,2.813861)
(100,2.813861)
(200,2.813861)
(300,2.813861)
};
\addlegendentry{AK}

\addplot[
color=BrickRed,
dashed,
thick,
mark=square*
]
coordinates {
(10,0.000000)
(20,0.000000)
(30,0.000000)
(50,0.000000)
(70,0.000000)
(100,0.000000)
(200,0.000000)
(300,0.000000)
};
\addlegendentry{$\tau_{\mathcal{X}}$}

\nextgroupplot[
width=14cm,
height=5cm,
xlabel={$N_{max}$},
title = {\textbf{Critical heat flux $N_{max}$ ablation}},
grid=both,
    legend style={
    at={(0.98,0.08)},
    anchor=south east,
    font=\tiny,
    legend columns=1,
    },
xtick={10,50,70,100,200,300},
xmin=5,
xmax=300,
ymin=0,
ymax=250,
]

\addplot[
color=NavyBlue,
mark=*,
thick
]
coordinates {
(10,230.473)
(20,206.318)
(30,198.335)
(50,191.051)
(70,191.314)
(100,201.304)
(200,212.969)
(300,212.969)
};
\addlegendentry{AK}

\addplot[
color=BrickRed,
dashed,
thick,
mark=square*
]
coordinates {
(10,171.840)
(20,171.840)
(30,171.840)
(50,171.840)
(70,171.840)
(100,171.840)
(200,171.840)
(300,171.840)
};
\addlegendentry{$\tau_{\mathcal{X}}$}

        \end{groupplot}

    \end{tikzpicture}
    \caption{Average kernel size, $AK$ or also $\operatorname{Kersize}(\mathcal D_{M'},2)_{K}$, for regression datasets as a function of the maximum number of nearest neighbors $N_{max}$ used in the threshold $\tau_{\mathcal{X}}$ and feasible set computation. For CHF the thresholds are chosen based on the worst-case noise and for concrete strength the input threshold is chosen as the $q^{th}$ quantile of the nearest-neighbor distances according to \eqref{eq:tauxqrule}. The quantile $q$ and number of anchor samples $K$ for train and test splits can be found in Table \ref{tab:regression-parameters}.}
\label{fig:regression-kersize-nmax-ablation2}
\end{figure}

\begin{table}[h!]
\centering
\small
\setlength{\tabcolsep}{15pt}
\begin{tabular}{l c c c c c}
\toprule
Model & RMSE$_{\mathrm{train}}$ & RMSE$_{\mathrm{test}}$ &
RMSE$_{D'}$ & $\widehat{L}$ \\
\midrule
$\phi_{\mathrm{GP}}^{o}$   & 0.000297 & 0.150666 & 0.145760 & 2.113529 \\
$\phi_{\mathrm{GP}}$       & 0.014644 & 0.016850 & 0.021984 & 0.930406 \\
$\phi_{\mathrm{SVR}}^{o}$  & 0.000177 & 0.309775 & 0.304252 & 3.889222 \\
$\phi_{\mathrm{SVR}}$      & 0.013384 & 0.016177 & 0.022371 & 0.921996 \\
\bottomrule
\end{tabular}

\caption{Different regressors errors in RMSE. The average kernel size is computed from the reference dataset $\mathcal{D}_{M'_{\mathrm{train}}}$. Models $\phi_n^{o}$, the GP and SVR, satisfy $\mathrm{RMSE}_{\mathrm{train}}<\frac12\operatorname{Kersize}$, whereas models $\phi_n$ satisfy $\frac12\operatorname{Kersize}\leq\mathrm{RMSE}_{\mathrm{train}}\leq\operatorname{Kersize}$. The table reports the training RMSE, the test RMSE, the RMSE on the constructed dataset $\mathcal{D}_{M'_{\mathrm{test}}}$, and the empirical Lipschitz estimate $\widehat{L}$. $M_{\mathrm{train}}=90$, $M_{\mathrm{test}}=810$, $M'_{\mathrm{train}}=650$ and $M'_{\mathrm{test}}=42.9k$.}
\label{tab:all-model-trade-off}
\end{table}

\begin{table}[h!]
\centering
\small
\setlength{\tabcolsep}{15pt}
\begin{tabular}{l c c c c}
\toprule
Model & RMSE$_{\mathrm{train}}$ & $1/\tau_{\mathcal{X}}(\tau_{\mathcal{Y}}-2\eta)$ & $\widehat{L}$ \\
\midrule
$\phi_{\mathrm{GP}}^{o}$   & 0.0003 & 2.100 & 2.11 \\
$\phi_{\mathrm{GP}}$       & 0.0146 & 0.051 & 0.930 \\
$\phi_{\mathrm{SVR}}^{o}$  & 0.0002 & 2.118 & 3.889 \\
$\phi_{\mathrm{SVR}}$      & 0.0134 & 0.231 & 0.922 \\
\bottomrule
\end{tabular}

\caption{We set $\eta = \text{RMSE}_{\mathrm{train}}$ and compute the quantity $1/\tau_{\mathcal{X}}(\tau_{\mathcal{Y}}-2\eta)$ which relates to the lower bound in Proposition \ref{prop:instabilityaccuracy} by taking the worst-case distance of similar targets and instead of the pointwise error the average error.}
\label{tab:all-model-trade-offquant}
\end{table}

\begin{figure}[h!]
\centering

\begin{minipage}[t]{0.47\linewidth}
\centering
\captionof{table}{Average kernel size, $AK$ or also $\operatorname{Kersize}(\mathcal D_{M'},2)_{K}$, for increasing input noise levels.}
\label{tab:kernel_size_noise}

\small
\setlength{\tabcolsep}{3pt}

\begin{tabular}{c|c|c|c}
\hline
$\sigma_x$ & AK & $M'$ & $\tau_{\mathcal{X}}$\\
\hline
0.01 & 0.00596 & 634.4k & 0.01414\\
0.05 & 0.00710 & 653.4k & 0.07071\\
0.10 & 0.01119 & 698.9k & 0.14142\\
0.20 & 0.02458 & 855.1k & 0.28284\\
0.30 & 0.04173 & 1087.6k & 0.42426\\
0.40 & 0.06339 & 1109.5k & 0.56569\\
0.50 & 0.08677 & 1093.0k & 0.70711\\
0.60 & 0.10327 & 1093.3k & 0.84853\\
0.70 & 0.13502 & 1035.2k & 0.98995\\
0.80 & 0.15497 & 1020.3k & 1.13137\\
0.90 & 0.18044 & 978.6k & 1.27279\\
1.00 & 0.20708 & 941.3k & 1.41421\\
\hline
\end{tabular}
\end{minipage}
\hfill
\begin{minipage}[t]{0.47\linewidth}
\centering

\captionof{table}{Average kernel size, $AK$ or also $\operatorname{Kersize}(\mathcal D_{M'},2)_{K}$, for increasing target noise levels.}
\label{tab:kernel_size_target_noise}

\small
\setlength{\tabcolsep}{3pt}

\begin{tabular}{c|c|c}
\hline
$\sigma_y$ & AK & $M'$\\
\hline
0.01 & 0.00596 & 634.4k\\
0.05 & 0.01556 & 733.2k\\
0.10 & 0.02552 & 742.2k\\
0.20 & 0.04741 & 746.8k\\
0.30 & 0.07155 & 744.7k\\
0.40 & 0.09789 & 745.1k\\
0.50 & 0.11382 & 748.1k\\
0.60 & 0.14349 & 746.9k\\
0.70 & 0.16264 & 745.7k\\
0.80 & 0.19173 & 745.1k\\
0.90 & 0.21385 & 745.7k\\
1.00 & 0.23710 & 745.1k\\
\hline
\end{tabular}
\end{minipage}

\end{figure}

\begin{figure}[ht]
    \centering
    \begin{tikzpicture}

        \begin{groupplot}[
            group style={
                group size=1 by 4,
                horizontal sep=1.0cm,
                vertical sep=1.5cm
            },
]

    \nextgroupplot[
width=14cm,
height=5cm,
title = {\textbf{Tropical Cyclone quantile ablation}},
ylabel={\small{$\operatorname{Kersize}(\mathcal D_{M'},2)_{K}$ and $\tau_{\mathcal{X}}(q)$}},
xmin=0.01, xmax=0.10,
ymin=0.5, ymax=8.2,
xtick={0.02,0.03,0.04,0.05,0.06,0.07,0.08,0.09,0.10},
ytick={2.0,3.0,4.0,5.0,6.0,7.0,8.0},
grid=major,
    legend style={
    at={(0.98,0.46)},
    anchor=south east,
    font=\tiny,
    legend columns=2,
    },
mark options={solid},
]


\addplot[
NavyBlue,
thick,
mark=*,
]
coordinates {
(0.01,7.531868)
(0.02,7.535607)
(0.03,7.541543)
(0.04,7.544380)
(0.05,7.555072)
(0.06,7.579909)
(0.07,7.601494)
(0.08,7.650106)
(0.09,7.792265)
(0.10,8.004868)
};

\addlegendentry{AK (train)}


\addplot[
RoyalBlue,
thick,
mark=triangle*,
]
coordinates {
(0.01,7.274664)
(0.02,7.274180)
(0.03,7.272130)
(0.04,7.269701)
(0.05,7.270748)
(0.06,7.272526)
(0.07,7.274562)
(0.08,7.276887)
(0.09,7.287913)
(0.10,7.296332)
};

\addlegendentry{AK (test)}


\addplot[
BrickRed,
thick,
dashed,
mark=square*,
]
coordinates {
(0.01,1.074639)
(0.02,1.361244)
(0.03,1.585607)
(0.04,1.804525)
(0.05,2.061460)
(0.06,2.397903)
(0.07,2.529677)
(0.08,2.643076)
(0.09,2.777048)
(0.10,2.941170)
};

\addlegendentry{$\tau_{\mathcal{X}}/100$ (train)}

\addplot[
Orange,
thick,
dashed,
mark=diamond*,
]
coordinates {
(0.01,0.935613)
(0.02,1.135205)
(0.03,1.419367)
(0.04,1.593978)
(0.05,1.724738)
(0.06,1.852423)
(0.07,1.955059)
(0.08,2.020739)
(0.09,2.111952)
(0.10,2.202165)
};

\addlegendentry{$\tau_{\mathcal{X}}/100$ (test)}

\nextgroupplot[
width=14cm,
height=5cm,
title={\textbf{Energy efficiency $Y_1$ quantile ablation}},
xmin=0.01, xmax=0.10,
ymin=0.50, ymax=0.70,
xtick={0.02,0.03,0.04,0.05,0.06,0.07,0.08,0.09,0.10},
ytick={0.50,0.54,0.58,0.62,0.66,0.70},
grid=major,
    legend style={
    at={(0.98,0.01)},
    anchor=south east,
    font=\tiny,
    legend columns=1,
    },
mark options={solid},
]

\addplot[
NavyBlue,
thick,
mark=*,
]
coordinates {
(0.01,0.553437)
(0.02,0.553437)
(0.03,0.553437)
(0.04,0.553437)
(0.05,0.553437)
(0.06,0.553437)
(0.07,0.553437)
(0.08,0.553437)
(0.09,0.553437)
(0.10,0.553437)
};

\addlegendentry{AK}

\addplot[
BrickRed,
thick,
dashed,
mark=square*,
]
coordinates {
(0.01,0.645182)
(0.02,0.645182)
(0.03,0.645182)
(0.04,0.645182)
(0.05,0.645182)
(0.06,0.645182)
(0.07,0.645182)
(0.08,0.645182)
(0.09,0.645182)
(0.10,0.645182)
};

\addlegendentry{$\tau_{\mathcal{X}}$}

\nextgroupplot[
width=14cm,
height=5cm,
title={\textbf{Energy efficiency $Y_2$ quantile ablation}},
xmin=0.01, xmax=0.10,
ymin=0.0, ymax=1.7,
xtick={0.02,0.03,0.04,0.05,0.06,0.07,0.08,0.09,0.10},
ytick={0.0,0.4,0.8,1.2,1.6},
grid=major,
    legend style={
    at={(0.98,0.08)},
    anchor=south east,
    font=\tiny,
    legend columns=1,
    },
mark options={solid},
]

\addplot[
NavyBlue,
thick,
mark=*,
]
coordinates {
(0.01,1.546744)
(0.02,1.546744)
(0.03,1.546744)
(0.04,1.546744)
(0.05,1.546744)
(0.06,1.546744)
(0.07,1.546744)
(0.08,1.546744)
(0.09,1.546744)
(0.10,1.546744)
};

\addlegendentry{AK Y2}

\addplot[
BrickRed,
thick,
dashed,
mark=square*,
]
coordinates {
(0.01,0.645182)
(0.02,0.645182)
(0.03,0.645182)
(0.04,0.645182)
(0.05,0.645182)
(0.06,0.645182)
(0.07,0.645182)
(0.08,0.645182)
(0.09,0.645182)
(0.10,0.645182)
};

\addlegendentry{$\tau_{\mathcal{X}}$ Y2}

        \nextgroupplot[width=14cm,
height=5cm,
xlabel={$q$},
title={\textbf{Concrete strength quantile ablation}},
xmin=0.01,
xmax=0.10,
ymin=0.00,
ymax=3.30,
xtick={0.02,0.03,0.04,0.05,0.06,0.07,0.08,0.09,0.10},
ytick={0.50,1.00,1.50,2.00,2.50,3.00},
grid=major,
    legend style={
    at={(0.98,0.08)},
    anchor=south east,
    font=\tiny,
    legend columns=1,
    },
mark options={solid},
]

\addplot[
NavyBlue,
thick,
mark=*,
]
coordinates {
(0.01,2.813861)
(0.02,2.813861)
(0.03,2.813861)
(0.04,2.813861)
(0.05,2.813861)
(0.06,3.154930)
(0.07,3.141419)
(0.08,3.129677)
(0.09,3.124495)
(0.10,3.111451)
};

\addlegendentry{AK}

\addplot[
BrickRed,
thick,
dashed,
mark=square*,
]
coordinates {
(0.01,0.000000)
(0.02,0.000000)
(0.03,0.000000)
(0.04,0.000000)
(0.05,0.000000)
(0.06,0.719100)
(0.07,1.107900)
(0.08,1.528500)
(0.09,1.794300)
(0.10,1.897600)
};

\addlegendentry{$100\tau_{\mathcal{X}}$}

        \end{groupplot}

    \end{tikzpicture}
    \caption{Average kernel size, $AK$ or also $\operatorname{Kersize}(\mathcal D_{M'},2)_{K}$, and scaled thresholds $\tau_{\mathcal{X}}(q)$ as functions of the quantile parameter $q$ for the regression datasets. Computed on subsets of $K=1k$ samples from the tropical cyclone training and test datasets. The threshold $\tau_{\mathcal{X}}(q)$ is computed for each quantile $q$. The number of anchor samples $K$ and $N_{\max}$ are in Table \ref{tab:regression-parameters}.}
\label{fig:regression-kersize-quantile-ablation1}
\end{figure}
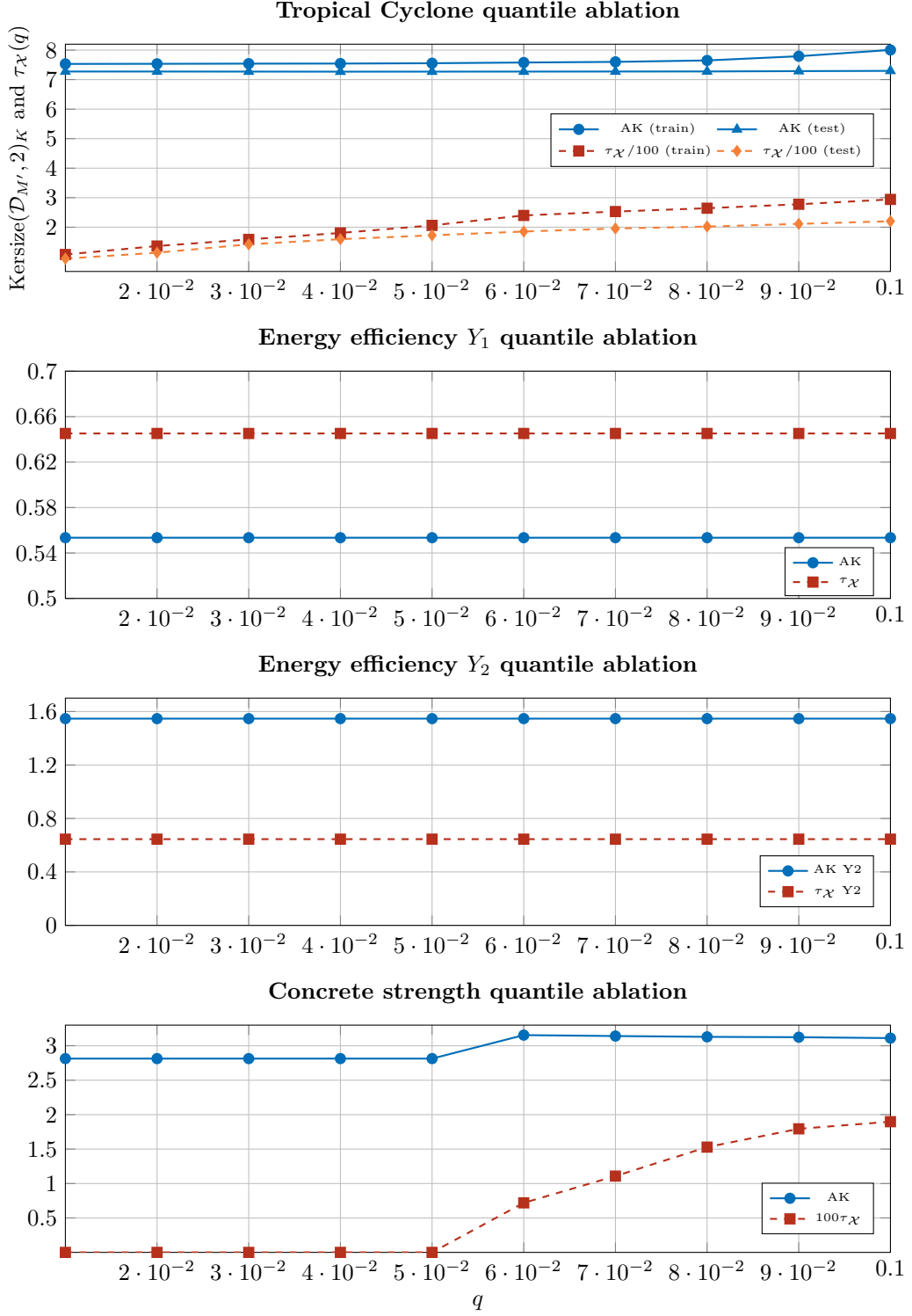

\end{document}